\documentclass[12pt]{article} 
\pdfoutput=1
\usepackage{amsmath,amsthm,amssymb,bm,mathrsfs,}
\usepackage{graphicx}
\usepackage[ruled,linesnumbered]{algorithm2e}
\usepackage{caption,enumerate,natbib,listings,setspace}
\usepackage[OT1]{fontenc}
\usepackage[colorlinks,linkcolor=red,anchorcolor=blue,citecolor=blue]{hyperref}
\usepackage[margin=1in]{geometry}
\usepackage[protrusion=false,expansion=true]{microtype}
\usepackage[title]{appendix}
\usepackage{bbm}
\usepackage{subcaption}
\usepackage{comment}
\theoremstyle{plain}
\let\hat\widehat

\newcommand{\argmin}{\mathop{\mathrm{argmin}}}

\newtheorem{lemma}{{\bf Lemma}}

\newtheorem{corollary}{{\bf Corollary}}

\newtheorem{theorem}{{\bf Theorem}}

\newtheorem{assumption}{{\bf Assumption}}

\newtheorem{definition}{{\bf Definition}}

\begin{document}

\title{\Large \bf Ranking Differential Privacy}
 
\author
{
Shirong Xu\thanks{Department of Statistics, University of California, Los Angeles. Email: shirong@stat.ucla.edu},
Will Wei Sun\thanks{Krannert School of Management, Purdue University. Email: sun244@purdue.edu.},
and
Guang Cheng\thanks{Department of Statistics, University of California, Los Angeles. Email: guangcheng@ucla.edu.}, 
} 
 

\date{}
\maketitle

\begin{abstract}
Rankings are widely collected in various real-life scenarios, leading to the leakage of personal information such as users' preferences on videos or news. To protect rankings, existing works mainly develop privacy protection on a single ranking within a set of ranking or pairwise comparisons of a ranking under the $\epsilon$-differential privacy. This paper proposes a novel notion called $\epsilon$-ranking differential privacy for protecting ranks. We establish the connection between the Mallows model \citep{mallows1957non} and the proposed $\epsilon$-ranking differential privacy. This allows us to develop a multistage ranking algorithm to generate synthetic rankings while satisfying the developed $\epsilon$-ranking differential privacy. Theoretical results regarding the utility of synthetic rankings in the downstream tasks, including the inference attack and the personalized ranking tasks, are established. For the inference attack, we quantify how $\epsilon$ affects the estimation of the true ranking based on synthetic rankings. For the personalized ranking task, we consider varying privacy preferences among users and quantify how their privacy preferences affect the consistency in estimating the optimal ranking function. Extensive numerical experiments are carried out to verify the theoretical results and demonstrate the effectiveness of the proposed synthetic ranking algorithm.
\end{abstract}

\baselineskip=24pt
\newpage
\section{Introduction}
Ranking data commonly arises from various business scenarios, such as recommender systems \citep{karatzoglou2013learning,oliveira2020rank} and search engines \citep{dwork2001rank,liu2007supervised}. Generally, rankings are collected and utilized to learn users' preferences to items for providing appropriate recommendations in the future. A typical example is the recommender system, which suggests new items for incoming users by pooling historical information on users' behaviors. Nevertheless, users' rankings are usually highly sensitive since they reveal their purchasing or political preferences \citep{yang2019collecting,lee2015efficient}. Therefore, developing an effective mechanism to achieve privacy protection of ranking data before being shared with an external party is of great need.

To ensure data privacy protection, it is a common practice to employ differential privacy (DP; \citealt{dwork2006differential}) as a standard metric. Differential privacy formalizes privacy guarantee in the mathematical language without imposing assumptions about data, possessing nice properties such as immunity to post-processing and privacy composition for sanitized data output by privacy-preserving mechanisms. Generally, differential privacy can be categorized into two main classes, including central differential privacy (CDP; \citealt{dwork2006our}) and local differential privacy (LDP; \citealt{wang2017locally}). LDP perturbs data on the users' side and submits privatized data to servers, whereas CDP relies on a trusted data collector to protect all data. LDP become popular in privacy protection and has now been applied in various real scenarios, including Google Chrome browser \citep{erlingsson2014rappor} and macOS \citep{tang2017privacy}.


In the literature, various research efforts have been devoted to developing an effective mechanism to protect ranking data under the differential privacy and establishing theoretical results regarding statistical inference based on privatized ranking data. Depending on the employment of either LDP or CDP, the privacy protection of ranking data can be divided into two main classes. In the central model of DP, a trusted curator collects non-private rankings from all users and carries out differentially private statistical inference \citep{hay2017differentially,lee2015efficient,shang2014application,lee2015efficient,sanchez2016utility,busa2021private}. Specifically, \citet{lee2015efficient} developed efficient algorithms for eliciting the true ranking of items under the central differential privacy and strategic manipulation. \citet{shang2014application} proposed to utilize the Gaussian noise to contaminate the histogram of collected rankings for rank aggregation. \citet{hay2017differentially} proposed several differentially private rank aggregation algorithms using Laplace noise to protect single ranking from the released output. \citet{li2022differentially} embedded the Laplace, the randomized response, and the exponential mechanisms into Condorcet voting, developing a novel family of randomized voting rules for protecting privacy of rankings. By contrast, in the local model of DP, ranking data are privatized via some local mechanisms before they are submitted to a curator \citep{yang2019collecting,song2022distributed,yan2020private,alabi2022private}. \citet{yan2020private} employed the Laplace noise or the randomized response mechanisms to randomly permute pairwise comparison preferences of ranks. \citet{song2022distributed} proposed to collect differentially private rankings locally for rank aggregation, which randomly permutes rankings via adding Gaussian noise to pairwise comparisons between items.

In this paper, we develop a novel notion called $\epsilon$-ranking differential privacy to protect the position of any single item in a ranking, where a smaller $\epsilon$ leads to more stringent privacy protection. Specifically, we first establish the definition of neighboring ranking that two rankings are viewed as neighbors if they have identical partial orders between items when one item is removed, and a synthetic ranking algorithm should have similar output distributions for two neighboring rankings as input. The key motivation is that existing works mainly focus on protecting pairwise comparisons between items' ranks or a single ranking within a set of rankings, formalizing privacy guarantees under the classical $\epsilon$-differential privacy, whereas few attempts have been made to directly protect the positions of items in a ranking. A key disadvantage of the application of the classical $\epsilon$-differential privacy to rankings is that, for a multidimensional object like a ranking, the privacy protection usually relies on the composition theorem for privacy accounting \citep{shang2014application,jeong2022ranking}, which suffers from a loose privacy bound and complicated downstream privacy-utility analyses. In contrast, the proposed $\epsilon$-ranking DP uses a single privacy parameter $\epsilon$ to measure the degree of privacy protection for a ranking instead of relying on the composition theorem \citep{kairouz2015composition}, avoiding an inaccurate computation of privacy budget. Under the developed $\epsilon$-ranking DP, we are capable of analyzing how privacy guarantee $\epsilon$ for each ranking ranking affects the performance of downstream inference tasks based on privacy-preserving rankings. For example, the proposed $\epsilon$-ranking DP allows for personalized privacy preferences in the personalized ranking task that users can control the privacy preferences over their rankings. Additionally, classical data perturbation methods like the Laplace noise addition fails to match with the ordinal nature of ranks, implicitly leading to a suboptimal privacy-utility tradeoff for downstream tasks. We conducted extensive experiments to validate our theoretical results and demonstrate the effectiveness of the synthetic ranking algorithm, showing that the synthetic ranking algorithm produces privacy-preserving rankings with more utility for downstream personalized learning task, which implicitly results from a better privacy-utility tradeoff.

The contributions of this paper are summarized as follows. First, we develop an synthetic ranking algorithm based on the multistage ranking algorithm \citep{fligner1988multistage,critchlow1991probability} to generate privacy-preserving rankings, which is proved to be advantageous over the linear Laplace noise addition in retaining more information of ranking regarding pairwise comparisons under the same privacy guarantee. The proposed algorithm essentially generates synthetic rankings via the Mallows model \citep{mallows1957non}, establishing a surprising connection between the Mallows model and the $\epsilon$-ranking DP. Second, we analyze the utility of the synthetic rankings by two downstream tasks, including the inference attack of a ranking and the personalized ranking task. In the inference attack, we theoretically quantify how $\epsilon$ affects the estimation of the central ranking based on generated synthetic ranking, providing an optimal rate of $\epsilon$ adaptive to the number of synthetic rankings for not correctly inferring the central ranking. For the personalized ranking task, we allow users to have personalized privacy preferences and theoretically quantify the relation between users' privacy preferences and the regret in estimating the optimal ranking function. Particularly, we derive the optimal order of privacy parameter $\epsilon$ adaptive to the number of users given the consistency in estimating the optimal ranking function. Our theoretical results show that when all users choose an adaptive privacy guarantee at the order $O\big(\sqrt{n^{-1}\log^{1+\zeta}(n)}\big)$ for any $\zeta>0$ the consistency in estimating the optimal ranking function is guaranteed. Interestingly, our theoretical result is similar to that of \citet{duchi2018right} in providing similar quantitative effect of $\epsilon$ on the convergence rate of estimation under the local models
of privacy.

The rest of the paper proceeds as follows. After introducing some necessary notations in Section \ref{Sec:Not}, Section \ref{Sec:Pre} introduces the backgrounds of differential privacy, ranking data, and the Mallows model. Section \ref{Sec:RDP} formalizes the definition of the $\epsilon$-ranking differential privacy and develops a synthetic ranking algorithm satisfying $\epsilon$-ranking DP. Section \ref{Sec:IF} establishes theoretical results concerning the inference attack of the central ranking used to generate synthetic rankings. Section \ref{Sec:PR} considers the situation that synthetic rankings are used for downstream personalized ranking task, for which we establish theoretical results concerning the consistency in estimating the optimal ranking function and quantify the corresponding the privacy-utility tradeoff. Section \ref{Sec:Expe} conducts extensive experiments to verify your theoretical results. A brief summary is provided in Section \ref{Sec:Summ} and all technical proofs are provided in the Appendix.

\subsection{Notation}
\label{Sec:Not}
For a positive integer $n$, denote $[n]=\{1, ... , n\}$ to be the $n$-set. For a set $S$, we let $|S|$ denote its cardinality. For two positive sequences $\{f_n\}_{n=1}^{\infty}$ and $\{g_n\}_{n=1}^{\infty}$, we denote that $f_n = O(g_n)$ if $\limsup_{n\rightarrow \infty} |f_n|/g_n <+\infty$. We let $f_n \asymp g_n$ if $f_n = O(g_n)$ and $g_n = O(f_n)$. For a random variable $X_n$ and a sequence $\{a_n\}_{n=1}^{\infty}$, we denote that $X_n = o_p(a_n)$ is $X_n/a_n$ converges to zero in probability and $X_n = O_p(a_n)$ if $X_n/a_n$ is stochastically bounded. For an integer $K$, $\Upsilon(K)$ denotes the set of permutations of $(1,2,\ldots,K)$. Let $I(\cdot)$ be the indicator function and $I(A)=1$ if $A$ holds true and 0 otherwise. For a vector $\bm{x}$, we let $\Vert \bm{x}\Vert_2$ denote its $l_2$-norm and $\Vert \bm{x}\Vert_{\infty}$ denote its $l_{\infty}$-norm.

\section{Preliminaries}
\label{Sec:Pre}
This section introduces  some basic concepts relating to differential privacy, ranking data, and the Mallows model \citep{mallows1957non,fligner1986distance} for ranking data, which paves the way for us to propose a novel variant of the differential privacy for ranking data.

\subsection{Differential Privacy}
Differential privacy has emerged as a rigorous framework for measuring the capacity of a randomized mechanism in privacy protection, which is reflected by bounding the discrepancy in the output distributions when any single record of the input changes. The most popular definition of differential privacy is $(\epsilon,\delta)$-differential privacy.

\begin{definition}
$\big( (\epsilon,\delta)$-differential privacy$\big)$ Let $\epsilon>0$ and $\delta \in [0,1)$. Let $S$ and $S'$ be two sets of records of same length and define $d:\mathcal{S}\times \mathcal{S}\rightarrow \mathbb{Z}_{ \geq 0}$ be the Hamming distance between two sets. We say a randomized mechanism $\mathcal{M}: \mathcal{S}\rightarrow \mathcal{Z}$ satisfies $(\epsilon,\delta)$-differential privacy with respect to $d$ if for any $S,S' \in \mathcal{S}$ such that $d(S,S')=1$, we have
\begin{align*}
\mathbb{P}(\mathcal{M}(D) \in Z) \leq e^\epsilon \mathbb{P}(\mathcal{M}(D') \in Z)+\delta,
\end{align*}
for any $Z \subset \mathcal{Z}$. If $\delta=0$, then $\mathcal{M}$ satisfies pure $\epsilon$-differential privacy.
\end{definition}

The intuition behind differential privacy is that inference on any single record in a dataset is inaccurate in the sense that the output distribution stays less affected by the change of any single record in the input dataset.

\subsection{Ranking Data and The Mallows Model}
In ranking data, a user ranks a set of items according to a specific criterion. A typical example of ranking is the preference ranking, which assigns ordinal ranks to items according to relative preferences. Let $\Omega=\{\mathcal{I}_1,\ldots,\mathcal{I}_m\}$ denote a set of $m$ items and $\phi_{\Omega}$ denote a ranking of items in $\Omega$. The ranking $\phi_{\Omega}$ usually appears as an ordered list indicating the positions of items in $\Omega$ under a specific metric, i.e., $\phi_{\Omega}(\mathcal{I}_i)=k$ means that the rank of item $\mathcal{I}_i$ is $k$. Without loss of generality, we let the item with rank $1$ refer to the most preferred item, and hence items with higher ranks are less preferred in our setting. For example, a preference ranking $\phi_{\Omega} = (\phi_{\Omega}(\mathcal{I}_1),\phi_{\Omega}(\mathcal{I}_2),\phi_{\Omega}(\mathcal{I}_3))=(3,2,1)$ indicates that item $3$ is the most preferred item and item $1$ is the least preferred one. For ease of notation, $\phi_{\Omega}$ will be abbreviated as $\phi$ in the sequel when it causes no confusion.


Let $\Phi$ denote the random variable of the observed ranking $\phi$, and $\Phi$ is a multi-variate random variable taking values in $\Upsilon(|\Omega|)$, where $\Upsilon(|\Omega|)$ denotes the set of all permutations of $(1,\ldots,|\Omega|)$. Naturally, we suppose that $\mathbb{P}(\Phi(\mathcal{I}_i)>\Phi(\mathcal{I}_j))>\mathbb{P}(\Phi(\mathcal{I}_j)>\Phi(\mathcal{I}_i))$ if and only if the item $\mathcal{I}_j$ is better than the item $\mathcal{I}_i$ in quality. Here the randomness of $\mathbb{P}(\Phi(\mathcal{I}_i)>\Phi(\mathcal{I}_j))$ comes from the randomness of the observed ranking. This is a mild assumption that is fulfilled for various ranking models, such as the random utility model \citep{walker2002generalized,soufiani2014computing,su2021you} and the Mallows model \citep{mallows1957non,JMLR:v23:21-1262}.

The Mallows model \citep{mallows1957non} is a popular parametric model to model ranking data. Specifically, the Mallows model with Kendall-$\tau$ distance \citep{fligner1986distance,mandhani2009tractable} generates synthetic rankings based on $\phi_0$. Specifically, a ranking $\phi$ will be generated with a higher probability if $\phi$ is more aligned with $\phi_0$ in pairwise comparisons of items' ranks. The generation of rankings follows the distribution as
\begin{align*}
\mathbb{P}_{\theta,\phi_0}(\phi) = 
\frac{1}{\Psi(\theta)}
\exp\Big(\theta
T(\phi,\phi_0)
\Big), \mbox{ for any } \phi \in \Upsilon(|\Omega|),
\end{align*}
where $\phi_0$ is the central ranking over the item set $\Omega$, $\Psi(\theta) = \sum_{\phi \in \Upsilon(|\Omega|)}\exp\big(\theta T(\phi,\phi_0)\big)$, $\theta$ is the dispersion parameter, and $T(\phi,\phi_0 )= \sum_{\mathcal{I}_i,\mathcal{I}_j\in \Omega}I\big(\big(\phi(\mathcal{I}_i)-\phi(\mathcal{I}_j)\big)\big(\phi_0(\mathcal{I}_i)-\phi_0(\mathcal{I}_j)\big)>0\big)$ can be viewed the number of concordant pairs, characterizing the difference between $\phi$ and $\phi_0$. The Mallows model is widely used to model the distribution of preference rankings \citep{desir2016assortment,busa2014preference}, and $\phi_0$ usually refers to the true ranking of items. For example, in recommender systems, $\phi_0$ refers to the true ranking of a set of items in quality, and $\phi$ is the observed ranking by a specific user.

\section{Ranking Differential Privacy}
\label{Sec:RDP}
Users' ranking data are extremely sensitive in revealing their preferences and behaviors, which are potentially interesting for marketing purposes \citep{jeckmans2013privacy}. Therefore, it is of great importance to protect privacy of preference rankings before sharing them to an external party. In this section, we propose the $\epsilon$-ranking differential privacy. Subsequently, we establish a connection between $\epsilon$-ranking DP and the Mallows model, which can be utilized to generate privacy-preserving rankings satisfying the proposed $\epsilon$-ranking DP. The connection is essentially derived from an inherent connection between the Mallows model and the exponential mechanism of differential privacy \citep{mcsherry2007mechanism,lantz2015subsampled}.

Ranks of items indicate their positions among the set of items and implicitly reveal their partial orders. Intuitively, in order to protect a ranking, the ranks of items should be noisy in accordance with the nature of differential privacy. To this end, we propose a new variant of neighboring set for rankings.
\begin{definition}
(Neighboring Ranking) Let $\phi$ and $\phi'$ be two ordinal rankings on the same item set $\Omega$. We say $\phi$ and $\phi'$ are neighboring rankings if there exists only one $\mathcal{I}_k \in \Omega$ such that for any $\mathcal{I}_i,\mathcal{I}_j\in \Omega \setminus \{ \mathcal{I}_k\}$ with $i \neq j$
\begin{align*}
\big(\phi(\mathcal{I}_i)-\phi(\mathcal{I}_j)\big)\big(\phi'(\mathcal{I}_i)-\phi'(\mathcal{I}_j)\big)>0.
\end{align*}
\end{definition}

The main idea of neighboring ranking is that two ordinal rankings are consistent in all partial orders of ranks of items except a specific item. Next, we propose our $\epsilon$-ranking differential privacy.
\begin{definition}
($\epsilon$-ranking Differential Privacy; $\epsilon$-ranking DP) Let $\phi$ and $\phi'$ be two neighboring ordinal rankings on the item set $\Omega$ and $\mathcal{M}(\phi)$ be a synthetic ranking algorithm producing ordinal rankings for users. We say $\mathcal{M}$ satisfies $\epsilon$-ranking differential privacy if
\begin{align*}
\sup_{\widetilde{\phi} \in \Upsilon(|\Omega|)}
\Big|\log  \frac{\mathbb{P}(\mathcal{M}(\phi)=\widetilde{\phi})}{\mathbb{P}(\mathcal{M}(\phi') =\widetilde{\phi})} \Big|
 \leq 
 \epsilon,
\end{align*}
where the randomness comes from the mechanism $\mathcal{M}$.
\end{definition}

The rationale of $\epsilon$-ranking DP is that the output distributions of $\mathcal{M}(\phi)$ and $\mathcal{M}(\phi')$ are similar in the sense that inference on the ordinal rank of a single item based on output synthetic rankings is statistically inaccurate. In other words, $\epsilon$-ranking DP protects any single rank within a ranking by ensuring that the change of any single rank in the ranking produces little effect on the output distribution of rankings. The proposed $\epsilon$-ranking DP uses a single privacy parameter $\epsilon$ to measure the degree of privacy protection.

\subsection{Privacy-Preserving Rankings}
In this section, we establish a connection between the Mallows model \citep{mallows1957non,fligner1986distance} and the proposed $\epsilon$-ranking DP, which is derived from the fact that the Mallows model belongs to the exponential family. Based on this connection, we propose a synthetic ranking algorithm to generate privacy-preserving rankings, which is developed based on the multistage ranking model \citep{fligner1988multistage,critchlow1991probability}.

\begin{lemma}
\label{lemma:eDP}
Let $\phi$ and $\phi'$ be two neighboring rankings. The Mallows model satisfies that $\big|\log
\big(
\frac{
\mathbb{P}_{\theta,\phi}(\widetilde{\phi})}{\mathbb{P}_{\theta,\phi'}(\widetilde{\phi})} \big)\big| \leq  \epsilon$ for any $\widetilde{\phi}\in \Upsilon(|\Omega|)$ given that $\theta = 2^{-1}\epsilon |\Omega|$.
\end{lemma}

Lemma $\ref{lemma:eDP}$ demonstrates a connection between the Mallows model and the $\epsilon$-ranking DP. Specifically, under an appropriate choice of the dispersion parameter $\theta = 2^{-1}\epsilon |\Omega|$, using the Mallows model to generate synthetic rankings satisfies the proposed $\epsilon$-ranking DP. In other words, protecting an observed ranking $\phi$ can be achieved by generating a counterpart ranking $\widetilde{\phi}$ via the Mallows model with $\phi$ being the input ranking, where the privacy guarantee is guarded under the developed $\epsilon$-ranking DP.

\begin{figure}[h!]
    \centering
  \includegraphics[scale=0.65]{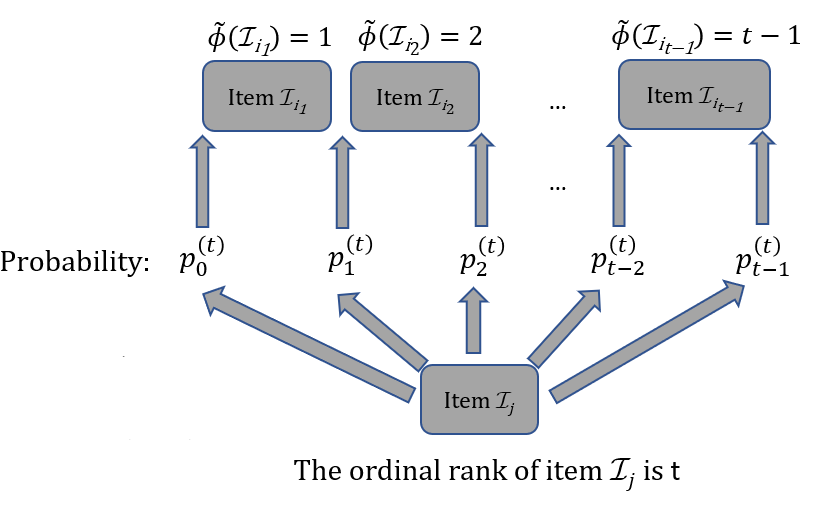}
    \caption{The $t$-th iteration of the proposed synthetic ranking algorithm.}
    \label{fig:algorithm}
\end{figure}

To generate privacy-preserving rankings via the Mallows model, we propose an algorithm based on the idea of the multistage ranking model \citep{fligner1988multistage,critchlow1991probability}. The overall idea of the developed algorithm can be viewed as a shuffling mechanism. To be more specific, the algorithm sequentially determines the ranks of items according to their ranks in $\phi$. As illustrated in Figure $\ref{fig:algorithm}$, the position of the item $\mathcal{I}_j$ in the synthetic ranking $\widetilde{\phi}$ is determined by the sampling procedure with probabilities $p_{i}^{(t)},i=1,\ldots,t-1$. Here it should be noted that $p_{i}^{(t)}>p_{j}^{(t)}$ for any $i>j$ since position $i$ is more aligned with the position of item $\mathcal{I}_j$ in $\phi$. The overall algorithm is summarized in Algorithm \ref{algorithm1}.

\begin{algorithm}[ht!]
		\SetKwInOut{Input}{Input}
		\SetKwInOut{Output}{Output}
		\caption{Privacy Preserving Ranking Algorithm}\label{algorithm1}
		\Input{The ranking $\phi$ of the item set $\Omega$ and privacy guarantee $\epsilon$ \\}
 \textbf{Initialization:} Let $\phi^{-1}$ be the inverse function of ranking and choose the item $\mathcal{I}_i$ satisfying $\phi^{-1}(1)=\mathcal{I}_i$ from $\Omega$ and set $\widetilde{\phi}(\mathcal{I}_i) = 1$ and $\chi^{(1)} = \{i \}$;	
		
		\For{$t=2,\ldots,|\Omega|$}{
		Select the item $\mathcal{I}_j$ from $\Omega$ such that the rank of $\mathcal{I}_j$ is $t$ and define $$\rho^{(t)} = \{0,\ldots,t-1  \}.$$ \\

Compute values 
$$
\tau(k,\chi^{(t-1)}) = \sum_{l \in \chi^{(t-1)}}
I \Big(
\big(t-
\phi(\mathcal{I}_l) \big)
\big(k+0.5 - \widetilde{\phi}(\mathcal{I}_l)  \big)>0
\Big),k \in \rho^{(t)}.
$$

Sample a value $k$ from $\rho^{(t)}$ and assign it to $V^{(t)}_{\epsilon}$ according to the probability as
$$
p_{k}^{(t)} = 
\mathbb{P}(V^{(t)}_{\epsilon} = k) =\frac{\exp\Big(
\epsilon(|\Omega|-1)^{-1} \tau(k,\chi^{(t-1)})
\Big)}{\sum_{k \in \rho^{(t)}}\exp\Big(
\epsilon(|\Omega|-1)^{-1} \tau(k,\chi^{(t-1)})
\Big)}
, k \in \rho^{(t)}.
$$ \\
Define $G^{(t)}(V^{(t)}_{\epsilon}) = \{i: \widetilde{\phi}(\mathcal{I}_i)> V^{(t)}_{\epsilon}+0.5, i \in \chi^{(t-1)} \}$ and rearrange the ranking as
\begin{align*}
&\widetilde{\phi}(\mathcal{I}_i) \gets \widetilde{\phi}(\mathcal{I}_i)+1, \mbox{ for }i \in G^{(t)}(V^{(t)}_{\epsilon}),\\
&\widetilde{\phi}(\mathcal{I}_j) \gets V^{(t)}_{\epsilon} + 1.
\end{align*}

$\chi^{(t)} \gets \chi^{(t-1)} \cup \{ j \}$ \\

		}
\textbf{Output:} Output synthetic ranking $\widetilde{\phi}$.	
	\end{algorithm}

Algorithm \ref{algorithm1} takes the ranking $\phi$ as an input and outputs a synthetic ranking $\widetilde{\phi}$. In Algorithm \ref{algorithm1}, the synthetic ranks of items are determined sequentially in a probabilistic manner such that positions with more similar partial orders to the input ranking are more likely to be chosen. Specifically, at the initialization step, the item with rank $1$ in $\phi$ (denoted as $\mathcal{I}_i$) is chosen first, and its rank in $\widetilde{\phi}$ is set as 1 temporarily. Then the ranks of other items in the synthetic ranking $\widetilde{\phi}$ are determined in an increasing order of their ranks of $\phi$. At the $t$-th iteration, the objective is to determine the relative position of the item with rank $t$ to the first $t-1$ items whose relative positions are already specified and their item indexes are stored in $\chi^{(t-1)}$. Since positions are chosen in a probabilistic manner, we let $V_{\epsilon}^{(t)}$ denote the random variable for the position taking values in $\rho^{(t)}$, and each element $k \in \rho^{(t)}$ corresponds to a relative position. For each position $k$, $\tau(k,\chi^{(t-1)})$ represents the number of concordant pairs at position $k$, and a larger $\tau(k,\chi^{(t-1)})$ indicates a higher probability $p_k^{(t)}$ that position $k$ will be chosen. After $V^{(t)}_{\epsilon}$ is determined, the synthetic ranks of first $t$ items will be rearranged as demonstrated in step 6. The overall computational complexity of Algorithm \ref{algorithm1} is $O(|\Omega|^2)$

\begin{lemma}
\label{lemma:Prob_u}
Let $\mathcal{A}_{\epsilon}$ be Algorithm $\ref{algorithm1}$. Given a ranking $\phi$ of the item set $\Omega=\{\mathcal{I}_1,\ldots,\mathcal{I}_m\}$, the synthetic ranking $\mathcal{A}_{\epsilon}(\phi)$ follows the following distribution,
$$
\mathbb{P}(\mathcal{A}_{\epsilon}(\phi) = \widetilde{\phi}) =
\frac{
		\exp\big(2^{-1}\epsilon|\Omega| T(\phi,\widetilde{\phi}) \big)}{\sum_{\widetilde{\phi} \in \Upsilon(|\Omega|)}\exp\big(2^{-1}\epsilon|\Omega| T(\phi,\widetilde{\phi} \big)},
		\mbox{ for any $\widetilde{\phi} \in \Upsilon(|\Omega|)$},
$$
where $T(\phi,\widetilde{\phi} )= \sum_{\mathcal{I}_i,\mathcal{I}_j\in \Omega}I\big(\big(\phi(\mathcal{I}_i)-\phi(\mathcal{I}_j)\big)\big(\widetilde{\phi}(\mathcal{I}_i)-\widetilde{\phi}(\mathcal{I}_j)\big)>0\big)$.
\end{lemma}

Lemma $\ref{lemma:Prob_u}$ shows that the synthetic ranking algorithm is essentially equivalent to sampling from the Mallows model, and the basic idea is employing importance sampling to generate synthetic ranking in that a ranking with larger value of $T(\phi,\widetilde{\phi} )$ is generated with a larger probability, providing a bridge to establish privacy parameter $\epsilon$.

\begin{theorem}
\label{Thm:ConsisandPriv}
 Algorithm $1$ possesses the following properties:
\begin{itemize}

\item[(1)] Consistency in ranking optimality: For any $\mathcal{I}_i,\mathcal{I}_j\in \Omega$ with $i \neq j$, it holds true that
\begin{align*}
\Big(\mathbb{P}\big(\Phi(\mathcal{I}_i)>\Phi(\mathcal{I}_j)\big)-\mathbb{P}\big(\Phi(\mathcal{I}_i)<\Phi(\mathcal{I}_j)\big)\Big)\Big(\mathbb{P}\big(\widetilde{\Phi}(\mathcal{I}_i)>\widetilde{\Phi}(\mathcal{I}_j)\big)-\mathbb{P}\big(\widetilde{\Phi}(\mathcal{I}_i)<\widetilde{\Phi}(\mathcal{I}_j)\big)\Big)>0,
\end{align*}
where $\widetilde{\Phi} = \mathcal{A}_{\epsilon}(\Phi)$ denotes the random synthetic ranking output by Algorithm \ref{algorithm1} with $\Phi$ being input ranking.

\item[(2)] Effect on ranking distribution: For any $\mathcal{I}_i,\mathcal{I}_j\in \Omega$ with $\mathcal{I}_i \neq \mathcal{I}_j$, it holds true that
\begin{align*}
 \frac{\exp\big((|\Omega|-1)^{-1}\epsilon\big)-1}{\exp\big((|\Omega|-1)^{-1}\epsilon\big)+1} \leq 
\frac{|2\widetilde{\eta}_{ij}-1|}{|2\eta_{ij}-1|} \leq 
\frac{\exp\big(\frac{2|\Omega|-3}{|\Omega|-1}\epsilon\big)-1}{\exp\big(\frac{2|\Omega|-3}{|\Omega|-1}\epsilon\big)+1},
\end{align*}
where $\eta_{ij} =\mathbb{P}\big(\Phi(\mathcal{I}_i)>\Phi(\mathcal{I}_j)\big)$ and $\widetilde{\eta}_{ij}=\mathbb{P}\big(\widetilde{\Phi}(\mathcal{I}_i)>\widetilde{\Phi}(\mathcal{I}_j)\big)$.

\item[(3)] At the iteration $t$, the expectation and variance of $V^{(t)}_{\epsilon}$ are given as
\begin{align*}
&
\mathbb{E}\big(
V^{(t)}_{\epsilon}
\big) =
\frac{(t-1)q_{\epsilon}^t}{q_{\epsilon}^t-1}-\frac{q_{\epsilon}^t-q_{\epsilon}}{(q_{\epsilon}-1)(q_{\epsilon}^t-1)}\xrightarrow{\epsilon\rightarrow +\infty} t-1 \\
&
\text{Var}\big(
V^{(t)}_{\epsilon}\big) = 
\frac{(t-1)^2 q_{\epsilon}^t}{q_{\epsilon}^t-1} - 
\frac{2\mathbb{E}\big(
V^{(t)}_{\epsilon}
\big) }{q_{\epsilon}-1}+\frac{q_{\epsilon}^t-q_{\epsilon}}{(q_{\epsilon}^t-1)(q_{\epsilon}-1)} -
\Big( \mathbb{E}\big(
V^{(t)}_{\epsilon}
\big) \Big)^2  \xrightarrow{\epsilon\rightarrow +\infty} 0 ,
\end{align*}
where $q_{\epsilon} = \exp\big(\epsilon(|\Omega|-1)^{-1} \big)$.
\end{itemize}
\end{theorem}

In Theorem $\ref{Thm:ConsisandPriv}$, we present several properties of the proposed synthetic ranking algorithm. To some extend, the properties in Theorem $\ref{Thm:ConsisandPriv}$ establishes quantitative effect of the Mallows model on the distribution of ranking. Specifically, property (1) shows that the partial order between any two items' ranks stays invariant at the population level after being processed by Algorithm \ref{algorithm1}, which indicates that $\mathcal{I}_i$ is more likely to have a higher rank than $\mathcal{I}_j$ in a synthetic ranking if this relation also holds for $\Phi$. This property also implicitly permits the invariance of the true ranking based on items' quality values. Property (2) characterizes the effect of the privacy parameter $\epsilon$ on the distribution of ranking. As $\epsilon$ decreases to zero, $\widetilde{\eta}_{ij}$ gets closer to 1/2, which means that the partial order between $\mathcal{I}_i$ and $\mathcal{I}_j$ is indistinguishable. Property (3) characterizes the relation between $\epsilon$ and $V_{\epsilon}^{(t)}$, where $V_{\epsilon}^{(t)}$ indicates the number of correct partial relations retained in the synthetic ranking in the $t$-th iteration. As $\epsilon$ increases to infinity, the distribution of $V_{\epsilon}^{(t)}$ converges to the constant $t-1$ as expected. This is natural since $\epsilon=\infty$ refers to the non-private case as in the definition of differential privacy, and in this case the output ranking $\widetilde{\phi}$ is identical to the input ranking $\phi$.

\subsection{Comparison to Laplace Noise}
In this section, we demonstrate the effectiveness of the synthetic ranking algorithm in comparison with the Laplace mechanism \citep{dwork2006differential} under the developed $\epsilon$-ranking differential privacy. In the domain of differential privacy, the Laplace mechanism has emerged as a popular technique to ensure privacy protection for numeric data due to its computational simplicity. For releasing a ranking $\phi$ of an item set $\Omega = \{\mathcal{I}_1,\ldots,\mathcal{I}_m\}$ in a secure manner, the Laplace mechanism adds element-wise noises to ranks of items, which is given as
\begin{align*}
\mathcal{M}_{\lambda}^{lap}(\phi)  = \big(\phi(\mathcal{I}_i) + \xi_i\big)_{i \in [m]},
\end{align*}
where $\xi_i$ are i.i.d. random samples drawn from the Laplace distribution with mean $0$ and scale $\lambda$.

\begin{lemma}
\label{LAPDP}
Let $\phi$ and $\phi'$ be two neighboring rankings on the item set $\Omega=\{\mathcal{I}_1,\ldots,\mathcal{I}_m\}$. Given that $\lambda = 2(m-1)\epsilon^{-1}$, it holds that
\begin{align*}
\Big|
\log
\frac{\mathbb{P}(\mathcal{M}^{lap}_{\lambda}(\phi)=\bm{r})}{\mathbb{P}(\mathcal{M}^{lap}_{_{\lambda}}(\phi')=\bm{r})}
\Big| \leq \epsilon,
\end{align*}
 for any $\bm{r} \in \mathbb{R}^{m}$.
\end{lemma}

Lemma \ref{LAPDP} shows that adding Laplace noise to ranks of items also achieves the proposed $\epsilon$-ranking differential privacy with a properly chosen scale $\lambda = 2(m-1)\epsilon^{-1}$.

\begin{lemma}
\label{RDPvLAP}
Let $\mathcal{A}_{\epsilon}$ and $\mathcal{M}^{lap}_{2(m-1)\epsilon^{-1}}$ denote Algorithm \ref{algorithm1} and the Laplace mechanism satisfying $\epsilon$-ranking differential privacy, respectively. For any ranking $\phi$ of length $m$, it holds true that for any $\epsilon>0$
\begin{align*}
\mathbb{E}\Big[
T\big(\phi,\mathcal{A}_{\epsilon}(\phi)\big)\Big]
>
\mathbb{E}\Big[
T\big(\phi,\mathcal{M}^{lap}_{2(m-1)\epsilon^{-1}}(\phi)\big)\Big],
\end{align*}
where $T(\phi,\widetilde{\phi})= \sum_{\mathcal{I}_i,\mathcal{I}_j\in \Omega}I\big(\big(\phi(\mathcal{I}_i)-\phi(\mathcal{I}_j)\big)\big(\widetilde{\phi}(\mathcal{I}_i)-\widetilde{\phi}(\mathcal{I}_j)\big)>0\big)$.
\end{lemma}

For two rankings $\phi$ and $\widetilde{\phi}$, $T(\phi,\widetilde{\phi})$ calculates the number of concordant pairs between $\phi$ and $\widetilde{\phi}$. Clearly, $T(\phi,\widetilde{\phi})$ attains the maximum value when $\phi$ and $\widetilde{\phi}$ have identical partial orders between items. In Lemma \ref{RDPvLAP}, we use the expected number of concordant pairs to measure the utility of the proposed synthetic ranking algorithm and the Laplace mechanism satisfying the same $\epsilon$-ranking DP, showing that even though the Laplace mechanism can achieve the same privacy guarantee, it is is less effective than our algorithm in preserving relative positions of items in the raw ranking. This is because the framework of differential privacy imposes privacy guarantee on top of the worst case.

    \begin{figure}[h!]
        \centering
        \begin{subfigure}[b]{0.32\textwidth}  
            \centering 
            \includegraphics[width=\textwidth]{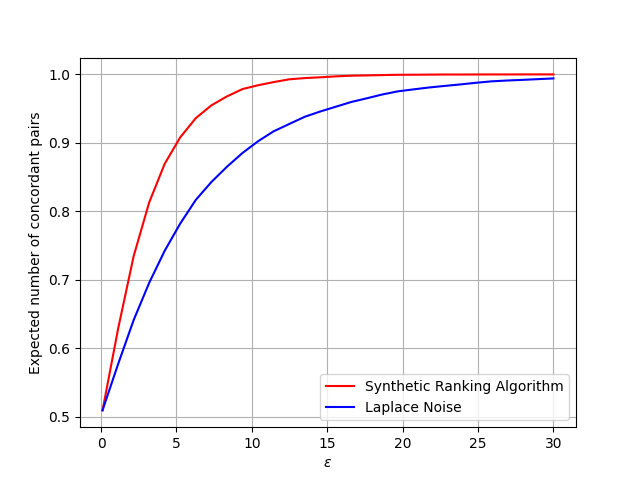}
            \caption[]%
            {{\small m=4}}    
        \end{subfigure}
        \begin{subfigure}[b]{0.32\textwidth}   
            \centering 
            \includegraphics[width=\textwidth]{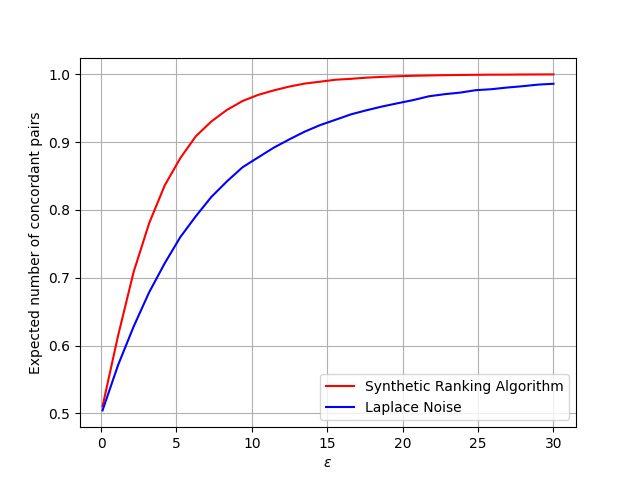}
            \caption[]%
            {{\small m=5}}    
        \end{subfigure}
        \begin{subfigure}[b]{0.32\textwidth}   
            \centering 
            \includegraphics[width=\textwidth]{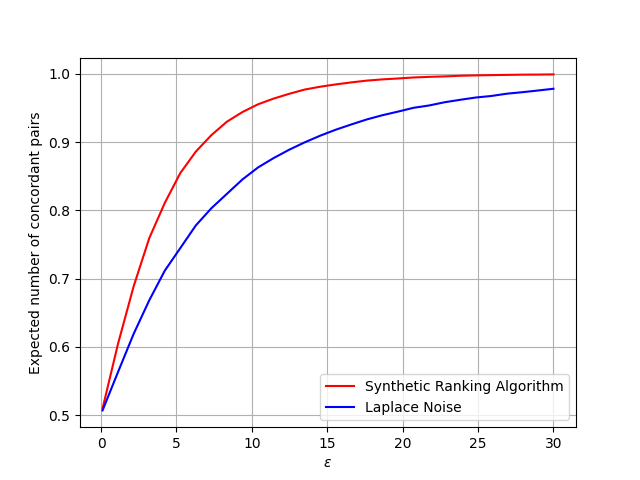}
            \caption[]%
            {{\small m=6}}    
        \end{subfigure}
        \caption{These figures present the averaged $T\big(\phi,\mathcal{A}_{\epsilon}(\phi)\big)$ (red) and $T\big(\phi,\mathcal{M}^{lap}_{2(m-1)\epsilon^{-1}}(\phi)\big)$ (blue) in 20,000 replications with $\epsilon \in [0.1,30]$, where $\phi$ is a ranking on the item set $\Omega=\{\mathcal{I}_1,\ldots,\mathcal{I}_m\}$ with $\phi(\mathcal{I}_i) = i,i=1,\ldots,m$.
        }
        \label{fig:LAPRDP}
    \end{figure}

To further validate theoretical results in Lemma \ref{RDPvLAP}, we carry out a simple experiment to illustrate the improvement of the proposed synthetic ranking algorithm relative to the Laplace mechanism. As can be seen in Figure \ref{fig:LAPRDP}, the proposed algorithm produces synthetic rankings preserving more partial orders among items in all cases. Particularly, the improvement becomes more significant as the size of ranking $m$ increases, showing that the proposed synthetic ranking algorithm is highly competitive when applied to a large ranking size.

\section{Inference Attack of Ranking}
\label{Sec:IF}
A critical challenge in privacy protection is against inference attacks \citep{williams2010probabilistic,sun2018truth,wu2022does}, which aims to recover some sensitive information contained in the true dataset based on the released data. For ranking data, obtaining rankings of items is usually of interest to attackers. For example, preference rankings revealing users' preferences can be used for advertisement purpose \citep{mayer2012third,chen2014economic}.

\begin{figure}[h!]
\centering
\includegraphics[scale=0.5]{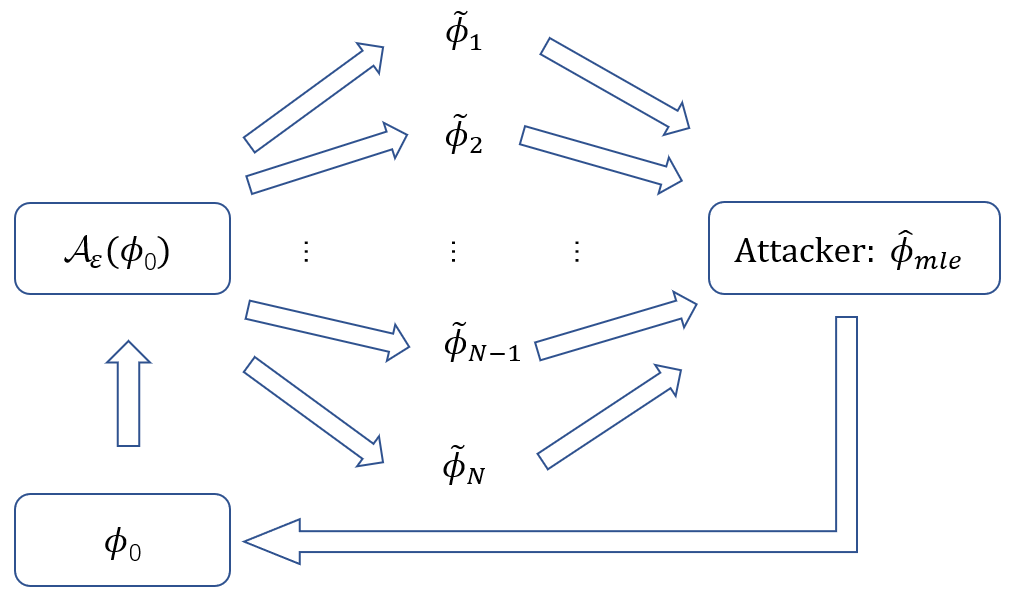}
\caption{The framework of inference attack of rankings.}
\label{Fig:Attack}
\end{figure}

In this section, we intend to study the effectiveness of the proposed synthetic ranking algorithm in resisting inference attacks based on synthetic rankings. Specifically, as illustrated in Figure \ref{Fig:Attack}, we assume that attackers can interact with the synthetic ranking algorithm frequently, requesting multiple synthetic rankings of the same central ranking $\phi_0$. In addition, attackers are assumed to have full knowledge of the Mallows model and privacy parameter $\epsilon$ and intend to estimate the observed ranking.

Let $\mathcal{S}=\{\widetilde{\phi}_i\}_{i=1}^N$ denote a set of synthetic rankings generated by Algorithm \ref{algorithm1} with the input ranking $\phi_0$ on the item set $\Omega$. We first consider a simple case that privacy parameter $\epsilon$ is known. Hence, it remains to estimate the observed ranking $\phi_0$ for attackers. The log-likelihood function of $\mathcal{S}$ can be written as
\begin{align*}
\mathcal{L}(\phi) = 
N \log \big\{\sum_{\widetilde{\phi} \in \Upsilon(|\Omega|)}\exp\big(2^{-1}\epsilon|\Omega| T(\phi,\widetilde{\phi} \big)\big) \big\} + \frac{\epsilon|\Omega|}{2}\sum_{i=1}^N T(\phi,\widetilde{\phi}_i).
\end{align*}
Let $\widehat{\phi}_{mle}$ denote the maximum likelihood estimation (MLE) of $\phi_0$, which is defined as
\begin{align}
\label{MLEE}
\widehat{\phi}_{mle} =\argmin_{\phi \in \Upsilon(|\Omega|)}\mathcal{L}(\phi).
\end{align}

The consistency of $\widehat{\phi}_{mle}$ to $\phi_0$ for any fixed $\epsilon$ is guaranteed by general results in \citet{choirat2012estimation}, which proves the consistency of $M$-estimators in discrete parameter models. In practice, the estimator $\widehat{\phi}_{mle}$ is difficult to obtain, since the feasible region of $\widehat{\phi}_{mle}$ is a discrete space. Therefore, the problem of finding $\widehat{\phi}_{mle}$ is known to be NP-hard \citep{meilua2010exponential,young1986optimal}. This explicitly demonstrates the merits of the proposed ranking algorithm in avoiding estimation of the observed ranking $\phi_0$ from a computational perspective. Specifically, it is computationally heavy for attackers to obtain the correct ranking via MLE, even though full knowledge of the synthetic ranking algorithm and the privacy parameter $\epsilon$ are provided.

\begin{theorem}
\label{Thm:ConsisPhi}
Let $\phi_0$ denote the observed ranking and $\mathcal{S}=\{\widetilde{\phi}_i\}_{i=1}^N$ denote a set of synthetic rankings generated by $\mathcal{A}_{\epsilon}(\phi_0)$. For any $\epsilon>0$ and $\Omega$, there exists a positive constant $C_0$ such that that
\begin{align*}
\mathbb{P}_{\mathcal{S}}\Big(
\widehat{\phi}_{mle} \neq \phi_0
\Big)
\geq 
C_0 
\frac{\exp(\epsilon(|\Omega|-1)^{-1})+1}{\exp(\epsilon(|\Omega|-1)^{-1})-1}
\sqrt{
\frac{2}{\pi N}}
\exp\Big\{
-N \Big(\exp(\epsilon(|\Omega|-1)^{-1})-1\Big)^2
\Big\}.
\end{align*}
Furthermore, $\mathbb{P}_{\mathcal{S}}\big(\widehat{\phi}_{mle} \neq \phi_0\big)$ is bounded away from 0 if $\epsilon = O \big( (|\Omega|-1)\sqrt{N^{-1}} \big)$.
\end{theorem}

Theorem $\ref{Thm:ConsisPhi}$ provides a lower bound for the estimation error of $\widehat{\phi}_{mle}$, quantifying the effect of privacy parameter $\epsilon$ on the convergence of $\widehat{\phi}_{mle}$. The implication of Theorem \ref{Thm:ConsisPhi} is two-fold. First, the lower bound enlarges when the length of ranking $|\Omega|$ increases, showing that the length of ranking amplifies the difficulty of the estimation of $\phi_0$. Additionally, setting $\epsilon$ to be adaptive to the number of synthetic rankings at the order $\epsilon = O\big( (|\Omega|-1)\sqrt{N^{-1}} \big)$ leads to the inconsistency of $\widehat{\phi}_{mle}$, where attackers estimate $\phi_0$ inaccurately with a fixed probability, regardless of the number of synthetic rankings requested.

\section{Ranking Differential Privacy in Personalized Ranking}
\label{Sec:PR}
This section is devoted to exploring the application of $\epsilon$-ranking DP to personalized ranking and establishing theoretical results regarding the quantitative relation between $\epsilon$-ranking DP and the consistency of the estimated ranking function based on synthetic rankings.

\subsection{Personalized Ranking}
A typical dataset in ranking problems consists of a set of triples $(\bm{x}_u,\bm{y}_i,\phi_u(\mathcal{I}_i))$, where $\bm{x}_u \in \mathbb{R}^p$ denotes the $p$-dimensional feature vector of user or query $u$, $\bm{y}_i \in \mathbb{R}^q$ denotes the $q$-dimensional feature vector of item $i$, and $\phi_u(\mathcal{I}_i)$ represents the preference or relevance rank from user $u$ to item $i$. We suppose that the item set is fixed, and the objective of personalized ranking problem \citep{balakrishnan2012collaborative} is to select most preferred or relevant items for a new user according to his/her preference or relevance.

To estimate the ranking function, it is a common practice to employ the pairwise ranking loss \citep{rendle2012bpr,balakrishnan2012collaborative}.
\begin{equation}
\label{RankLoss}
R_n(f) =\frac{1}{nm(m-1)} \sum_{u=1}^n  \sum_{i \neq j}
I\Big(\phi_u(\mathcal{I}_i)>\phi_u(\mathcal{I}_{j})\Big)
I\Big(
f(\bm{x}_u,\bm{y}_i) \leq f(\bm{x}_u,\bm{y}_{j})
\Big).
\end{equation}
As can be seen in $(\ref{RankLoss})$, error occurs when $f$ disagrees with two observed ranks in relative orders, and hence minimizing (\ref{RankLoss}) stimulates $f$ to be consistent with observed ranks in ranking as much as possible.

Let $\phi_u$ denote a realization of $\Phi_u$ and $\eta_{uij}=\mathbb{P}\big(\Phi_u(\mathcal{I}_i)>\Phi_u(\mathcal{I}_{j})\big)$ denote the probability that user $u$ gives higher rank on $\mathcal{I}_i$ than $\mathcal{I}_{j}$. The ranking risk can be written as
\begin{align*}
R(f) = \frac{1}{m(m-1)}
\sum_{i<j}
\mathbb{E}
\Big[
\eta_{uij}I\big(
f(\bm{x}_u,\bm{y}_i) \leq f(\bm{x}_u,\bm{y}_{j})
\big)
+
(1-\eta_{uij}) I\big(
f(\bm{x}_u,\bm{y}_i) \geq f(\bm{x}_u,\bm{y}_{j})
\big)
\Big],
\end{align*}
where the expectation is taken with respect to the randomness in users.

Let $f^*=\argmin_{f }R(f)$ be the optimal ranking function. Specifically, the behavior of $f^*$ on the item pair $(i,j)$ is characterized by the relative order of $\eta_{uij}$ and $\eta_{uji}$ as illustrated in Lemma $\ref{Bayes_ranker}$. It is important to note that $f^*$ is not unique due to translation invariance of optimality, and hence a biased estimator with correct ranking on preferences also leads to the optimal performance in practice. Specifically, $f^*+C$ for any constant $C$ is still an optimal ranking function.

\begin{lemma}
(\citealp{dai2021scalable})
\label{Bayes_ranker}
The optimal ranking function $ f^*$ satisfies that for any user $u$ and $i,j \in [m]$,
\begin{equation}
    \Big(f^*(\bm{x}_u,\bm{y}_{i})-f^*(\bm{x}_u,\bm{y}_{j})\Big)
\Big(\eta_{uij} - \eta_{uji} \Big)
 \geq 0,
\end{equation}
where the equality holds if and only if $\eta_{uij} = \eta_{uji}$ and the minimal risk is $$
R^* = \mathbb{E}\Big[\frac{1}{m(m-1)}\sum_{i<j}\min\big(
\eta_{uij} , \eta_{uji}
\big)\Big].
$$
\end{lemma}

Since the indicator function is not differentiable, minimizing $(\ref{RankLoss})$ is computationally intractable. A common approach is to employ a surrogate loss function as a substitution for the indicator function. Specifically, the empirical risk equipped with a surrogate loss can be written as
\begin{align}
\label{RankLoss_surro}
R_{n,\upsilon}(f) =\frac{1}{nm(m-1)} \sum_{u=1}^n   \sum_{i \neq j}
I\Big(\phi_u(\mathcal{I}_i)>\phi_u(\mathcal{I}_{j})\Big) \upsilon \Big(
f(\bm{x}_u,\bm{y}_{i}) - f(\bm{x}_u,\bm{y}_{j})
\Big),
\end{align}
where $\upsilon$ denote a surrogate loss function, such as logistic loss $\upsilon(x)=\text{log}(1+\exp(-x))$ \citep{rendle2012bpr,zhu2005kernel}, exponential loss $\upsilon(x)=\exp(-x)$ \citep{schapire2003boosting}, and $\psi$-loss $\upsilon(x)=\min(1,(1-x)_+ )$ \citep{shen2003psi,dai2021scalable}. It should be noted that $\upsilon$ cannot be hinge loss since the optimal ranking function under the hinge loss is not attainable in general \citep{gao2015consistency,dai2021scalable}.

\subsection{Differentially Private Personalized Ranking}
To protect the observed rankings of users $\{\phi_u\}_{u=1}^n$, the proposed synthetic ranking algorithm can be utilized to generate synthetic rankings $\{\widetilde{\phi}_u\}_{u=1}^n$ for the subsequent personalized learning. Let $\widetilde{\phi}_u = \mathcal{A}_{\epsilon_u}(\phi_u)$ denote the synthetic ranking generated by Algorithm \ref{algorithm1} satisfying the $\epsilon_u$-ranking DP for user $u$, where $\epsilon_u$ denotes the personalized privacy preference of user $u$. Then, the differentially private pair-wise ranking task can be organized as
\begin{align}
\label{DP_RankLoss_surro}
\widetilde{R}_{n,\upsilon}(f) =\frac{1}{nm(m-1)} \sum_{u=1}^n   \sum_{i \neq j}
I\Big(\widetilde{\phi}_u(\mathcal{I}_i)>\widetilde{\phi}_u(\mathcal{I}_{j})\Big) \upsilon \Big(
f(\bm{x}_u,\bm{y}_{i}) - f(\bm{x}_u,\bm{y}_{j})
\Big).
\end{align}
Let $\mathcal{F}$ denote the class of ranking functions of interest. The estimated ranking function $\widetilde{f}$ is obtained as
\begin{align}
\label{Object}
\widetilde{f} = \argmin_{f \in \mathcal{F}} 
\widetilde{R}_{n,\upsilon}(f)+ \lambda_n J(f),
\end{align}
where $\lambda_n$ is a tuning parameter vanishing with $n$ and $J(\cdot)$ is a regularization term.

Denote by $\widetilde{\Phi}_u = \mathcal{A}_{\epsilon_u}\big(\Phi_u)$ the random variable of synthetic ranking of user $u$. Let $\widetilde{\eta}_{uij}=\mathbb{P}\big(\widetilde{\Phi}_u(\mathcal{I}_i)>\widetilde{\Phi}_u(\mathcal{I}_{j})\big)$ denote the probability that the user $u$ gives a higher rank to $\mathcal{I}_i$ than $\mathcal{I}_{j}$ in the synthetic ranking $\widetilde{\phi}_u$. Correspondingly, the ranking risk of synthetic rankings can be written as
\begin{align*}
\widetilde{R}(f) = \frac{1}{m(m-1)}\sum_{i < j}\mathbb{E}
\Big[
\widetilde{\eta}_{uij}I\big(
f(\bm{x}_u,\bm{y}_i) \leq f(\bm{x}_u,\bm{y}_{j})
\big)
+
(1-\widetilde{\eta}_{uij}) I\big(
f(\bm{x}_u,\bm{y}_i) \geq f(\bm{x}_u,\bm{y}_{j})
\big)
\Big].
\end{align*}

Let $\widetilde{f}^*=\argmin_{f}\widetilde{R}(f)$ denote the optimal ranking function minimizing $\widetilde{R}(f)$ and $\widetilde{R}^* = \widetilde{R}(\widetilde{f}^*)$ denote the minimal synthetic ranking risk. 

\begin{lemma}
\label{Invariant_syn}
For any $\epsilon_u>0$, it holds that 
\begin{align*}
\big(f^*(\bm{x}_u,\bm{y}_i)-f^*(\bm{x}_u,\bm{y}_j)\big)
\big(\widetilde{f}^*(\bm{x}_u,\bm{y}_i)-\widetilde{f}^*(\bm{x}_u,\bm{y}_j)\big) >0,
\end{align*}
for any user $u$ and $i,j \in [m]$. Particularly, when $\epsilon_u=\epsilon$ for any $u$, it holds true that
\begin{align}
\label{Excess_Risk_rel}
\big(
R(f) -R^*\big)
\frac{\exp((|\Omega|-1)^{-1}\epsilon)-1}{\exp((|\Omega|-1)^{-1}\epsilon)+1}
\leq 
\widetilde{R}(f) - \widetilde{R}^* \leq 
\big(
R(f) - R^*\big)
\frac{\exp\big(\frac{2|\Omega|-3}{|\Omega|-1}\epsilon\big)-1}{\exp\big(\frac{2|\Omega|-3}{|\Omega|-1}\epsilon\big)+1}.
\end{align}
\end{lemma}

Lemma \ref{Invariant_syn} shows that $\widetilde{f}^*$ is consistent with $f^*$ in ranking items under any privacy guarantee $\epsilon$, demonstrating there is no bias in optimality of using synthetic rankings for the personalized learning. Furthermore, (\ref{Excess_Risk_rel}) establishes the quantitative relation between the excess risks under the raw ranking and the synthetic ranking distributions, from which we can derive the convergence rate of $R(\widetilde{f}) -R^*$ from that of $\widetilde{R}(\widetilde{f}) - \widetilde{R}^*$. By the upper bound in (\ref{Excess_Risk_rel}), we can see that $\widetilde{R}(f) - \widetilde{R}^*$ tends to 0 for any $f$ if $\epsilon$ goes to 0. This is as expected since $\epsilon=0$ refers to the most private case, where synthetic ranks of items are randomly determined, and all ranking functions degenerate simultaneously.

\subsection{Consistency in Differentially Private Personalized Ranking }
In this section, we establish theoretical results regarding to the asymptotic behavior of $\widetilde{f}$, quantifying how $\epsilon$-ranking DP affects the convergence rate of $R(\widetilde{f})-R^*$. We first define the ranking $\upsilon$-risk with respect to synthetic rankings as
\begin{align*}
\widetilde{R}_{\upsilon}(f) = \frac{1}{m(m-1)} \sum_{i \neq j}\mathbb{E}
\Big[
I\big(\widetilde{\Phi}_{u}(\mathcal{I}_i)-\widetilde{\Phi}_{u}(\mathcal{I}_i)>0\big)\upsilon\big(f(\bm{x}_u,\bm{y}_{i})-f(\bm{x}_u,\bm{y}_{j})\big)
\Big],
\end{align*}
where the expectation is taken with respect to the randomness from users and the synthetic ranking algorithm. We denote by $\widetilde{f}^*_{\upsilon}= \argmin_{f}\widetilde{R}_{\upsilon}(f)$ the optimal ranking funtion under $\upsilon$-loss. It is important to note that $\widetilde{f}^*$ is not unique, and $\widetilde{f}^*_{\upsilon}$ can also be the optimal ranking function in minimizing $\widetilde{R}(f)$ when $\upsilon(\cdot)$ is properly chosen, such as logistic loss \citep{gao2015consistency}, exponential loss \citep{gao2015consistency}, and $\psi$-loss \citep{dai2021scalable}.

In this paper, we only consider those loss functions such that $\widetilde{R}(\widetilde{f}^*_{\upsilon}) = \widetilde{R}(\widetilde{f}^*)$, and $\widetilde{f}^*$ will be referred to as $\widetilde{f}^*_{\upsilon}$ in the sequel. We denote $e(\widetilde{f} , \widetilde{f}^*) = \widetilde{R}(\widetilde{f}) - \widetilde{R}(\widetilde{f}^*)$ and $e_{\upsilon}(\widetilde{f} , \widetilde{f}^*) = \widetilde{R}_{\upsilon}(\widetilde{f})-\widetilde{R}_{\upsilon}(\widetilde{f}^*)$ as its excess risk under $0$-1 loss and $\upsilon$-loss, respectively. To quantify how $\epsilon$-ranking DP affects the excess risk, we first derive the convergence rate of $e(\widetilde{f} , \widetilde{f}^*)$ from that $e_{\upsilon}(\widetilde{f} , \widetilde{f}^*)$, which combined with (\ref{Excess_Risk_rel}) characterizes the asymptotic behavior of $R(\widetilde{f})-R^*$. The derived convergence behavior of $e(\widetilde{f},\widetilde{f}^*)$ from $e_{\upsilon}(\widetilde{f} , \widetilde{f}^*)$ is known to depend on the surrogate loss $\upsilon$ \citep{zhang2004statistical,gao2015consistency}. Before introducing the main theory, we first list some assumptions.

\begin{assumption}
\label{Assum_B}
Let $\mathcal{F}_{\delta} = \{f \in \mathcal{F}: e_{\upsilon}(f , \widetilde{f}^*) \leq \delta \}$ denote a subset of $\mathcal{F}$ such that the excess $\upsilon$-risk of any $f \in \mathcal{F}_{\delta}$ is smaller than $\delta$. Assume that there exist a constant $\alpha>0$ and a sufficiently small constant $\delta>0$ such that $\sup_{f  \in \mathcal{F}_{\delta}}e(f,\widetilde{f}^*) \leq \delta^{\alpha}$.
\end{assumption}

Assumption \ref{Assum_B} establishes a conversion relationship between $e_{\upsilon}(f , \widetilde{f}^*)$ and $e(f , \widetilde{f}^*)$, which influences the asymptotic behavior of $e(f , \widetilde{f}^*)$. Specifically, $\alpha$ depends on the choice of loss function, and particularly $\alpha=1$ for $\psi$-loss \citep{dai2021scalable} and $\alpha=1/2$ for exponential loss and logistic loss \citep{zhang2004statistical,bartlett2006convexity,gao2015consistency}. Additionally, Assumption \ref{Assum_B} indicates that the convergence rate of $e(\widetilde{f} , \widetilde{f}^*)$ can be derived from that of $e_{\upsilon}(\widetilde{f} , \widetilde{f}^*)$. Let $\widetilde{f}^*_{\mathcal{F}}= \argmin_{f \in \mathcal{F}}\widetilde{R}_{\upsilon}(f)$ be the best ranking function in $\mathcal{F}$ in approximating $\widetilde{f}^*$. The excess risk $e_{\upsilon}(\widetilde{f} , \widetilde{f}^*)$ admits the decomposition as
\begin{align*}
e_{\upsilon}(\widetilde{f} , \widetilde{f}^*) =  e_{\upsilon}(\widetilde{f} , \widetilde{f} _{\mathcal{F}}^*)+e(\widetilde{f} _{\mathcal{F}}^*, \widetilde{f}^*) = \widetilde{R}_{\upsilon}(\widetilde{f}) - \widetilde{R}_{\upsilon}(\widetilde{f} _{\mathcal{F}}^*)+
\widetilde{R}_{\upsilon}(\widetilde{f} _{\mathcal{F}}^*) - \widetilde{R}_{\upsilon}(\widetilde{f}^*),
\end{align*}
where $e_{\upsilon}(\widetilde{f} , \widetilde{f}_{\mathcal{F}}^*)$ and $e_{\upsilon}(\widetilde{f} _{\mathcal{F}}^*,\widetilde{f}^*)$ are usually referred to as estimation error and approximation error, respectively. 

In this paper, we mainly focus on the estimation error in the consistency of $\widetilde{f}$. Therefore, in Assumption \ref{APPROX}, we assume that the class of ranking functions is correctly specified and the approximation error is ignorable.
\begin{assumption}
\label{APPROX}
The class of ranking functions $\mathcal{F}$ is properly chosen such that $e_{\upsilon}(\widetilde{f} _{\mathcal{F}}^*,\widetilde{f}^*) =0$.
\end{assumption}

\begin{assumption}
\label{Assum:privacy}
We assume that each user $u$ has personal privacy preference $\epsilon_u$ and $\Psi_u =  \frac{\exp\big((|\Omega|-1)^{-1}\epsilon_u\big)+1}{\exp\big((|\Omega|-1)^{-1}\epsilon_u\big)-1}$ is a sub-Gaussian random variable. 
\end{assumption}

Assumption \ref{Assum:privacy} considers varying privacy preference among users, which generally holds true in real-life applications \citep{watson2015mapping}. Additionally, Assumption \ref{Assum:privacy} imposes a constraint on the distribution of users' privacy preference that $\Psi_u$ is sub-Gaussian distribution, which implies that fewer users pursue stronger privacy protection on their rankings.

\begin{assumption}
\label{Assum:low-noise}
For any user $u$, we assume that there exists some constants $C_2>0$ and $0<\gamma \leq +\infty$ such that $\mathbb{P}\big(|2\eta_{uij}-1| \leq \beta \big) \leq C_2 \beta^{\gamma}$ for $0 \leq \beta \leq 1$.
\end{assumption}

Assumption \ref{Assum:low-noise} is known as the low-noise assumption \citep{bartlett2006convexity,shen2003psi} characterizing the behavior of $2\eta_{uij}-1$ around the decision boundary, which affects the asymptotic behavior of $\widetilde{f}$. Particularly, $\gamma=+\infty$ implies that the partial orders of preferences are deterministic, resulting in optimal convergence rate of $e(\widetilde{f},\widetilde{f}^*)$.

\begin{lemma}
\label{lemma:low-noise}
Under Assumption \ref{Assum:low-noise}, it holds that for any user $u$,
$
\mathbb{P}\big(
|2\widetilde{\eta}_{uij}-1| \leq \beta
\big) \leq C_2 \Psi_{u}^{\gamma}\beta^{\gamma}
$, where $\Psi_u =  \frac{\exp\big((|\Omega|-1)^{-1}\epsilon_u\big)+1}{\exp\big((|\Omega|-1)^{-1}\epsilon_u\big)-1}$.
\end{lemma}

\begin{assumption}
\label{Assum_3}
Let $\Theta$ denote all parameters of functions in $\mathcal{F}$. We assume that $\Vert \Theta \Vert_{\infty} \leq C_{\mathcal{F}}$ and there exists some positive constants $C_1$ such that for any $f_1,f_2 \in \mathcal{F}$
\begin{align*}
\big(f_1(\bm{x}_u,\bm{y}_i) - f_2(\bm{x}_u,\bm{y}_i)\big)^2 \leq C_1 \big( \Vert \bm{x}_u \Vert_2^2 + \Vert \bm{y}_i \Vert_2^2\big)
\Vert \Theta_{f_1} -\Theta_{f_2} \Vert_{\infty}^2,
\end{align*}
where $\Theta_f$ denote the parameters of $f$.
\end{assumption}

Assumption \ref{Assum_3} is a mild assumption, ensuring the smoothness of functions in $\mathcal{F}$ with respect to input features and parameters, establishing the connection between the metric entropy of $\mathcal{F}$ and the associated parameter space.

\begin{theorem}
\label{Thm: Main}
Under Assumptions \ref{Assum_B}-\ref{Assum_3}, it holds that for any minimizer $\widetilde{f}$ defined in $(\ref{Object})$, there exist some constants $C_3,C_4>0$ such that
\begin{align*}
\mathbb{P}\big(    e(\widetilde{f},f^*) \geq \delta_n^{\alpha}      \big) \leq 
8C_3  \exp\Big(
-C_4n\mathbb{E}^{-\frac{1}{1+\gamma}}(\Psi_{u}^{\gamma}) \delta_{n}^{\frac{\gamma+2}{\gamma+1}}
\Big),
\end{align*}
where $\mathbb{E}(\Psi_{u}^{\gamma})=O(\delta_n)$, $\mathbb{E}^{\frac{1}{\gamma+2}}(\Psi_{u}^{\gamma})  \big(|\Theta| n^{-1}\log(n/|\Theta|)\big)^{\frac{\gamma+1}{\gamma+2}} =O(\delta_n)$, and $|\Theta|$ is the number of parameters of $f \in \mathcal{F}$ and $\lambda_n J_0 \asymp \delta_n $ with $J_0 =\max\{ J(f^*_{\mathcal{F}}),1\}$.
\end{theorem}

Theorem $\ref{Thm: Main}$ quantifies the asymptotic behavior of $e(\widetilde{f},f^*)$, as well as its relation to the privacy guarantee. Specifically, the convergence rate $\delta_n^{\alpha}$ is governed by $\gamma$, $\alpha$, averaged privacy preference $\mathbb{E}(\Psi_{u}^{\gamma})$, and the complexity of $\mathcal{F}$. Particularly, when $\alpha=1$ and $\gamma=\epsilon=\infty$, the best convergence rate is obtained as $O_p\Big(|\Theta| n^{-1}\log(n/|\Theta|)\Big)$, which matches the existing theoretical results in \citet{dai2021scalable}. Furthermore, it is interesting to note that, when $\mathbb{E}^{-1}(\Psi_{u}^{\gamma}) = o(1)$ , the rate $\delta_n$ gets slower compared with that of non-private case. For example, when $\epsilon_u=\epsilon$ for all users, $\mathbb{E}^{-1}(\Psi_{u}^{\gamma})$ is of the order $\epsilon^{\gamma}$ with $\epsilon=o(1)$, indicating that the convergence to the optimal ranking function is slower when all users pursue more protection of their rankings corresponding to a smaller $\epsilon$.

\begin{corollary}
\label{Coro_1}
Under assumptions of Theorem $\ref{Thm: Main}$, we have $$R(\widetilde{f}) - R(f^*) = O_p\Big(
\mathbb{E}(\Psi_{u}) 
\mathbb{E}^{\frac{\alpha}{\gamma+2}}(\Psi_{u}^{\gamma})  \big(|\Theta| n^{-1}\log(n/|\Theta|)\big)^{\frac{\alpha(\gamma+1)}{\gamma+2}}\Big).$$
Particularly, when $\epsilon_u=\epsilon=o(1)$ for any user $u$ and $\upsilon(\cdot)$ is chosen such that $\alpha=1$, we have
\begin{align*}
R(\widetilde{f}) - R(f^*)  = 
O_p\Big(
\big(|\Theta|(|\Omega|-1)^2n^{-1}\epsilon^{-2}\log(n/|\Theta|)\big)^{\frac{\gamma+1}{\gamma+2}}\Big).
\end{align*}
Furthermore, we have $R(\widetilde{f}) - R(f^*) = o_p(1)$ if $|\Omega|\sqrt{|\Theta| n^{-1} \log^{1+\zeta}(n/|\Theta|)}=O(\epsilon)$ for some $\zeta >0$.

\end{corollary}

Corollary $\ref{Coro_1}$ presents the convergence rate of excess risk of differentially private ranking function $\widetilde{f}$, quantifying the effect of synthetic ranking algorithm. First, if all users choose a constant privacy, the convergence rate becomes slower by a multiplicative constant compared with the non-private case. Corollary $\ref{Coro_1}$ also sheds light on the best rate of privacy guarantee achievable for all users under the consistency of pairwise learning. Specifically, when the privacy guarantee $\epsilon$ of all users satisfies $|\Omega|\sqrt{|\Theta| n^{-1} \log^{1+\zeta}(n/|\Theta|)}=O(\epsilon)$ for some $\zeta >0$ for any $\zeta>0$, pairwise learning can still estimate the optimal ranking function $f^*$ well with appropriately chosen parameter space $\mathcal{F}$ and loss function $\upsilon(\cdot)$.

\section{Experiment}
\label{Sec:Expe}
In this section, we conduct a series of numerical experiments on simulated datasets to validate our theoretical results and demonstrate the effectiveness of the proposed algorithm. 

\subsection{Privacy Guarantee}
\begin{figure}[htb]
        \centering
        \begin{subfigure}[b]{0.32\textwidth}  
            \centering 
            \includegraphics[width=\textwidth]{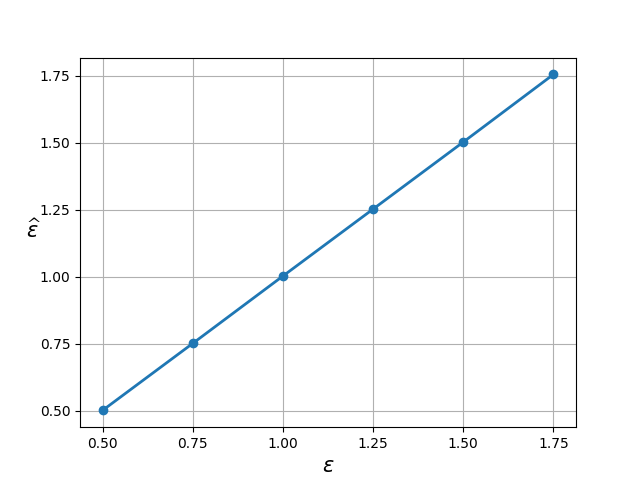}
            \caption[]%
            {{\small m=3}}    
        \end{subfigure}
        \begin{subfigure}[b]{0.32\textwidth}   
            \centering 
            \includegraphics[width=\textwidth]{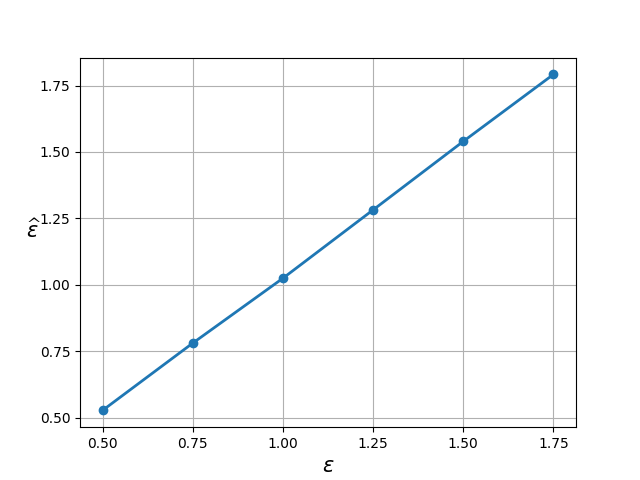}
            \caption[]%
            {{\small m=4}}    
        \end{subfigure}
        \begin{subfigure}[b]{0.32\textwidth}   
            \centering 
            \includegraphics[width=\textwidth]{Fig/PG_M5.png}
            \caption[]%
            {{\small m=5}}    
        \end{subfigure}
        \caption{Estimated privacy guarantee $\widehat{\epsilon}$ v.s. the pre-specified privact guarantee}
        \label{fig:PG}
    \end{figure}

In the first simulation, we aim to provide empirical validations of the privacy guarantee of the proposed synthetic ranking algorithm as stated in Lemma \ref{lemma:eDP}. To this end, we consider a ranking $\sigma$ with size $m$ such that $\phi(\mathcal{I}_i) = i$ for $i=1,\ldots,m$. Let $\mathcal{S}$ denote the set of all possible neighboring rankings of $\phi$ and $\mathcal{C}$ denote the set of all possible permutations of $\phi$. For each $\phi' \in \mathcal{S}$, we implement the proposed algorithm with privacy guarantee $\epsilon$ on $\phi$ and $\phi'$ in $N$ times and then estimate the privacy guarantee by
\begin{align}
\label{Mea:Ratio}
\hat{\epsilon} = 
\max_{\phi'  \in \mathcal{S}}\max_{\widetilde{\phi}\in \mathcal{C}} \Big|\log \Big(
\frac{\sum_{i=1}^{N}I(\mathcal{A}_{\epsilon}(\phi)= \widetilde{\phi})}{\sum_{i=1}^{N}I(\mathcal{A}_{\epsilon}(\phi')= \widetilde{\phi})} \Big)\Big|,
\end{align}
where $\mathcal{A}_{\epsilon}(\phi)$ denote the output ranking of the algorithm in the $i$-th replication. We consider cases that $(m,\epsilon) \in \{ 3,4,5\} \times \{ 0.5 +0.25 * i,i=0,\ldots,6\}$ with $N=10^7$ and the results are reported in Figure \ref{fig:PG}.

As can be seen in Figure $\ref{fig:PG}$, the estimated privacy guarantee $\hat{\epsilon}$ perfectly matches the pre-specified privacy guarantee $\epsilon$, which is consistent with our theoretical results established in Lemma $\ref{lemma:eDP}$.

\subsection{Inference Attack}
This simulation intends to verify theoretical results in Theorem \ref{Thm:ConsisPhi} that an adaptive scheme of $\epsilon$ regarding the number of synthetic rankings results in the deterioration of the estimation of ranking. To this end, the simulation setting is organized as follows. First, for a ranking $\phi_0$ of size $m$, we generate a set of synthetic rankings $\mathcal{S}_m = \{\widetilde{\phi}_{i}\}_{i=1}^N$. Second, the estimation of $\phi_0$ is implemented via maximum likelihood estimation based on $\mathcal{S}_m$ as in (\ref{MLEE}). We repeat the above steps in $R$ replications. Let $\widehat{\phi}_{mle}^{(i)}$ denote the maximum likelihood estimator in the $i$-th replication. Then we estimate the probability of the inconsistency of MLE as
\begin{align*}
\widehat{\mathbb{P}}(\widehat{\phi}_{mle} \neq \phi_0)
=
\frac{1}{R}\sum_{i=1}^{R}
I(\widehat{\phi}_{mle}^{(i)} \neq \phi_0).
\end{align*}

In the first example, we aim to verify that $\widehat{\phi}_{mle}$ converges to $\phi_0$ in probability under a fixed privacy guarantee. Specifically, we consider rankings of size $m=3,4,5$ with $\phi_0(\mathcal{I}_i)=i$ for $i=1,\ldots,m$ and $R=1,000$. The numbers of synthetic rankings $N$ are set as $\{10,20,30,\ldots,100\}$. The privacy guarantee is set as $\epsilon \in \{1,2,3,4\}$.

    \begin{figure}[h!]
        \centering
        \begin{subfigure}[b]{0.475\textwidth}
            \centering
            \includegraphics[width=\textwidth]{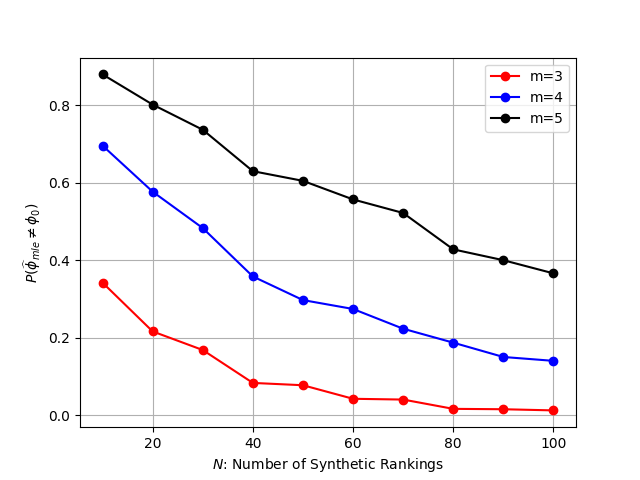}
            \caption[]%
            {{\small $\epsilon=1$}}    
        \end{subfigure}
        \begin{subfigure}[b]{0.475\textwidth}  
            \centering 
            \includegraphics[width=\textwidth]{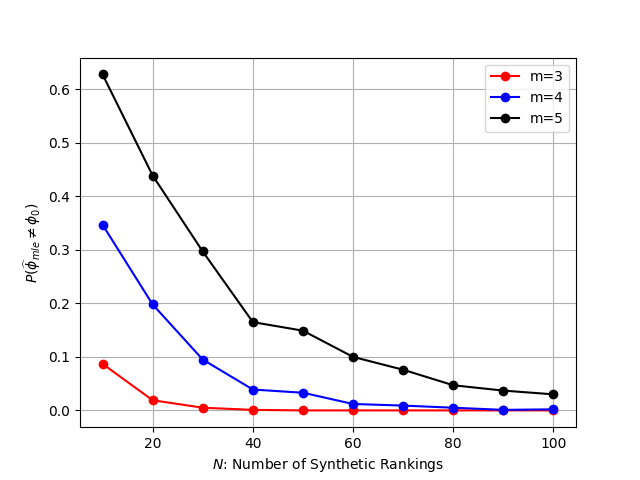}
            \caption[]%
            {{\small $\epsilon=2$}}    
        \end{subfigure}
        \begin{subfigure}[b]{0.475\textwidth}   
            \centering 
            \includegraphics[width=\textwidth]{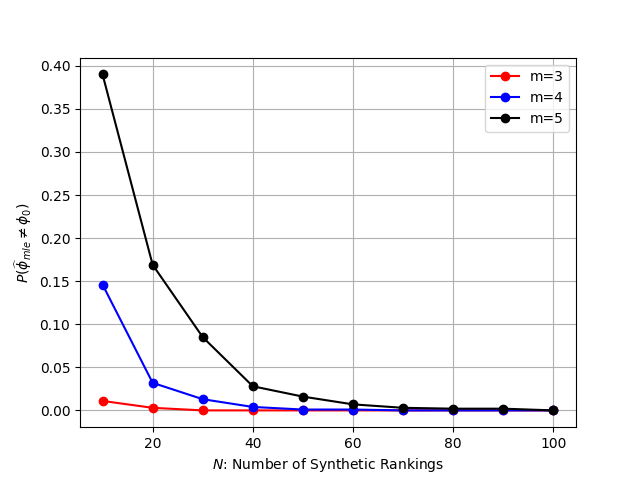}
            \caption[]%
            {{\small $\epsilon=3$}}    
        \end{subfigure}
        \begin{subfigure}[b]{0.475\textwidth}   
            \centering 
            \includegraphics[width=\textwidth]{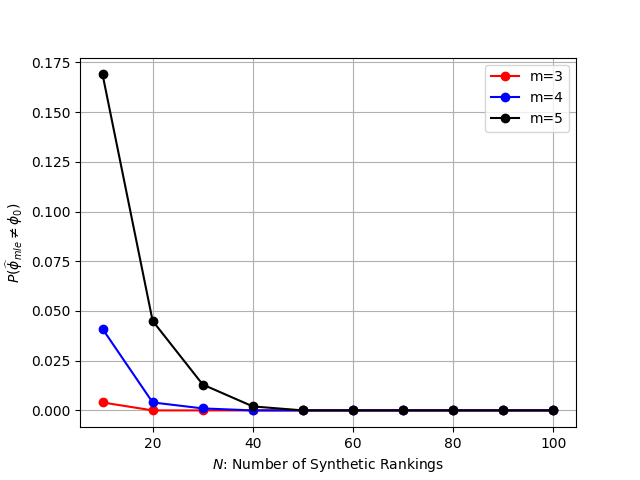}
            \caption[]%
            {{\small $\epsilon=4$}}    
        \end{subfigure}
        \caption{The empirical probabilities of the event $\widehat{\phi}_{mle}=\phi_0$ in 1,000 replications with different privacy guarantees $\epsilon=1,2,3,4$ and lengths of rankings $m=3,4,5$.}
        \label{fig:simu5-2}
    \end{figure}

The empirical probabilities of incorrect estimation of $\widehat{\phi}_{mle}$ under different privacy and ranking settings are plotted in Figure \ref{fig:simu5-2}. Clearly, $\widehat{\phi}_{mle}$ converges to the true ranking $\phi_0$ in probability as the number of synthetic rankings increases, illustrating that the true ranking used to generated synthetic rankings can be accurately estimated as long as attackers gain enough synthetic rankings. Furthermore, it is not surprising to observe that longer ranking length deteriorates the estimation of true ranking. This is because the developed $\epsilon$-ranking differential privacy enforces the protection of each rank, which enhances the difficulty of the estimation procedure as the length of ranking increases.

In the second example, we aim to verify that the convergence of $\widehat{\phi}_{mle}$ to $\phi_0$ can be prevented under an adaptive privacy guarantee, and the optimal adaptive privacy guarantee scheme is $\epsilon \asymp (m-1)/\sqrt{N}$. To this end, we follow the same simulation settings as in the first example, except that $\epsilon$ is set to be adaptive to the number of synthetic rankings as $\epsilon \asymp (m-1)/\sqrt{N}$ and $\epsilon \asymp (m-1)\log(N)/\sqrt{N}$ and the number of replications for estimating $\mathbb{P}(\widehat{\phi}_{mle} \neq \phi_0)$ is set as 5,000.

    \begin{figure}[htb]
        \centering
        \begin{subfigure}[b]{0.475\textwidth}
            \centering
            \includegraphics[width=\textwidth]{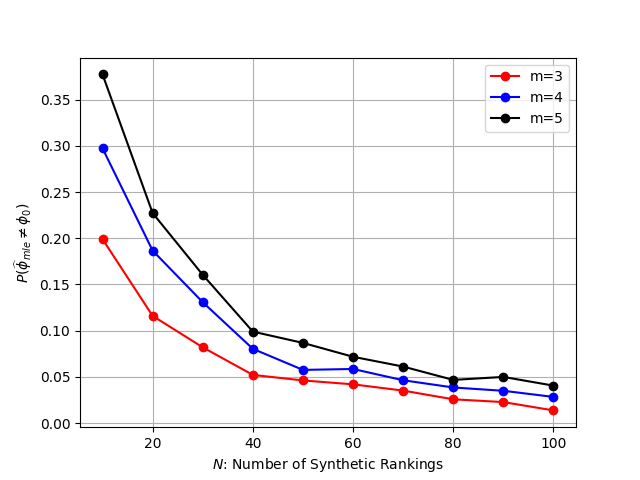}
            \caption[]%
            {{\small $\epsilon = \frac{\log(N)(m-1)}{\sqrt{N}} $}}    
        \end{subfigure}
        \begin{subfigure}[b]{0.475\textwidth}  
            \centering 
            \includegraphics[width=\textwidth]{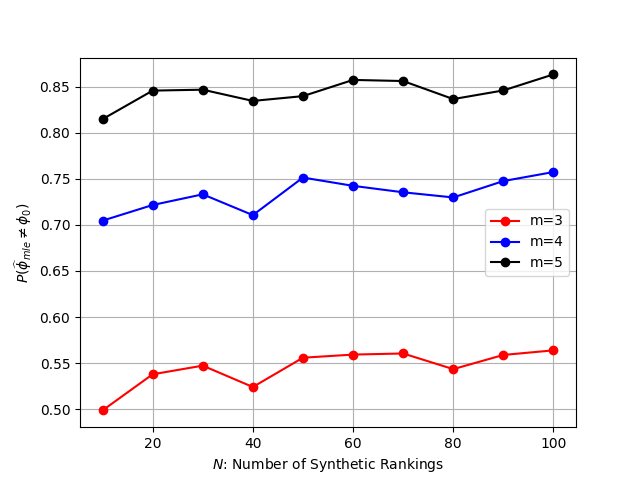}
            \caption[]%
            {{\small $\epsilon = \frac{m-1}{\sqrt{N}} $}}    
        \end{subfigure}
        \caption{The empirical probabilities of the event $\widehat{\phi}_{mle}=\phi_0$ in 5,000 replications with an adaptive $\epsilon$ and different lengths of rankings $m=3,4,5$.}
        \label{fig:SimuAda}
    \end{figure}

Figure \ref{fig:SimuAda} validates our theoretical results in Theorem \ref{Thm:ConsisPhi}, showing that the convergence of $\widehat{\mathbb{P}}(\widehat{\phi}_{mle} \neq \phi_0)$ is still achievable when $\epsilon$ decreases with the number of synthetic rankings at the rate $\epsilon \asymp (m-1)\log(N)/\sqrt{N}$. By contrast, without the logarithmic term, the consistency of $\widehat{\phi}_{mle}$ is invalidated and the empirical probabilities are bounded away from some fixed constants as stated in Theorem \ref{Thm:ConsisPhi}. These two cases show that the slowest rate of $\epsilon$ for reconciling the inconsistency of $\widehat{\phi}_{mle}$ and the utility of synthetic rankings is $\epsilon \asymp (m-1)/\sqrt{N}$.

\subsection{Differentially Private Personalized Ranking}
\label{Sim:DPPL}
In this simulation, we aim to verify the advantage of the proposed synthetic ranking algorithm in preserving more information in pairwise comparisons of rankings in downstream ranking learning task in comparison to the Laplace mechanism. First, we generated features of users and items by a $4$-dimensional uniform distribution, that is $x_{ul},y_{il}\sim \text{Unif}(-3,3)$ for $l=1,\ldots,4$. Then the preferences of user $u$ to items are generated by the model $r_{ui} = \bm{\alpha}^T \bm{x}_u +\bm{\beta}^T \bm{y}_i$, where the item set is fixed at $\Omega=\{\mathcal{I}_1,\ldots,\mathcal{I}_m\}$. The ranking of user $u$ is generated by sorting the associated preferences $\{r_{ui}\}_{i=1}^m$. Subsequently, we generate synthetic rankings via the synthetic ranking algorithm and the Laplace mechanism satisfying $\epsilon$-ranking differential privacy, which are denoted as $\widetilde{\phi}=\mathcal{A}_{\epsilon}(\phi_u)$ and $\widetilde{\phi}=\mathcal{M}_{2(m-1)\epsilon^{-1}}(\phi_u)$, respectively. Finally, we estimate the ranking function by minimizing the following regularized pairwise learning task.
\begin{align}
&\widetilde{f} =\argmin_{f \in \mathcal{F}} \frac{1}{n}\sum_{i=1}^n \sum_{i\neq j}I(\widetilde{\phi}_u\big(\mathcal{I}_i)>\widetilde{\phi}_u(\mathcal{I}_j)\big)
\upsilon(f(\bm{x}_u,\bm{y}_i)-f(\bm{x}_u,\bm{y}_j)),\label{Ftilde}
 \\
&\widetilde{f}_{lap} = 
\argmin_{f \in \mathcal{F}}
\frac{1}{n}\sum_{i=1}^n \sum_{i\neq j}I(\widetilde{\phi}_u^{lap}\big(\mathcal{I}_i)>\widetilde{\phi}_u^{lap}(\mathcal{I}_j)\big)
\upsilon(f(\bm{x}_u,\bm{y}_i)-f(\bm{x}_u,\bm{y}_j)),\label{Flaptilde}
\end{align}
where $\mathcal{F}$ is the class of two-tower models \citep{wang2021cross,yang2020mixed}, which employs two parallel neural networks to learn representations of users' and items' features and computes preference scores by dot product of their representations.

Furthermore, to compare the performance of $\widetilde{f}_{lap}$ and $\widetilde{f}$, we generate a set of new users with their features following same setting as above. Specifically, we let $\{\bm{x}_u'\}_{u=1}^{N}$ denote the set of features of new users and generate their preferences as $r_{ui}'=\bm{\alpha}^T \bm{x}_{u}' +\bm{\beta}^T \bm{y}_i$. For a ranking function $f$, we employ the following metric to evaluate its performance,
\begin{align}
\label{PRIS_test}
L_{pair}(f) = 
\frac{1}{Nm(m-1)}\sum_{u =1}^N \sum_{i \neq j}I\big(r_{ui}'>r_{uj}'\big)I\big(f(\bm{x}_{u}',i) > f(\bm{x}_{u}',j)\big). 
\end{align}

\begin{figure}[h!]
        \centering
        \begin{subfigure}[b]{0.475\textwidth}
            \centering
            \includegraphics[width=\textwidth]{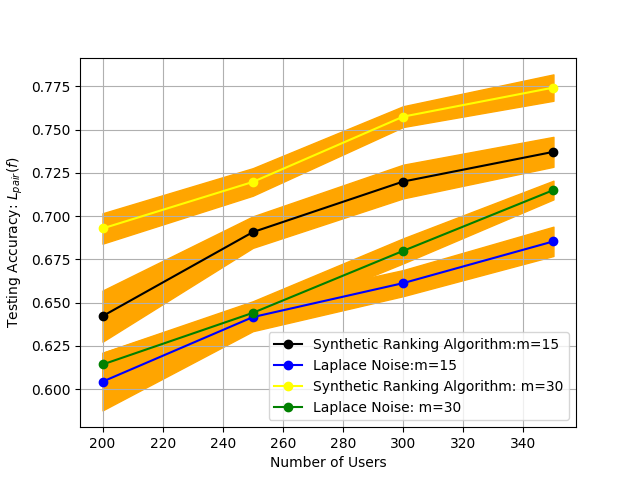}
            \caption[]%
            {{\small $\epsilon=1$}}    
        \end{subfigure}
        \begin{subfigure}[b]{0.475\textwidth}  
            \centering 
            \includegraphics[width=\textwidth]{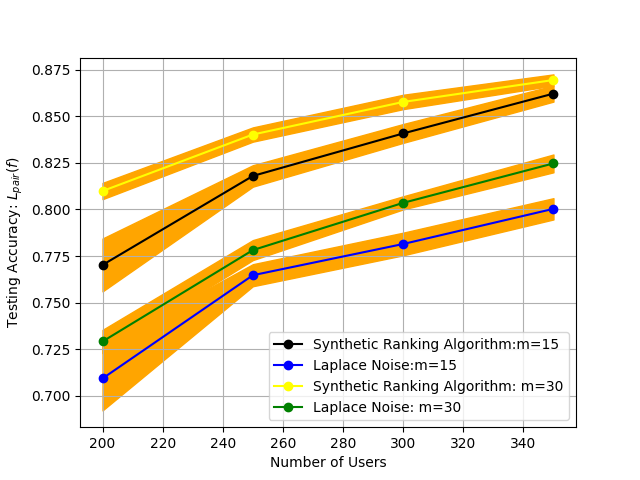}
            \caption[]%
            {{\small $\epsilon=2$}}    
        \end{subfigure}
                \begin{subfigure}[b]{0.475\textwidth}  
            \centering 
            \includegraphics[width=\textwidth]{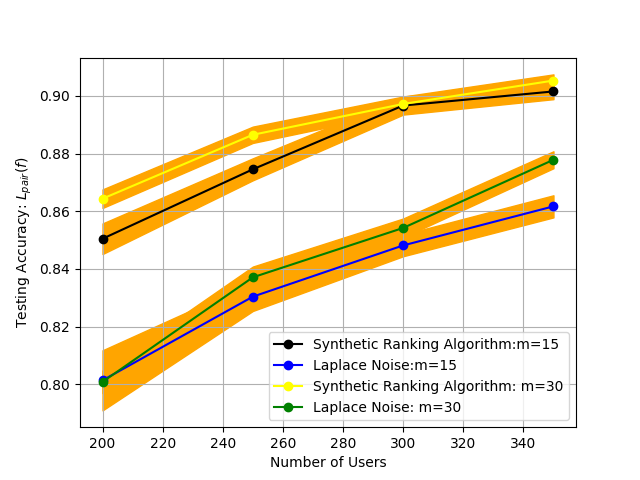}
            \caption[]%
            {{\small $\epsilon=3$}}    
        \end{subfigure}
                \begin{subfigure}[b]{0.475\textwidth}  
            \centering 
            \includegraphics[width=\textwidth]{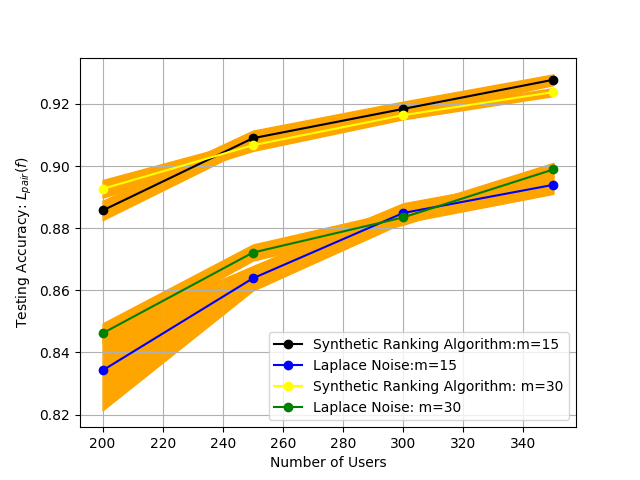}
            \caption[]%
            {{\small $\epsilon=4$}}    
        \end{subfigure}
        \caption{The averaged pairwise accuracies in 50 replications with different privacy guarantees $\epsilon=1,2,3,4$ and lengths of rankings $m=15,30$.}
        \label{fig:simu5_3}
    \end{figure}

For the hyper-parameter selection, we set two neural networks in the two-tower model to be a $3$-layer multilayer perceptron with 10 hidden units in each layer. We set $N=1,000$ and consider cases that $(m,n,\epsilon) \in \{15,30\} \times \{200,250,300,350\} \times \{1,2,3,4 \}$. In each case, 100 users will be used for validation and the early-stopping method monitoring the validation error is employed to tune the parameters of neural networks. We repeat each case in 50 runs and report the averaged pairwise testing accuracies in Figure \ref{fig:simu5_3}.

As seen in Figure \ref{fig:simu5_3}, the averaged testing accuracies of the proposed method are significantly higher than those of the Laplace noise in all scenarios, showing that the proposed synthetic ranking algorithm outperforms the Laplace noise in generating more utility-preserving rankings for the pairwise learning task.

\subsection{Real Application}
In this section, we evaluate the performance of the developed synthetic ranking algorithm over the Sushi dataset, aiming to show that our synthetic ranking algorithm maintains more information for downstream learning tasks in achieving a better generalization performance compared with adding Laplace noise under the same privacy guarantee. 

The Sushi dataset is publicly available at \url{https://www.kamishima.net/sushi/}, which consists of preference rankings of 5,000 users over 10 kinds of sushis collected via a questionnaire survey, including ``shrimp", ``sea eel", ``tuna", ``squid", ``sea urchin", ``salmon roe", ``tamago", ``fatty tuna", ``tuna roll", and ``cucumber roll". For this application, we only consider the preference ranks of  ``tuna", ``salmon roe", ``tamago", ``fatty tuna", and ``cucumber roll", whose preference ranks are more consistent among users compared with other sushis. 

We compare the utility of rankings generated by the synthetic ranking algorithm and the counterpart which adds Laplace noise via the pairwise learning task as in Section \ref{Sim:DPPL}. To be more specific, we first split the dataset into a training dataset of 3,000 users and a testing dataset of 2,000 users. Second, the rankings in the training dataset are permuted by the synthetic ranking algorithm and the Laplace noise satisfying the same $\epsilon$-ranking DP. Then, we estimate the ranking functions $\widetilde{f}$ and $\widetilde{f}_{lap}$ as in (\ref{Ftilde}) and (\ref{Flaptilde}), respectively, where the ranking function to be $f(\bm{x}_u,\bm{y}_i)=\bm{\alpha}\bm{x}_u+\bm{\beta}\bm{y}_i$. We evaluate the performance of $\widetilde{f}$ and $\widetilde{f}_{lap}$ by the pairwise test accuracy defined in (\ref{PRIS_test}) on the users in the testing dataset.

\begin{figure}[h]
\centering
\includegraphics[scale=0.75]{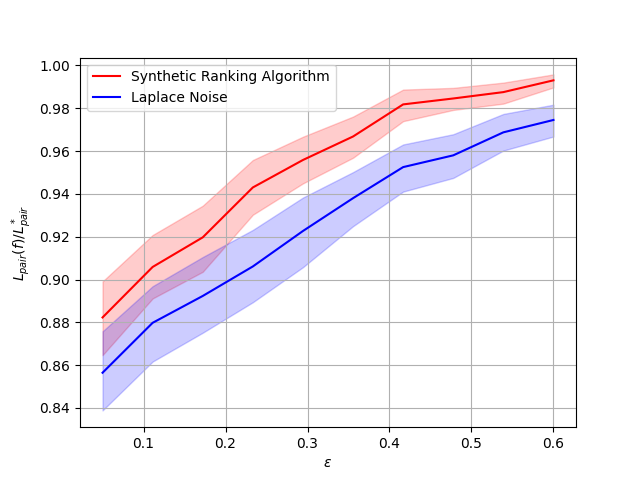}
\caption{The averaged relative utility in 50 replications with different privacy guarantees $\epsilon$ being evenly spaced values ranging from 0.06 to 0.5. $L_{pair}^*$ denotes the averaged pairwise accuracy of the ranking function trained on non-privatized training dataset in 50 replications.}
\label{Fig:Real}
\end{figure}

Figure \ref{Fig:Real} presents similar results as in Section \ref{Sim:DPPL} that the proposed method produces more utility-preserving synthetic rankings for the pairwise learning task compared with the Laplace noise under the same privacy guarantee.

\section{Summary}
\label{Sec:Summ}
In this paper, we propose a new privacy metric called $\epsilon$-ranking differential privacy for ranking data. The basic idea is to generate a synthetic ranking in placement of the real ranking for sharing while resisting inference on a specific rank within the ranking list. Based on this notion, we further develop a synthetic ranking algorithm grounded in the multistage ranking model to synthetic rankings satisfying the developed $\epsilon$-ranking differential privacy, which inherently permutes ordinal ranks of items in a probabilistic manner. Furthermore, we analyze the utility of synthetic rankings yielded by the proposed algorithm in the downstream inference attack and personalized ranking tasks, for which we establish statistical consistency characterizing the corresponding utility-privacy tradeoff. It is worth noting that, unlike existing methods employing noise addition approaches to produce differentially private rankings, the proposed method implicitly perturbs rankings nonlinearly while accommodating the requirement of strict privacy protection.

\renewcommand\refname{References}
\bibliographystyle{apalike}
\bibliography{Private_RS_1226}

\begin{thebibliography}{}

\bibitem[Alabi et~al., 2022]{alabi2022private}
Alabi, D., Ghazi, B., Kumar, R., and Manurangsi, P. (2022).
\newblock Private rank aggregation in central and local models.
\newblock In {\em Proceedings of the AAAI Conference on Artificial
  Intelligence}, volume~36, pages 5984--5991.

\bibitem[Arratia and Gordon, 1989]{arratia1989tutorial}
Arratia, R. and Gordon, L. (1989).
\newblock Tutorial on large deviations for the binomial distribution.
\newblock {\em Bulletin of Mathematical Biology}, 51(1):125--131.

\bibitem[Balakrishnan and Chopra, 2012]{balakrishnan2012collaborative}
Balakrishnan, S. and Chopra, S. (2012).
\newblock Collaborative ranking.
\newblock In {\em Proceedings of the fifth ACM International Conference on Web
  Search and Data Mining}, pages 143--152.

\bibitem[Bartlett et~al., 2006]{bartlett2006convexity}
Bartlett, P.~L., Jordan, M.~I., and McAuliffe, J.~D. (2006).
\newblock Convexity, classification, and risk bounds.
\newblock {\em Journal of the American Statistical Association},
  101(473):138--156.

\bibitem[Busa-Fekete et~al., 2021]{busa2021private}
Busa-Fekete, R., Fotakis, D., and Zampetakis, E. (2021).
\newblock Private and non-private uniformity testing for ranking data.
\newblock {\em Advances in Neural Information Processing Systems},
  34:9480--9492.

\bibitem[Busa-Fekete et~al., 2014]{busa2014preference}
Busa-Fekete, R., H{\"u}llermeier, E., and Sz{\"o}r{\'e}nyi, B. (2014).
\newblock Preference-based rank elicitation using statistical models: The case
  of mallows.
\newblock In {\em International Conference on Machine Learning}, pages
  1071--1079. PMLR.

\bibitem[Chen and Stallaert, 2014]{chen2014economic}
Chen, J. and Stallaert, J. (2014).
\newblock An economic analysis of online advertising using behavioral
  targeting.
\newblock {\em Mis Quarterly}, 38(2):429--A7.

\bibitem[Choirat and Seri, 2012]{choirat2012estimation}
Choirat, C. and Seri, R. (2012).
\newblock Estimation in discrete parameter models.
\newblock {\em Statistical Science}, 27(2):278--293.

\bibitem[Critchlow et~al., 1991]{critchlow1991probability}
Critchlow, D.~E., Fligner, M.~A., and Verducci, J.~S. (1991).
\newblock Probability models on rankings.
\newblock {\em Journal ofMathematical Psychology}, 35(3):294--318.

\bibitem[Dai et~al., 2021]{dai2021scalable}
Dai, B., Shen, X., Wang, J., and Qu, A. (2021).
\newblock Scalable collaborative ranking for personalized prediction.
\newblock {\em Journal of the American Statistical Association},
  116(535):1215--1223.

\bibitem[D{\'e}sir et~al., 2016]{desir2016assortment}
D{\'e}sir, A., Goyal, V., Jagabathula, S., and Segev, D. (2016).
\newblock Assortment optimization under the mallows model.
\newblock {\em Advances in Neural Information Processing Systems}, 29.

\bibitem[Duchi and Ruan, 2018]{duchi2018right}
Duchi, J.~C. and Ruan, F. (2018).
\newblock The right complexity measure in locally private estimation: It is not
  the fisher information.
\newblock {\em arXiv preprint arXiv:1806.05756}.

\bibitem[Dwork, 2006]{dwork2006differential}
Dwork, C. (2006).
\newblock Differential privacy.
\newblock In {\em International Colloquium on Automata, Languages, and
  Programming}, pages 1--12. Springer.

\bibitem[Dwork et~al., 2006]{dwork2006our}
Dwork, C., Kenthapadi, K., McSherry, F., Mironov, I., and Naor, M. (2006).
\newblock Our data, ourselves: Privacy via distributed noise generation.
\newblock In {\em Annual International Conference on The Theory and
  Applications of Cryptographic Techniques}, pages 486--503. Springer.

\bibitem[Dwork et~al., 2001]{dwork2001rank}
Dwork, C., Kumar, R., Naor, M., and Sivakumar, D. (2001).
\newblock Rank aggregation methods for the web.
\newblock In {\em Proceedings of the 10th International Conference on World
  Wide Web}, pages 613--622.

\bibitem[Erlingsson et~al., 2014]{erlingsson2014rappor}
Erlingsson, {\'U}., Pihur, V., and Korolova, A. (2014).
\newblock Rappor: Randomized aggregatable privacy-preserving ordinal response.
\newblock In {\em Proceedings of the 2014 ACM SIGSAC Conference on Computer and
  Communications Security}, pages 1054--1067.

\bibitem[Fligner and Verducci, 1986]{fligner1986distance}
Fligner, M.~A. and Verducci, J.~S. (1986).
\newblock Distance based ranking models.
\newblock {\em Journal of the Royal Statistical Society: Series B
  (Methodological)}, 48(3):359--369.

\bibitem[Fligner and Verducci, 1988]{fligner1988multistage}
Fligner, M.~A. and Verducci, J.~S. (1988).
\newblock Multistage ranking models.
\newblock {\em Journal of the American Statistical Association},
  83(403):892--901.

\bibitem[Gao and Zhou, 2015]{gao2015consistency}
Gao, W. and Zhou, Z.-H. (2015).
\newblock On the consistency of auc pairwise optimization.
\newblock In {\em Twenty-Fourth International Joint Conference on Artificial
  Intelligence}.

\bibitem[Hay et~al., 2017]{hay2017differentially}
Hay, M., Elagina, L., and Miklau, G. (2017).
\newblock Differentially private rank aggregation.
\newblock In {\em Proceedings of the 2017 SIAM International Conference on Data
  Mining}, pages 669--677. SIAM.

\bibitem[Jeckmans et~al., 2013]{jeckmans2013privacy}
Jeckmans, A.~J., Beye, M., Erkin, Z., Hartel, P., Lagendijk, R.~L., and Tang,
  Q. (2013).
\newblock Privacy in recommender systems.
\newblock In {\em Social Media Retrieval}, pages 263--281. Springer.

\bibitem[Jeong et~al., 2022]{jeong2022ranking}
Jeong, M., Dytso, A., and Cardone, M. (2022).
\newblock Ranking recovery under privacy considerations.
\newblock {\em Transactions on Machine Learning Research}.

\bibitem[Kairouz et~al., 2015]{kairouz2015composition}
Kairouz, P., Oh, S., and Viswanath, P. (2015).
\newblock The composition theorem for differential privacy.
\newblock In {\em International Conference on Machine Learning}, pages
  1376--1385. PMLR.

\bibitem[Karatzoglou et~al., 2013]{karatzoglou2013learning}
Karatzoglou, A., Baltrunas, L., and Shi, Y. (2013).
\newblock Learning to rank for recommender systems.
\newblock In {\em Proceedings of the 7th ACM Conference on Recommender
  Systems}, pages 493--494.

\bibitem[Koltchinskii, 2011]{koltchinskii2011oracle}
Koltchinskii, V. (2011).
\newblock {\em Oracle inequalities in empirical risk minimization and sparse
  recovery problems: {\'E}cole D’{\'E}t{\'e} de Probabilit{\'e}s de
  Saint-Flour XXXVIII-2008}, volume 2033.
\newblock Springer Science \& Business Media.

\bibitem[Lantz et~al., 2015]{lantz2015subsampled}
Lantz, E., Boyd, K., and Page, D. (2015).
\newblock Subsampled exponential mechanism: Differential privacy in large
  output spaces.
\newblock In {\em Proceedings of the 8th ACM Workshop on Artificial
  Intelligence and Security}, pages 25--33.

\bibitem[Lee, 2015]{lee2015efficient}
Lee, D.~T. (2015).
\newblock Efficient, private, and eps-strategyproof elicitation of tournament
  voting rules.
\newblock In {\em Twenty-Fourth International Joint Conference on Artificial
  Intelligence}.

\bibitem[Li et~al., 2022]{li2022differentially}
Li, Z., Liu, A., Xia, L., Cao, Y., and Wang, H. (2022).
\newblock Differentially private condorcet voting.
\newblock {\em arXiv preprint arXiv:2206.13081}.

\bibitem[Liu et~al., 2007]{liu2007supervised}
Liu, Y.-T., Liu, T.-Y., Qin, T., Ma, Z.-M., and Li, H. (2007).
\newblock Supervised rank aggregation.
\newblock In {\em Proceedings of the 16th International Conference on World
  Wide Web}, pages 481--490.

\bibitem[Mallows, 1957]{mallows1957non}
Mallows, C.~L. (1957).
\newblock Non-null ranking models. i.
\newblock {\em Biometrika}, 44(1/2):114--130.

\bibitem[Mandhani and Meila, 2009]{mandhani2009tractable}
Mandhani, B. and Meila, M. (2009).
\newblock Tractable search for learning exponential models of rankings.
\newblock In {\em Artificial Intelligence and Statistics}, pages 392--399.
  PMLR.

\bibitem[Mayer and Mitchell, 2012]{mayer2012third}
Mayer, J.~R. and Mitchell, J.~C. (2012).
\newblock Third-party web tracking: Policy and technology.
\newblock In {\em 2012 IEEE Symposium on Security and Privacy}, pages 413--427.
  IEEE.

\bibitem[McSherry and Talwar, 2007]{mcsherry2007mechanism}
McSherry, F. and Talwar, K. (2007).
\newblock Mechanism design via differential privacy.
\newblock In {\em 48th Annual IEEE Symposium on Foundations of Computer Science
  (FOCS'07)}, pages 94--103. IEEE.

\bibitem[Meil{\u{a}} and Bao, 2010]{meilua2010exponential}
Meil{\u{a}}, M. and Bao, L. (2010).
\newblock An exponential model for infinite rankings.
\newblock {\em Journal of Machine Learning Research}, 11(113):3481--3518.

\bibitem[Nadarajah, 2007]{nadarajah2007linear}
Nadarajah, S. (2007).
\newblock The linear combination, product and ratio of laplace random
  variables.
\newblock {\em Statistics}, 41(6):535--545.

\bibitem[Oliveira et~al., 2020]{oliveira2020rank}
Oliveira, S.~E., Diniz, V., Lacerda, A., Merschmanm, L., and Pappa, G.~L.
  (2020).
\newblock Is rank aggregation effective in recommender systems? an experimental
  analysis.
\newblock {\em ACM Transactions on Intelligent Systems and Technology (TIST)},
  11(2):1--26.

\bibitem[Pearce and Erosheva, 2022]{JMLR:v23:21-1262}
Pearce, M. and Erosheva, E.~A. (2022).
\newblock A unified statistical learning model for rankings and scores with
  application to grant panel review.
\newblock {\em Journal of Machine Learning Research}, 23(210):1--33.

\bibitem[Rendle et~al., 2012]{rendle2012bpr}
Rendle, S., Freudenthaler, C., Gantner, Z., and Schmidt-Thieme, L. (2012).
\newblock Bpr: Bayesian personalized ranking from implicit feedback.
\newblock {\em arXiv preprint arXiv:1205.2618}.

\bibitem[S{\'a}nchez et~al., 2016]{sanchez2016utility}
S{\'a}nchez, D., Domingo-Ferrer, J., Mart{\'\i}nez, S., and Soria-Comas, J.
  (2016).
\newblock Utility-preserving differentially private data releases via
  individual ranking microaggregation.
\newblock {\em Information Fusion}, 30:1--14.

\bibitem[Sason, 2015]{sason2015reverse}
Sason, I. (2015).
\newblock On reverse pinsker inequalities.
\newblock {\em arXiv preprint arXiv:1503.07118}.

\bibitem[Schapire, 2003]{schapire2003boosting}
Schapire, R.~E. (2003).
\newblock The boosting approach to machine learning: An overview.
\newblock {\em Nonlinear Estimation and Classification}, pages 149--171.

\bibitem[Shang et~al., 2014]{shang2014application}
Shang, S., Wang, T., Cuff, P., and Kulkarni, S. (2014).
\newblock The application of differential privacy for rank aggregation: Privacy
  and accuracy.
\newblock In {\em 17th International Conference on Information Fusion
  (FUSION)}, pages 1--7. IEEE.

\bibitem[Shen et~al., 2003]{shen2003psi}
Shen, X., Tseng, G.~C., Zhang, X., and Wong, W.~H. (2003).
\newblock On $\psi$-learning.
\newblock {\em Journal of the American Statistical Association},
  98(463):724--734.

\bibitem[Song et~al., 2022]{song2022distributed}
Song, B., Lan, Q., Li, Y., and Li, G. (2022).
\newblock Distributed differentially private ranking aggregation.
\newblock In {\em Pacific-Asia Conference on Knowledge Discovery and Data
  Mining}, pages 236--248. Springer.

\bibitem[Soufiani et~al., 2014]{soufiani2014computing}
Soufiani, H.~A., Parkes, D., and Xia, L. (2014).
\newblock Computing parametric ranking models via rank-breaking.
\newblock In {\em International Conference on Machine Learning}, pages
  360--368. PMLR.

\bibitem[Su, 2021]{su2021you}
Su, W. (2021).
\newblock You are the best reviewer of your own papers: An owner-assisted
  scoring mechanism.
\newblock {\em Advances in Neural Information Processing Systems},
  34:27929--27939.

\bibitem[Sun et~al., 2018]{sun2018truth}
Sun, H., Dong, B., Wang, H., Yu, T., and Qin, Z. (2018).
\newblock Truth inference on sparse crowdsourcing data with local differential
  privacy.
\newblock In {\em 2018 IEEE International Conference on Big Data (Big Data)},
  pages 488--497. IEEE.

\bibitem[Tang et~al., 2017]{tang2017privacy}
Tang, J., Korolova, A., Bai, X., Wang, X., and Wang, X. (2017).
\newblock Privacy loss in apple's implementation of differential privacy on
  macos 10.12.
\newblock {\em arXiv preprint arXiv:1709.02753}.

\bibitem[Tang, 2019]{tang2019mallows}
Tang, W. (2019).
\newblock Mallows ranking models: maximum likelihood estimate and regeneration.
\newblock In {\em International Conference on Machine Learning}, pages
  6125--6134. PMLR.

\bibitem[Walker and Ben-Akiva, 2002]{walker2002generalized}
Walker, J. and Ben-Akiva, M. (2002).
\newblock Generalized random utility model.
\newblock {\em Mathematical social sciences}, 43(3):303--343.

\bibitem[Wang et~al., 2021]{wang2021cross}
Wang, J., Zhu, J., and He, X. (2021).
\newblock Cross-batch negative sampling for training two-tower recommenders.
\newblock In {\em Proceedings of the 44th International ACM SIGIR Conference on
  Research and Development in Information Retrieval}, pages 1632--1636.

\bibitem[Wang et~al., 2017]{wang2017locally}
Wang, T., Blocki, J., Li, N., and Jha, S. (2017).
\newblock Locally differentially private protocols for frequency estimation.
\newblock In {\em 26th USENIX Security Symposium (USENIX Security 17)}, pages
  729--745.

\bibitem[Watson et~al., 2015]{watson2015mapping}
Watson, J., Lipford, H.~R., and Besmer, A. (2015).
\newblock Mapping user preference to privacy default settings.
\newblock {\em ACM Transactions on Computer-Human Interaction (TOCHI)},
  22(6):1--20.

\bibitem[Williams and McSherry, 2010]{williams2010probabilistic}
Williams, O. and McSherry, F. (2010).
\newblock Probabilistic inference and differential privacy.
\newblock {\em Advances in Neural Information Processing Systems}, 23.

\bibitem[Wu et~al., 2022]{wu2022does}
Wu, R., Zhou, J.~P., Weinberger, K.~Q., and Guo, C. (2022).
\newblock Does label differential privacy prevent label inference attacks?
\newblock {\em arXiv preprint arXiv:2202.12968}.

\bibitem[Yan et~al., 2020]{yan2020private}
Yan, Z., Li, G., and Liu, J. (2020).
\newblock Private rank aggregation under local differential privacy.
\newblock {\em International Journal of Intelligent Systems},
  35(10):1492--1519.

\bibitem[Yang et~al., 2019]{yang2019collecting}
Yang, J., Cheng, X., Su, S., Chen, R., Ren, Q., and Liu, Y. (2019).
\newblock Collecting preference rankings under local differential privacy.
\newblock In {\em 2019 IEEE 35th International Conference on Data Engineering
  (ICDE)}, pages 1598--1601. IEEE.

\bibitem[Yang et~al., 2020]{yang2020mixed}
Yang, J., Yi, X., Zhiyuan~Cheng, D., Hong, L., Li, Y., Xiaoming~Wang, S., Xu,
  T., and Chi, E.~H. (2020).
\newblock Mixed negative sampling for learning two-tower neural networks in
  recommendations.
\newblock In {\em Companion Proceedings of the Web Conference 2020}, pages
  441--447.

\bibitem[Young, 1986]{young1986optimal}
Young, H.~P. (1986).
\newblock Optimal ranking and choice from pairwise comparisons.
\newblock {\em Information Pooling and Group Decision Making}, pages 113--122.

\bibitem[Zhang, 2004]{zhang2004statistical}
Zhang, T. (2004).
\newblock Statistical behavior and consistency of classification methods based
  on convex risk minimization.
\newblock {\em The Annals of Statistics}, 32(1):56--85.

\bibitem[Zhu and Hastie, 2005]{zhu2005kernel}
Zhu, J. and Hastie, T. (2005).
\newblock Kernel logistic regression and the import vector machine.
\newblock {\em Journal of Computational and Graphical Statistics},
  14(1):185--205.

\end{thebibliography}

\newpage
\begin{center}
{\Large\bf Supplementary Materials} \\
\medskip
{\Large\bf ``Ranking Differential Privacy"}  \\
\bigskip
\textbf{Shirong Xu, Will Wei Sun, and Guang Cheng}
\vspace{0.2in} 
\end{center}
\bigskip

\noindent
\textbf{Proof of Lemma \ref{lemma:eDP}}:  Let $\phi$ and $\phi'$ be two neighboring ranking differing such that
\begin{align*}
\big(\phi(\mathcal{I}_i)-\phi(\mathcal{I}_j)\big)\big(\phi'(\mathcal{I}_i)-\phi'(\mathcal{I}_j)\big)>0,
\end{align*}
for any $\mathcal{I}_i,\mathcal{I}_j \in \Omega \setminus \{I_{k}\}$. Therefore, for any synthetic ranking $\widetilde{\phi}$ and $\mathcal{I}_i,\mathcal{I}_j \in \Omega \setminus \{\mathcal{I}_{k}\}$, it holds that
\begin{align*}
I\Big(
(\phi(\mathcal{I}_i)-\phi(\mathcal{I}_j))(\widetilde{\phi}(\mathcal{I}_i)-\widetilde{\phi}(\mathcal{I}_j))>0
\Big) -
I\Big(
(\phi'(\mathcal{I}_i)-\phi'(\mathcal{I}_j))(\widetilde{\phi}(\mathcal{I}_i)-\widetilde{\phi}(\mathcal{I}_j))>0
\Big) = 0.
\end{align*}
Let $M_\Omega=\{(\mathcal{I}_i,\mathcal{I}_j): \mathcal{I}_i,\mathcal{I}_j\in \Omega, \mathcal{I}_i \neq \mathcal{I}_j\}$. For any synthetic ranking $\widetilde{\phi}  $, we get
\begin{align*}
\Big| 
T(\phi,\widetilde{\phi}  ) - 
T(\phi ',\widetilde{\phi})
\Big|  
&=  
\Big| \frac{1}{|M_\Omega|}
\sum_{(i,j)\in M_\Omega}
I\Big((\phi(\mathcal{I}_i)-\phi  (\mathcal{I}_j))(\widetilde{\phi}  (\mathcal{I}_i)-\widetilde{\phi}  (\mathcal{I}_j))>0\Big)  \\
&-  \frac{1}{|M_\Omega|}
\sum_{(i,j)\in M_\Omega}
I\Big((\phi  '(\mathcal{I}_i)-\phi  '(\mathcal{I}_j))(\widetilde{\phi}  (\mathcal{I}_i)-\widetilde{\phi}  (\mathcal{I}_j))>0\Big) 
\Big| \\
& =
\Big| \frac{2}{|M_\Omega|}
\sum_{\mathcal{I}_j \in \Omega \setminus \{ \mathcal{I}_k \}}
I\Big((\phi  (\mathcal{I}_k)-\phi  (\mathcal{I}_j))(\widetilde{\phi}  (\mathcal{I}_k)-\widetilde{\phi}  (\mathcal{I}_j))>0\Big)\\
&-
\frac{2}{|M_\Omega|}
\sum_{I_j \in \Omega\setminus \{ \mathcal{I}_k \}}
I\Big((\phi  '(\mathcal{I}_k)-\phi  '(\mathcal{I}_j))(\widetilde{\phi}  (\mathcal{I}_k)-\widetilde{\phi}  (\mathcal{I}_j))>0\Big)
\Big|  \\
&\leq \frac{2}{|\Omega|}.
\end{align*}
Therefore, for any neighboring rankings $\phi  $ and $\phi  '$, it holds that
$$
\max_{\widetilde{\phi}}
\Big| 
T(\phi  ,\widetilde{\phi}  ) - 
T(\phi  ',\widetilde{\phi}  )
\Big|  \leq \frac{2}{|\Omega|}.
$$
The desired result immediately follows by setting $\theta=2^{-1}|\Omega|\epsilon$. \qed

\noindent
\textbf{Proof of Lemma $\ref{lemma:Prob_u}$:} 
Let $\mathcal{A}$ denote algorithm $\ref{algorithm1}$ and $\widetilde{\phi}=\mathcal{A}(\phi)$ denote the output synthetic ranking. Let $\Omega = \{\mathcal{I}_1,\mathcal{I}_2,\ldots,\mathcal{I}_L \}$ denote the item set. Suppose that $\widetilde{\phi}$ satisfies that $\widetilde{\phi}(\mathcal{I}_i)  = s_{\mathcal{I}_i}$ for $i \in [L]$ denote the output synthetic ranking, where $\bm{s}=(s_{\mathcal{I}_i})_{i \in [L]}$ is a permutation of $(i)_{i \in [L]}$. 

According to the Algorithm \ref{algorithm1}, at the iteration $t \in \{ 2,\ldots,L\}$, the selected item $\mathcal{I}_j$ has rank $t$ among the items, that is $\mathcal{I}_j = \phi^{-1}(t)$. Notice that all selected items in the previous items all have smaller ordinal ranks. Therefore it is straightforward to see that
\begin{align*}
\mathbb{P}(V_{\epsilon}^{(t)} = k)  \propto
\exp\big(
\epsilon (L-1)^{-1}k
\big),
\end{align*}
for $k = 0,\ldots,t-1$. With this, we further have
\begin{align*}
\mathbb{P}(V_{\epsilon}^{(t)} = k) =
\frac{\exp\big(
\epsilon k (L-1)^{-1}
\big)}{\sum_{k=0}^{t-1}\exp\big(
\epsilon  k (L-1)^{-1}
\big)},k=0,\ldots,t-1.
\end{align*}

Define $H_t(\bm{s}) = \{i:\phi(\mathcal{I}_i)<t, s_{\mathcal{I}_i}<s_{\phi^{-1}(t)},i\in [L]  \}$ for $t=2,\ldots,L$. It is easy to verify that $|H_t(\bm{s})| =k$ implies that $V^{(t)}=k$. Notice that $|H_1(\bm{s})|=0$, it then follows that
\begin{align}
\label{Prob_sigma}
\mathbb{P}\Big(\mathcal{A}(\phi) = \widetilde{\phi}\Big) = 
\prod_{t=2}^{L}
\frac{\exp(\epsilon (L-1)^{-1}|H_t(\bm{s})|)}{\sum_{k=0}^{t-1} \exp(\epsilon (L-1)^{-1} k)} \propto \exp\Big(\epsilon (L-1)^{-1} \sum_{t=1}^L |H_t(\bm{s})|\Big).
\end{align}
Next, it remains to prove that $\sum_{t=1}^L |H_t(\bm{s})| = 2^{-1}|\Omega|(|\Omega|-1)T(\phi,\widetilde{\phi})$. By the definition of $H_t(\bm{s})$, we have
\begin{align*}
|H_t(\bm{s})| &= \sum_{i: \phi(\mathcal{I}_i) < t} I\Big(
(\phi(\mathcal{I}_i)-t)
(s_{\mathcal{I}_i} - s_{\phi^{-1}(t)})>0
\Big) \\
& =\sum_{i: \phi(\mathcal{I}_i) < t} I\Big(
(\phi(\mathcal{I}_i)-\phi(\mathcal{I}_j))
(\widetilde{\phi}(\mathcal{I}_i) - \widetilde{\phi}(\mathcal{I}_j))>0
\Big),
\end{align*}
where $\mathcal{I}_j = \phi^{-1}(t)$. Summing over $t$ from $1$ to $L$ yields that
\begin{align}
\label{Sum_H_i}
\sum_{t=1}^L |H_t(\bm{s})| =&\sum_{t=1}^L \sum_{i: \phi(\mathcal{I}_i)  < t} I\Big(
(\phi(\mathcal{I}_i)-t)
(s_{\mathcal{I}_i} - s_{\phi^{-1}(t)})>0
\Big) \notag \\
=&
\sum_{j=1}^L \sum_{ \{ i: \phi(\mathcal{I}_i)  < \phi(\mathcal{I}_j) \}}
I\Big(
(\phi(\mathcal{I}_i)-\phi(\mathcal{I}_j))
(\widetilde{\phi}(\mathcal{I}_i) - \widetilde{\phi}(\mathcal{I}_j))>0
\Big) \notag \\
=&\frac{1}{2}
\sum_{j=1}^L \sum_{i \neq j}
I\Big(
(\phi(\mathcal{I}_i)-\phi(\mathcal{I}_j))
(\widetilde{\phi}(\mathcal{I}_i) - \widetilde{\phi}(\mathcal{I}_j))>0
\Big) \notag \\
=&\frac{1}{2}|\Omega|(|\Omega|-1)
 T(\phi,\widetilde{\phi}).
\end{align}
Plugging $(\ref{Sum_H_i})$ into $(\ref{Prob_sigma})$, it follows that
\begin{align*}
\mathbb{P}\Big(\mathcal{A}(\phi) 
=\widetilde{\phi}
\Big)\propto \exp(2^{-1}\epsilon |\Omega|T(\phi,\widetilde{\phi})).
\end{align*}
This completes the proof. \qed


\noindent
\textbf{Proof of Theorem $\ref{Thm:ConsisandPriv}$}: First, we prove the property (1). Let $\Omega = \{ \mathcal{I}_1,\ldots,\mathcal{I}_L \}$ be the item set. We consider the case that $\mathbb{P}\big(\Phi(\mathcal{I}_i)>\Phi(\mathcal{I}_j)\big)>\mathbb{P}\big(\Phi(\mathcal{I}_i)<\Phi(\mathcal{I}_j)\big)$, which implies that $2\eta_{ij}-1>0$. Furthermore,  For any $\mathcal{I}_i,\mathcal{I}_j \in \Omega$, we define two sets of rankings as
\begin{align*}
&Y_{ij}^{g} =\big \{\phi:\big(\phi(\mathcal{I}_i)\big)_{\mathcal{I}_i\in \Omega}\in \Upsilon(L), \phi(\mathcal{I}_i)>\phi(\mathcal{I}_j)\big \}, \\
&Y_{ij}^{l} = \big \{\phi:\big(\phi(\mathcal{I}_i)\big)_{\mathcal{I}_i\in \Omega} \in \Upsilon(L), \phi(\mathcal{I}_i)<\phi(\mathcal{I}_j)\big\},
\end{align*}
where $Y_{ij}^{g}$ and $Y_{ij}^{l}$ denote the ranking sets whose elements satisfying $\phi(\mathcal{I}_i)>\phi(\mathcal{I}_j)$ and $\phi(\mathcal{I}_j)>\phi(\mathcal{I}_i)$, respectively.

Let $\widetilde{\Phi}=\mathcal{A}(\Phi)$ denote the random variable of synthetic ranking. Next, we intend to prove $\mathbb{P}\big( \widetilde{\Phi} (\mathcal{I}_i)>\widetilde{\Phi} (\mathcal{I}_j) \big)>\mathbb{P}\big( \widetilde{\Phi} (\mathcal{I}_i)<\widetilde{\Phi}(\mathcal{I}_j) \big)$, which are given by
\begin{align*}
&\mathbb{P}\big( \widetilde{\Phi} (\mathcal{I}_i)>\widetilde{\Phi} (\mathcal{I}_j) \big) = 
\mathbb{P}\big(
\widetilde{\Phi} \in Y_{ij}^{g}
\big)=
\sum_{\widetilde{\phi}  \in Y_{ij}^{g}} 
\mathbb{P}\Big(\mathcal{A}(\Phi) 
=\widetilde{\phi} 
\Big),  \\
&\mathbb{P}\big( \widetilde{\Phi} (\mathcal{I}_i)<\widetilde{\Phi} (\mathcal{I}_j) \big) = 
\mathbb{P}\big(
\widetilde{\Phi} \in Y_{ij}^{l}
\big)=
\sum_{\widetilde{\phi}  \in Y_{ij}^{l}} 
\mathbb{P}\Big(\mathcal{A}(\Phi) 
=\widetilde{\phi} 
\Big),
\end{align*}
respectively.

Consider two synthetic rankings $\widetilde{\phi}_1 \in Y_{ij}^g$ and $\widetilde{\phi}_2 \in Y_{ij}^l$ satisfying that $\widetilde{\phi}_1(\mathcal{I}_i) =\widetilde{\phi}_2(\mathcal{I}_j)=k_1$, $\widetilde{\phi}_1(\mathcal{I}_j) =\widetilde{\phi}_2(\mathcal{I}_i)= k_2$ with $k_1>k_2$, and $\widetilde{\phi}_1(\mathcal{I}_l)=\widetilde{\phi}_2(\mathcal{I}_l)$ for $l \in [L]\setminus \{i,j \}$. Notice that $\widetilde{\phi}_1$ and $\widetilde{\phi}_2$ are a couple of ordinal rankings by the one-to-one correspondence between $\widetilde{\phi}_1 \in Y_{ij}^g$ and $\widetilde{\phi}_2 \in Y_{ij}^l$, therefore it is easy to see that $|Y_{ij}^l| = |Y_{ij}^g|$. By Lemma $\ref{lemma:Prob_u}$, we have 
\begin{align*}
\frac{\mathbb{P}\Big(\mathcal{A}(\phi) 
=\widetilde{\phi}_1
\Big)}{\mathbb{P}\Big(\mathcal{A}(\phi) 
=\widetilde{\phi}_2
\Big)}
= \exp\Big(2^{-1}\epsilon L \big(T(\phi,\widetilde{\phi}_1)-T(\phi,\widetilde{\phi}_2)\big)\Big),
\end{align*}
Therefore, it suffices to compare the values of quality function $T$ to obtain the relative order of $\mathbb{P}\big(\mathcal{A}(\phi\big) 
=\widetilde{\phi}_1\big)$ and $\mathbb{P}\big(\mathcal{A}(\phi) 
=\widetilde{\phi}_2\big)$.

For any ranking $\phi$, define $S(\phi,k_2,k_1) = \{i:k_2<\phi(\mathcal{I}_i)<k_1, i \in [L] \}$ as the subset of $\Omega $ with ordinal ranks smaller than $k_1$ and greater than $k_2$. By the definitions of $\widetilde{\phi}_1$ and $\widetilde{\phi}_2$, it is straightforward to verify that 
$$
S(\widetilde{\phi}_1,k_2,k_1) = S(\widetilde{\phi}_2, k_2, k_1).
$$ 
Further, we denote that $\Delta_{ij}(\phi,\widetilde{\phi}_1) =I\big((\phi(\mathcal{I}_i)-\phi(\mathcal{I}_j))(\widetilde{\phi}_1(\mathcal{I}_i)-\widetilde{\phi}_1(\mathcal{I}_j))>0 \big) $. By the definition of $T$, we get
\begin{align}
\label{T_diff}
&T(\phi,\widetilde{\phi}_1) -
T(\phi,\widetilde{\phi}_2) 
=
\frac{1}{L(L-1)}\sum_{\mathcal{I}_i ,\mathcal{I}_j \in \Omega}
\Big(
\Delta_{ij}(\phi,\widetilde{\phi}_1) - 
\Delta_{ij}(\phi,\widetilde{\phi}_2) \Big)\notag  \\
=&
\frac{2}{L(L-1)}\sum_{l \in S(\widetilde{\phi}_1,k_2,k_1) }
\Big(
\Delta_{il}(\phi,\widetilde{\phi}_1) +
\Delta_{jl}(\phi,\widetilde{\phi}_1) -
 \Delta_{jl}(\phi,\widetilde{\phi}_2)- 
\Delta_{il}(\phi,\widetilde{\phi}_2) \Big) \notag \\
+& \frac{2}{L(L-1)}\Big( \Delta_{ij}(\phi,\widetilde{\phi}_1) - 
\Delta_{ij}(\phi,\widetilde{\phi}_2) \Big).
\end{align}
Next, we consider two cases that $\phi  \in Y_{ij}^g$ and $\phi  \in Y_{ij}^l$. Notice that $\phi  \in Y_{ij}^g$ implies $\phi(\mathcal{I}_i)>\phi(\mathcal{I}_j)$ and $\Delta_{ij}(\phi,\widetilde{\phi}_1) - 
\Delta_{ij}(\phi,\widetilde{\phi}_2)=1$. For any $l \in S(\widetilde{\phi}_1,k_2,k_1)$, it can be verified that 
\begin{align*}
\Delta_{il}(\phi,\widetilde{\phi}_1) +
\Delta_{jl}(\phi,\widetilde{\phi}_1) -
 \Delta_{jl}(\phi,\widetilde{\phi}_2)- 
\Delta_{il}(\phi,\widetilde{\phi}_2) =
\begin{cases}
2, \phi(\mathcal{I}_j)<\phi(\mathcal{I}_l)< \phi(\mathcal{I}_i),\\
0, \mbox{otherwise}.
\end{cases}
\end{align*}
Therefore, if $\phi  \in Y_{ij}^g$, $(\ref{T_diff})$ can be written as
\begin{align*}
T(\phi,\widetilde{\phi}_1) -
T(\phi,\widetilde{\phi}_2) = 
\frac{4| S(\widetilde{\phi}_1,k_2,k_1) \cap S(\phi,\phi(\mathcal{I}_j),\phi(\mathcal{I}_i)) |+2}{L(L-1)}.
\end{align*}
Hence, it follows that
\begin{align*}
\frac{2}{L(L-1)} \leq 
T(\phi,\widetilde{\phi}_1) -
T(\phi,\widetilde{\phi}_2) \leq \frac{4(L-2)+2}{L(L-1)}.
\end{align*} 
Consequently, for any $\phi \in Y_{ij}^g$, it holds that
\begin{align*}
&\frac{\mathbb{P}\big(\mathcal{A}(\phi) \in Y_{ij}^g\big)}{\mathbb{P}\big(\mathcal{A}(\phi) \in Y_{ij}^l\big)}
 =
\frac{\sum_{\widetilde{\phi}_1  \in Y_{ij}^g} \mathbb{P}\big(
\mathcal{A}(\phi)=
\widetilde{\phi}_1 
\big)}{\sum_{\widetilde{\phi}_2  \in Y_{ij}^l} \mathbb{P}\big(
\mathcal{A}(\phi)=
\widetilde{\phi}_2 
\big)} 
 \geq \exp\Big((L-1)^{-1}\epsilon\Big), \\
& \frac{\mathbb{P}\big(\mathcal{A}(\phi) \in Y_{ij}^g\big)}{\mathbb{P}\big(\mathcal{A}(\phi) \in Y_{ij}^l\big)}
 =
\frac{\sum_{\widetilde{\phi}_1  \in Y_{ij}^g} \mathbb{P}\big(
\mathcal{A}(\phi)=
\widetilde{\phi}_1
\big)}{\sum_{\widetilde{\phi}_2  \in Y_{ij}^l} \mathbb{P}\big(
\mathcal{A}(\phi)=
\widetilde{\phi}_2
\big)} 
\leq \exp\Big(\frac{2L-3}{L-1}\epsilon\Big).
\end{align*}
Applying a similar argument for the case $\phi  \in Y_{ij}^l$, we have
\begin{align*}
\exp\Big((L-1)^{-1}\epsilon\Big) \leq 
\frac{\mathbb{P}\big(\mathcal{A}(\phi) \in Y_{ij}^l\big)}{\mathbb{P}\big(\mathcal{A}(\phi) \in Y_{ij}^g\big)}
\leq \exp\Big(\frac{2L-3}{L-1}\epsilon\Big).
\end{align*}
Notice that $Y_{ij}^g$ and $Y_{ij}^l$ are mutually exclusive and $\Upsilon = Y_{ij}^g \cup Y_{ij}^l$, we obtain that
\begin{align}
\label{Gen_sig_dff1}
&\mathbb{P}\big(\widetilde{\Phi}(\mathcal{I}_i)>\widetilde{\Phi} (\mathcal{I}_j)\big)-
\mathbb{P}\big(\widetilde{\Phi} (\mathcal{I}_j)>\widetilde{\Phi} (\mathcal{I}_i)\big) \notag \\
=&
\Big(
\mathbb{P}\big(\widetilde{\Phi}  \in Y_{ij}^g|\Phi\in Y_{ij}^g\big) - 
\mathbb{P}\big(\widetilde{\Phi}  \in Y_{ij}^l|\Phi\in Y_{ij}^g\big)  \Big)
 \mathbb{P}\big(\Phi\in Y_{ij}^g\big)\notag \\
+&
\Big(
\mathbb{P}\big(\widetilde{\Phi}  \in Y_{ij}^g|\Phi\in Y_{ij}^l\big)  - 
\mathbb{P}\big(\widetilde{\Phi}  \in Y_{ij}^l|\Phi\in Y_{ij}^l\big) 
 \Big)
\mathbb{P}\big(\Phi\in Y_{ij}^l\big).
\end{align}
By the symmetry between $Y_{ij}^l$ and $Y_{ij}^g$, it follows that 
\begin{align}
\label{SymmetryY}
&\mathbb{P}\big(\widetilde{\Phi}  \in Y_{ij}^g|\Phi\in Y_{ij}^g\big) - 
\mathbb{P}\big(\widetilde{\Phi}  \in Y_{ij}^l|\Phi\in Y_{ij}^g\big)\notag \\
 =&
\mathbb{P}\big(\widetilde{\Phi}  \in Y_{ij}^l|\Phi\in Y_{ij}^l\big) -
\mathbb{P}\big(\widetilde{\Phi}  \in Y_{ij}^g|\Phi\in Y_{ij}^l\big).
\end{align}
Plugging $(\ref{SymmetryY})$ into $(\ref{Gen_sig_dff1})$ yields that
\begin{align}
\label{Equ:equiv}
&\mathbb{P}\big(\widetilde{\Phi} (\mathcal{I}_i)>\widetilde{\Phi} (\mathcal{I}_j)\big)-
\mathbb{P}\big(\widetilde{\Phi}(\mathcal{I}_j)>\widetilde{\Phi} (\mathcal{I}_i)\big) 
\notag
\\
=&
\Big(
\mathbb{P}\big(\widetilde{\Phi}  \in Y_{ij}^g|\Phi\in Y_{ij}^g\big) - 
\mathbb{P}\big(\widetilde{\Phi}  \in Y_{ij}^l|\Phi\in Y_{ij}^g\big)
\Big)\Big(
 \mathbb{P}\big(\Phi\in Y_{ij}^g\big)-
  \mathbb{P}\big(\Phi\in Y_{ij}^l\big)
\Big) \notag \\
=&
\Big(
\mathbb{P}\big(\widetilde{\Phi}  \in Y_{ij}^g|\Phi\in Y_{ij}^g\big) - 
\mathbb{P}\big(\widetilde{\Phi}  \in Y_{ij}^l|\Phi\in Y_{ij}^g\big)
\Big)\Big(
\mathbb{P}\big(\Phi(\mathcal{I}_i)>\Phi(\mathcal{I}_j)\big)  - 
\mathbb{P}\big(\Phi(\mathcal{I}_j)>\Phi(\mathcal{I}_i)\big)\Big) \notag \\
=& 
\Big(
\mathbb{P}\big(\widetilde{\Phi}  \in Y_{ij}^g|\Phi\in Y_{ij}^g\big) - 
\mathbb{P}\big(\widetilde{\Phi}  \in Y_{ij}^l|\Phi\in Y_{ij}^g\big)
\Big) (2\eta_{ij}-1)>0.
\end{align}
By the fact that $\mathbb{P}\big(\widetilde{\Phi}  \in Y_{ij}^g|\Phi\in Y_{ij}^g\big) - 
\mathbb{P}\big(\widetilde{\Phi}  \in Y_{ij}^l|\Phi\in Y_{ij}^g\big)>0$, it follows that
\begin{align*}
\Big(\mathbb{P}\big(\Phi(\mathcal{I}_i)>\Phi(I_j)\big)-\mathbb{P}\big(\Phi(\mathcal{I}_i)<\Phi(\mathcal{I}_j)\big)\Big)\Big(\mathbb{P}\big(\widetilde{\Phi}(I_i)>\widetilde{\Phi}(\mathcal{I}_j)\big)-\mathbb{P}\big(\widetilde{\Phi}(\mathcal{I}_i)<\widetilde{\Phi}(\mathcal{I}_j)\big)\Big)>0,
\end{align*}
This completes the proof of property $(1)$.

Next, we turn to prove property (2). For ease of notation, we let $A_1 = \mathbb{P}\big(\widetilde{\Phi} \in Y_{ij}^g|\Phi\in Y_{ij}^g\big)$ and $A_2 = \mathbb{P}\big(\widetilde{\Phi} \in Y_{ij}^l|\Phi\in Y_{ij}^g\big)$. By the fact that $A_1 +A_2 = 1$ and $ \exp\big((L-1)^{-1}\epsilon\big) \leq A_1/A_2 \leq \exp(\frac{2L-3}{L-1}\epsilon )$, we have
\begin{align*}
A_2 \leq \exp\big(-(L-1)^{-1}\epsilon\big)  A_1 \mbox{ and }
A_1 \leq \exp\big(\frac{2L-3}{L-1}\epsilon\big)  A_2.
\end{align*}
These combined with the fact that $A_1 +A_2 = 1$ imply that 
\begin{align*}
\frac{\exp\big((L-1)^{-1}\epsilon\big)-1}{\exp\big((L-1)^{-1}\epsilon\big)+1} \leq  A_1 - A_2
\leq 
\frac{\exp\big(\frac{2L-3}{L-1}\epsilon\big)-1}{\exp\big(\frac{2L-3}{L-1}\epsilon\big)+1}.
\end{align*}
Combined with (\ref{Equ:equiv}), we get
\begin{align*}
 \frac{\exp\big((L-1)^{-1}\epsilon\big)-1}{\exp\big((L-1)^{-1}\epsilon\big)+1} \leq 
\frac{|2\widetilde{\eta}_{ij}-1|}{|2\eta_{ij}-1|} \leq 
\frac{\exp\big(\frac{2L-3}{L-1}\epsilon\big)-1}{\exp\big(\frac{2L-3}{L-1}\epsilon\big)+1},
\end{align*}
where $\eta_{ij} = \mathbb{P}\big(\Phi(I_i)>\Phi(I_j)\big)$ and $\widetilde{\eta}_{ij} = \mathbb{P}\big(\widetilde{\Phi}(I_i)>\widetilde{\Phi}(I_j)\big)$. This completes the proof of property (2).

Next, we prove the property (3). In the $t$-th iteration, $V^{(t)}_{\epsilon}$ follows the distribution
\begin{align*}
\mathbb{P}(V^{(t)}_{\epsilon} = k) =\frac{\exp\Big(
\epsilon(|\Omega |-1)^{-1} \tau(k,\chi^{(t-1)})
\Big)}{\sum_{k \in \rho^{(t)}}\exp\Big(
\epsilon(|\Omega |-1)^{-1} \tau(k,\chi^{(t-1)})
\Big)}
, k =0,\ldots,t-1.
\end{align*}
Notice that items are selected sequentially by their ordinal ranks, hence $t>\sigma (l)$ for $l \in \chi^{(t)}$ and $V^{(t)}_{\epsilon}=k$ implies that $\tau(k,\chi^{(t)})=k$. Therefore, we have
\begin{align*}
\mathbb{E}\big[
V^{(t)}_{\epsilon}
\big] = \sum_{k=0}^{t-1} \mathbb{P}(V^{(t)}_{\epsilon} = k) k =
\sum_{k=0}^{t-1}
\frac{k\exp\Big(
\epsilon(|\Omega |-1)^{-1} \tau(k,\chi^{(t)})
\Big)}{\sum_{k \in \rho^{(t)}}\exp\Big(
\epsilon(|\Omega |-1)^{-1} \tau(k,\chi^{(t)})
\Big)}.
\end{align*}
For ease of notation, we denote that $Q =\sum_{k \in \rho^{(t)}}\exp\Big(
\epsilon(|\Omega |-1)^{-1} \tau(k,\chi^{(t)})
\Big)$. We have
\begin{align*}
\mathbb{E}\big[
V^{(t)}_{\epsilon}
\big]  = Q^{-1} \sum_{k=1}^{t-1} k\exp\Big(
\epsilon(|\Omega |-1)^{-1} k
\Big).
\end{align*}
Straightforward algebra shows that
\begin{align*}
\mathbb{E}\big[
V^{(t)}_{\epsilon}
\big] =& \frac{Q^{-1} (t-1)\exp\big(
\epsilon(|\Omega |-1)^{-1} t
\Big) -  Q^{-1} \sum_{k=1}^{t-1} \exp\big(
\epsilon(|\Omega |-1)^{-1} k
\big)}{\exp\big(
\epsilon(|\Omega |-1)^{-1} 
\big) - 1} \\
=&
\frac{(t-1)q_{\epsilon}^t}{q_{\epsilon}^t-1}-\frac{q_{\epsilon}^t-q_{\epsilon}}{(q_{\epsilon}-1)(q_{\epsilon}^t-1)},
\end{align*}
where $q_{\epsilon} = \exp\big(
\epsilon(|\Omega |-1)^{-1} 
\big)$. As $\epsilon$ goes to infinity, it can be verified that $\mathbb{E}\big[
V^{(t)}_{\epsilon}
\big]$ converges to $t-1$. 

Note that $\text{Var}(V^{(t)}_{\epsilon}) = \mathbb{E}(V^{(t)}_{\epsilon})^2 - \big( \mathbb{E}(V^{(t)}_{\epsilon}) \big)^2$, it remains to compute $\mathbb{E}(V^{(t)}_{\epsilon})^2$.
\begin{align*}
\mathbb{E}(V^{(t)}_{\epsilon})^2
 = 
Q^{-1} \sum_{k=1}^{t-1} k^2\exp\Big(
\epsilon(|\Omega |-1)^{-1} k
\Big).
\end{align*}
Applying a similar argument, we get
\begin{align*}
\mathbb{E}(V^{(t)}_{\epsilon})^2 &= 
\frac{Q^{-1} (t-1)^2\exp\Big(
\epsilon(|\Omega |-1)^{-1} t
\Big)}{\exp\Big(
\epsilon(|\Omega |-1)^{-1} 
\Big)-1}- \frac{
Q^{-1}
\sum_{k=1}^{t-1}
(2k-1)
\exp\Big(
\epsilon(|\Omega |-1)^{-1} k
\Big)}{\exp\Big(
\epsilon(|\Omega |-1)^{-1} 
\Big)-1} \\
&=
\frac{ (t-1)^2\exp\Big(
\epsilon(|\Omega |-1)^{-1} t
\Big)}{\exp\Big(
\epsilon(|\Omega |-1)^{-1} t
\Big)-1}- \frac{Q^{-1}
\sum_{k=1}^{t-1}
(2k-1)
\exp\Big(
\epsilon(|\Omega |-1)^{-1} k
\Big)}{\exp\Big(
\epsilon(|\Omega |-1)^{-1} 
\Big)-1} \\
&=
\frac{ (t-1)^2\exp\Big(
\epsilon(|\Omega |-1)^{-1} t
\Big)}{\exp\Big(
\epsilon(|\Omega |-1)^{-1} t
\Big)-1} - 
\frac{2\mathbb{E}(V^{(t)}_{\epsilon}) }{\exp\Big(
\epsilon(|\Omega |-1)^{-1} 
\Big)-1}\\
&+
\frac{\exp\Big(
\epsilon(|\Omega |-1)^{-1} t
\Big)-\exp\Big(
\epsilon(|\Omega |-1)^{-1}
\Big)}{\exp\Big(
\epsilon(|\Omega |-1)^{-1} t
\Big)-1}
\frac{1}{\exp\Big(
\epsilon(|\Omega  |-1)^{-1} 
\Big)-1}.
\end{align*}
It is easy to verify that $\lim_{\epsilon \rightarrow +\infty} \mathbb{E}(V^{(t)}_{\epsilon})^2 = (t-1)^2$. This completes the whole proof. \qed

\noindent
\textbf{Proof of Theorem \ref{Thm:ConsisPhi}}: Notice that the minimizer of $\mathcal{L}(\phi)$ lies in a discrete space. Therefore, it is impossible to derive an analytic form of $\widehat{\phi}$. Without loss of generality, we suppose the true ranking $\phi$ satisfies $\phi(I_i)=i$ for $i=1,\ldots,L$.

The proof we present here resembles that of Theorem 3.2 in \citet{tang2019mallows}. Let $\mathcal{Q}_1=\{\phi: \phi(I_1)=2,\phi(I_2)=1\}$ denote the set of rankings that item $I_1$ and $I_2$ exchange their ranks and $\mathcal{Q}_2=\{\phi: \phi(I_1)=1,\phi(I_2)=2\}$ denote the set of rankings that item $I_1$ and $I_2$ have correct ranks. Here it should be noted that the true ranking is an element of $\mathcal{Q}_2$. For any $\phi_1 \in \mathcal{Q}_1$, there exists an $\phi_2 \in \mathcal{Q}_2$ such that $\phi_1(I_l)=\phi_2(I_l)$ for $l \in \{3,\ldots,L\}$. Let $\Pi_{ij}(\mathcal{S}) = \sum_{l=1}^N I\big(\widetilde{\phi}_l(I_i)>\widetilde{\phi}_l(I_j) \big)$ denote the frequency that the rank of item $I_i$ is larger than that of item $I_j$. If $\Pi_{12}(\mathcal{S})>N/2$, it is easy to verify that 
\begin{align*}
\mathcal{L}(\phi_1) > \mathcal{L}(\phi_2).
\end{align*}
Therefore, it holds that 
$$
\mathcal{L}(\phi) \leq \max_{\phi_2 \in \mathcal{Q}_2}\mathcal{L}(\phi) < \max_{\phi_1 \in \mathcal{Q}_1}\mathcal{L}(\phi).
$$
Hence, $\Pi_{12}(\mathcal{S})>N/2$ implies that $\widehat{\phi} \neq \phi$.

\begin{align*}
\mathbb{P}_{\mathcal{S}}\Big(
\widehat{\phi}_{mle} \neq \phi
\Big) \geq 
\mathbb{P}_{\mathcal{S}}\Big(
\Pi_{12}(\mathcal{S})>N/2
\Big).
\end{align*}
Recall that, in Algorithm \ref{algorithm1}, the ranks of $I_1$ and $I_2$ are determined by the probability 
$$
\mathbb{P}\big(\widetilde{\phi}_l(I_2)<\widetilde{\phi}_l(I_1)\big)=\frac{1}{1+\exp(\epsilon(|\Omega|-1)^{-1})}, \mbox{ for } l=1,\ldots,N.
$$
For ease of notation, we denote $p_{\epsilon,\Omega} = \frac{\exp(\epsilon(|\Omega|-1)^{-1})}{1+\exp(\epsilon(|\Omega|-1)^{-1})}$. It then follows that
\begin{align*}
\mathbb{P}_{\mathcal{S}}\Big(
\Pi_{12}(\mathcal{S})>N/2
\Big) \geq& \mathbb{P}_{\mathcal{S}}\Big(
\Pi_{21}(\mathcal{S}) \leq \lfloor N/2 \rfloor 
\Big) 
= \sum_{k=0}^{\lfloor N/2 \rfloor }
{N \choose k}  p_{\epsilon,\Omega}^{k}(1-p_{\epsilon,\Omega})^{N-k}.
\end{align*}

By Theorem 2 of \citet{arratia1989tutorial}, there exists some positive constants $C_0$ such that
By approximating the binomial coefficient with Stirling's formula, we get
\begin{align}
\label{LOW}
\mathbb{P}_{\mathcal{S}}\Big(
\Pi_{12}(\mathcal{S})>N/2
\Big) \geq C_0 
\frac{\exp(\epsilon(|\Omega|-1)^{-1})+1}{\exp(\epsilon(|\Omega|-1)^{-1})-1}
\sqrt{
\frac{2}{\pi N}}
\exp\Big(
-N D_{KL}(1/2 \Vert p_{\epsilon,\Omega})
\Big),
\end{align}
where $D_{KL}(1/2 \Vert p_{\epsilon,\Omega})$ denotes the KL-divergence between two Bernoulli random variables with parameters $1/2$ and $p_{\epsilon,\Omega}$, respectively. Applying the reverse Pinsker inequality \citep{sason2015reverse} to the right-hand side of $(\ref{LOW})$, we further have
\begin{align*}
\mathbb{P}_{\mathcal{S}}\Big(
\Pi_{12}(\mathcal{S})>N/2
\Big) \geq &
C_0 
\frac{\exp(\epsilon(|\Omega|-1)^{-1})+1}{\exp(\epsilon(|\Omega|-1)^{-1})-1}
\sqrt{
\frac{2}{\pi N}}
\exp\Big\{
-2N \frac{\Big(\exp(\epsilon(|\Omega|-1)^{-1})-1\Big)^2}{1+\exp(\epsilon(|\Omega|-1)^{-1})}
\Big\}  \\
\geq &
C_0 
\frac{\exp(\epsilon(|\Omega|-1)^{-1})+1}{\exp(\epsilon(|\Omega|-1)^{-1})-1}
\sqrt{
\frac{2}{\pi N}}
\exp\Big\{
-2N \frac{\Big(\exp(\epsilon(|\Omega|-1)^{-1})-1\Big)^2}{2}
\Big\} \\
\geq &
C_0 
\frac{\exp(\epsilon(|\Omega|-1)^{-1})+1}{\exp(\epsilon(|\Omega|-1)^{-1})-1}
\sqrt{
\frac{2}{\pi N}}
\exp\Big\{
-N \Big(\exp(\epsilon(|\Omega|-1)^{-1})-1\Big)^2
\Big\}.
\end{align*}
Denote that $C(\epsilon) =\exp(\epsilon(|\Omega|-1)^{-1})-1$, the lower bound can be re-written as
\begin{align}
\label{LOWERBOUND}
\mathbb{P}_{\mathcal{S}}\Big(
\Pi_{12}(\mathcal{S})>N/2
\Big) \geq 
C_0\sqrt{
\frac{8}{C^2(\epsilon)\pi N}}
\exp\Big\{
-N C^2(\epsilon)
\Big\}.
\end{align}
Further, by setting $C(\epsilon)= O\big(N^{-1/2}\big)$, the right hand side of (\ref{LOWERBOUND}) is bounded away from 0 for any $N \geq 1$. The desired result immediately follows by seeing that $C(\epsilon)= O\big(N^{-1/2}\big)$ implies $\epsilon =O\big((|\Omega|-1)N^{-1/2}\big) $. \qed

\noindent
\textbf{Proof of Lemma \ref{LAPDP}}: Without loss of generality, we suppose the raw ranking $\phi$ is
\begin{align*}
\phi(\mathcal{I}_i) = i, i\in [m].
\end{align*}
Let $\widetilde{\phi}_{lap} =\mathcal{M}^{lap}_{\lambda}(\phi)$ and $\widetilde{\phi}_{lap}' =\mathcal{M}^{lap}_{\lambda}(\phi')$ be outputs of the Laplace mechanism. For any $\bm{r}\in \mathbb{R}^m$, the joint density functions of $\widetilde{\phi}_{lap} =\bm{r}$ and $\widetilde{\phi}_{lap}' =\bm{r}$ can be written as
\begin{align*}
&\mathbb{P}(\widetilde{\phi}_{lap} =\bm{r})=
\frac{1}{2^m\lambda^m}
\exp\Big(
-\frac{\sum_{i=1}^m |r_i-\phi(\mathcal{I}_i)|}{\lambda}
\Big), \\
&\mathbb{P}(\widetilde{\phi}_{lap}' =\bm{r})=
\frac{1}{2^m\lambda^m}
\exp\Big(
-\frac{\sum_{i=1}^m |r_i-\phi'(\mathcal{I}_i)|}{\lambda}
\Big).
\end{align*}
By the definition of neighboring ranking, we can easily verify that
\begin{align*}
\Big|
\sum_{i=1}^m |r_i-\phi(\mathcal{I}_i)|-
\sum_{i=1}^m |r_i-\phi'(\mathcal{I}_i)|
\Big| \leq 2(m-1),
\end{align*}
where the equality holds when $\phi'$ satisfies $\phi'(\mathcal{I}_i)=i+1$ for $i=1,\ldots,m-1$ and $\phi'(\mathcal{I}_1)=m$ and $r_i \geq m$ for $i =1,\ldots,m$. Therefore, by setting $\lambda=2(m-1)/\epsilon$, we have
\begin{align*}
\Big|
\log \frac{\mathbb{P}(\widetilde{\phi}_{lap} =\bm{r})}{\mathbb{P}(\widetilde{\phi}_{lap}' =\bm{r})}\Big|
\leq \epsilon.
\end{align*}
This completes the proof. \qed

\noindent
\textbf{Proof of Lemma \ref{RDPvLAP}}: Without loss of generality, we suppose that raw ranking $\phi$ satisfies that $\phi(\mathcal{I}_i) = i$ for $i=1,\ldots,m$.

Denote that $\widetilde{\phi}_{lap} = \mathcal{M}_{2(m-1)\epsilon^{-1}}^{lap}(\phi)$. For any $i>j$, we define a new random variable as $Z_{ij} = \widetilde{\phi}_{lap}(\mathcal{I}_i)-\widetilde{\phi}_{lap}(\mathcal{I}_j)$. By Corollary 2 of \citet{nadarajah2007linear}, the cumulative distribution function of $Z_{ij}$ takes the form as
\begin{align*}
F_{Z_{ij}}(z) = 
\begin{cases}
\frac{1}{2}
\exp\Big(\frac{\epsilon(z-i+j)}{2(m-1)}\Big)-
\frac{(z-i+j)\epsilon}{8(m-1)}\exp\Big(\frac{\epsilon(z-i+j)}{2(m-1)}\Big),
z< i - j,\\
1-\frac{1}{2}
\exp\Big(-\frac{\epsilon(z-i+j)}{2(m-1)}\Big)+
\frac{(z-i+j)\epsilon}{8(m-1)}\exp\Big(-\frac{\epsilon(z-i+j)}{2(m-1)}\Big)
z>i-j.
\end{cases}
\end{align*}
Therefore, we have
\begin{align*}
&\mathbb{P}\Big(
\widetilde{\phi}_{lap}(\mathcal{I}_i)
>
\widetilde{\phi}_{lap}(\mathcal{I}_j)
\Big) = \mathbb{P}\Big(Z_{ij}>0\Big) = 1 - F_{Z_{ij}}(0) \\
=&1-
\frac{1}{2}
\exp\Big(-\frac{\epsilon(i-j)}{2(m-1)}\Big)-
\frac{(i-j)\epsilon}{8(m-1)}\exp\Big(-\frac{\epsilon(i-j)}{2(m-1)}\Big).
\end{align*}
Let $U_{\epsilon}^{t-1}=\mathbb{E}\big[\sum_{j=1}^{t-1}I(\widetilde{\phi}_{lap}(\mathcal{I}_t)
>
\widetilde{\phi}_{lap}(\mathcal{I}_j))\big]$ denote the expected number of correct partial orders of item $t$ and those items with lower ranks. It then follows that for any $t \geq 2$,
\begin{align*}
\mathbb{E}\big[
U_{\epsilon}^{t-1}
\big]=
&\sum_{j=1}^{t-1}
\mathbb{P}\Big(
\widetilde{\phi}_{lap}(\mathcal{I}_t)
>
\widetilde{\phi}_{lap}(\mathcal{I}_j)\Big)\\
=&
t-1 - \frac{\exp(-t\lambda_{\epsilon})}{2}\sum_{j=1}^{t-1}\exp\big(j\lambda_{\epsilon}\big)-
\sum_{j=1}^{t-1}\frac{j\lambda_{\epsilon}}{4}\exp\big(-j\lambda_{\epsilon}\big) \\
=&
t-1- \frac{1-\exp(-(t-1)\lambda_{\epsilon})}{2(\exp(\lambda_{\epsilon})-1)}+
\frac{\lambda_{\epsilon}(t-1)\exp(-(t-1)\lambda_{\epsilon})}{4(\exp(\lambda_{\epsilon})-1)}
-
\lambda_{\epsilon}
\frac{\exp(-\lambda_{\epsilon})-\exp(-t\lambda_{\epsilon})}{4(1-\exp(-\lambda_{\epsilon}))^2} \\
=&
t-1- \frac{1-p_{\epsilon}^{-(t-1)}}{2(p_{\epsilon}-1)}+
\frac{\lambda_{\epsilon}(t-1)p_{\epsilon}^{-(t-1)}}{4(p_{\epsilon}-1)}
-
\lambda_{\epsilon}
\frac{p_{\epsilon}-p_{\epsilon}^{-(t-2)}}{4(p_{\epsilon}-1)^2} \\
=&
t-1- \frac{p_{\epsilon}^{t-1}-1}{2(p_{\epsilon}^t-p_{\epsilon}^{t-1})}+
\frac{\lambda_{\epsilon}(t-1)}{4(p_{\epsilon}^t-p_{\epsilon}^{t-1})}
-
\lambda_{\epsilon}
\frac{p_{\epsilon}-p_{\epsilon}^{-(t-2)}}{4(p_{\epsilon}-1)^2}
\end{align*}
where $p_{\epsilon} = \exp(\lambda_{\epsilon})$ and $\lambda_{\epsilon} =\epsilon2^{-1}(m-1)^{-1}$. By property (3) of Theorem \ref{Thm:ConsisandPriv}, we have
\begin{align*}
\mathbb{E}\big[V_{\epsilon}^{t-1}\big] = 
t-1 + \frac{t}{q_{\epsilon}^{t}-1} - \frac{1}{q_{\epsilon}-1},
\end{align*}
where $q_{\epsilon} = p_{\epsilon}^2$.

Next, we turn to prove $\mathbb{E}\big[V_{\epsilon}^{t-1}\big] - \mathbb{E}\big[
U_{\epsilon}^{t-1}
\big]>0$ for any $2 \leq t \leq M$ and $\epsilon >0$.
\begin{align*}
\mathbb{E}\big[V_{\epsilon}^{t-1}\big] - \mathbb{E}\big[
U_{\epsilon}^{t-1}
\big] = 
 \frac{t}{q_{\epsilon}^{t}-1} - \frac{1}{q_{\epsilon}-1}
+ \frac{p_{\epsilon}^{t-1}-1}{2(p_{\epsilon}^t-p_{\epsilon}^{t-1})}-
\frac{\lambda_{\epsilon}(t-1)}{4(p_{\epsilon}^t-p_{\epsilon}^{t-1})}
+
\lambda_{\epsilon}
\frac{p_{\epsilon}-p_{\epsilon}^{-(t-2)}}{4(p_{\epsilon}-1)^2}.
\end{align*}
Notice that 
\begin{align*}
-
\frac{\lambda_{\epsilon}(t-1)}{4(p_{\epsilon}^t-p_{\epsilon}^{t-1})}
+
\lambda_{\epsilon}
\frac{p_{\epsilon}-p_{\epsilon}^{-(t-2)}}{4(p_{\epsilon}-1)^2} =
\frac{\lambda_{\epsilon}(p_{\epsilon}^t-t p_{\epsilon}+t-1)}{4(p_{\epsilon}^t-p_{\epsilon}^{t-1})(p_{\epsilon}-1)} \geq 0,
\end{align*}
for any $p_{\epsilon} \geq 1$. Then, when $p_{\epsilon}\geq 2$, we have
\begin{align*}
\mathbb{E}\big[V_{\epsilon}^{t-1}\big] - \mathbb{E}\big[
U_{\epsilon}^{t-1}
\big] \geq  \frac{t}{q_{\epsilon}^{t}-1} +\frac{p_{\epsilon}^{t-1}(p_{\epsilon}-1)-(p_{\epsilon}+1)}{2(p_{\epsilon}+1)(p_{\epsilon}^t-p_{\epsilon}^{t-1})}>0.
\end{align*}
Next, we consider the case that $p_{\epsilon} <2$. By the fact that $3/2\lambda_{\epsilon} \geq p_{\epsilon}-1$ when $p_{\epsilon} <2$, we get
\begin{align*}
\frac{\lambda_{\epsilon}(p_{\epsilon}^t-t p_{\epsilon}+t-1)}{4(p_{\epsilon}^t-p_{\epsilon}^{t-1})(p_{\epsilon}-1)} \geq \frac{(p_{\epsilon}^t-t p_{\epsilon}+t-1)}{6(p_{\epsilon}^t-p_{\epsilon}^{t-1})}.
\end{align*}
Next, we analyze the ratio 
\begin{align}
\label{RA}
Ra(p_{\epsilon},t) = &
\frac{\frac{1}{p_{\epsilon}^2-1}- \frac{t}{p_{\epsilon}^{2t}-1}}{ \frac{p_{\epsilon}^{t-1}-1}{2(p_{\epsilon}^t-p_{\epsilon}^{t-1})}+
\frac{(p_{\epsilon}^t-t p_{\epsilon}+t-1)}{6(p_{\epsilon}^t-p_{\epsilon}^{t-1})}} =
\frac{6(p_{\epsilon}^{2t}-tp_{\epsilon}^{2}+t-1)(p_{\epsilon}^t-p_{\epsilon}^{t-1})}{(p_{\epsilon}^t+3p_{\epsilon}^{t-1}-t p_{\epsilon}+t-4)(p_{\epsilon}^2-1)(p_{\epsilon}^{2t}-1)}\notag  \\
=&
\frac{6(p_{\epsilon}^{2t}-tp_{\epsilon}^{2}+t-1)p_{\epsilon}^{t-1}}{(p_{\epsilon}^t+3p_{\epsilon}^{t-1}-t p_{\epsilon}+t-4)(p_{\epsilon}+1)(p_{\epsilon}^{2t}-1)} \notag \\
=&
\frac{6p_{\epsilon}^{t-1}}{(1+p_{\epsilon})(p_{\epsilon}^{t}+1)}
\frac{(p_{\epsilon}^{2t}-tp_{\epsilon}^{2}+t-1)}{
p_{\epsilon}^{2t}+3p_{\epsilon}^{2t-1}-t p_{\epsilon}^{t+1}+(t-5)p_{\epsilon}^t-3p_{\epsilon}^{t-1}+t p_{\epsilon}-t+4}.
\end{align}
Using L'Hospital's rule, we have $\lim_{p_{\epsilon}\rightarrow 1}Ra(p_{\epsilon},t)=1$ for any $t \geq 2$. Furthermore, it can easily verified that the numerator and denominator of (\ref{RA}) are both positive and increasing on $p_{\epsilon} \in [1,2]$ for any $t \geq 2$ and the denominator is larger than the numerator for any $p_{\epsilon} \in [1,2]$ and $t \geq 2$. Therefore, it follows that $Ra(p_{\epsilon},t) \leq 1$, which implies that
\begin{align*}
\mathbb{E}\big[V_{\epsilon}^{t-1}\big] - \mathbb{E}\big[
U_{\epsilon}^{t-1}
\big] \geq 0,
\end{align*}
where the equality holds if and only if $p_{\epsilon}=1$ indicating $\epsilon=0$.

To sum up, we get
\begin{align*}
\mathbb{E}\Big[
T\big(\phi,\mathcal{A}_{\epsilon}(\phi)\big)\Big] = 2\sum_{t=2}^{m}\mathbb{E}[V_{\epsilon}^{t-1}] >
2\sum_{t=2}^{m}\mathbb{E}[U_{\epsilon}^{t-1}] =\mathbb{E}\Big[
T\big(\phi,\mathcal{M}^{lap}_{2(m-1)\epsilon^{-1}}(\phi)\big)\Big],
\end{align*}
for any $\epsilon>0$. This completes the proof. \qed

\noindent
\textbf{Proof of Lemma \ref{Invariant_syn}}: By property (1) of Theorem \ref{Thm:ConsisandPriv}, we get
\begin{align*}
\Big(\mathbb{P}\big(\Phi_u(\mathcal{I}_i)>\Phi(\mathcal{I}_j)\big)-\mathbb{P}\big(\Phi_u(\mathcal{I}_i)<\Phi(\mathcal{I}_j)\big)\Big)\Big(\mathbb{P}\big(\widetilde{\Phi}_u(\mathcal{I}_i)>\widetilde{\Phi}_u(\mathcal{I}_j)\big)-\mathbb{P}\big(\widetilde{\Phi}_u(\mathcal{I}_i)<\widetilde{\Phi}_u(\mathcal{I}_j)\big)\Big)>0.
\end{align*}
This combined with Lemma \ref{Bayes_ranker} yields that
\begin{equation}
    \Big(f^*(\bm{x}_u,\bm{y}_{i})-f^*(\bm{x}_u,\bm{y}_{j})\Big)
    \Big(\widetilde{f}^*(\bm{x}_u,\bm{y}_{i})-\widetilde{f}^*(\bm{x}_u,\bm{y}_{j})\Big)
> 0,
\end{equation}
for any $u \in [n]$ and $i,j \in [m]$. By the definitions of $E_{uij}(f)$ and $\widetilde{E}_{uij}(f)$, we have
\begin{align*}
E_{uij}(f) - E_{uij}^* = |2\eta_{uij}-1|I\Big(
    \big(f(\bm{x}_u,\bm{y}_{i})-f(\bm{x}_u,\bm{y}_{j})\big)
    \big(f^*(\bm{x}_u,\bm{y}_{i})-f^*(\bm{x}_u,\bm{y}_{j})\big)>0
    \Big), \\
    \widetilde{E}_{uij}(f) - \widetilde{E}_{uij}^* = |2\widetilde{\eta}_{uij}-1|I\Big(
    \big(f(\bm{x}_u,\bm{y}_{i})-f(\bm{x}_u,\bm{y}_{j})\big)
    \big(\widetilde{f}^*(\bm{x}_u,\bm{y}_{i})-\widetilde{f}^*(\bm{x}_u,\bm{y}_{j})\big)>0
    \Big).
\end{align*}
The desired result immediately follows from property (2) of Theorem \ref{Thm:ConsisandPriv}. \qed

\noindent
\textbf{Proof of Lemma \ref{lemma:low-noise}}: By the property (1) in Theorem \ref{Thm:ConsisandPriv}, we have
\begin{align*}
 \frac{\exp\big((|\Omega_u|-1)^{-1}\epsilon_u\big)-1}{\exp\big((|\Omega_u|-1)^{-1}\epsilon_u\big)+1} \leq 
\frac{|2\widetilde{\eta}_{uij}-1|}{|2\eta_{uij}-1|} \leq 
\frac{\exp\big(\epsilon_u\big)-1}{\exp\big(\epsilon_u\big)+1}.
\end{align*}
Let $\Psi_u =  \frac{\exp\big((|\Omega_u|-1)^{-1}\epsilon_u\big)+1}{\exp\big((|\Omega_u|-1)^{-1}\epsilon_u\big)-1}$, then it is straightforward to see that $|2\widetilde{\eta}_{uij}-1| \leq \beta$ implies $|2\eta_{uij}-1| \leq \Psi_u \beta$. Following from Assumption C yields that
\begin{align*}
\mathbb{P}\Big(
|2\widetilde{\eta}_{uij}-1| \leq \beta\Big)
\leq 
\mathbb{P}\Big(
|2\eta_{uij}-1| \leq \Psi_u\beta\Big) \leq 
C_1 (\Psi_u)^{\gamma}\beta^{\gamma}.
\end{align*}
This completes the proof. \qed

\begin{lemma}
\label{Var_Mean_inequ}
Under Assumption \ref{Assum:low-noise}, there exists some positive constants $C_4>0$ such that 
$$
\text{Var}\big( G_f(u) - G_{f^*}(u)\big) \leq 2 C_4
\mathbb{E}^{\frac{1}{1+\gamma}}\big[
\Psi_{u}^{\gamma}\big]
\big(e_{\upsilon}(f,f^*)\big)^{\gamma/(\gamma+1)}.
$$
\end{lemma}

\noindent
\textbf{Proof of Lemma \ref{Var_Mean_inequ}} :
By the law of total variance, we have
\begin{align*}
\text{Var}\big( G_f(u) - G_{f^*}(u)\big)  = 
\mathbb{E}\big[
\text{Var}_u\big( G_f(u) - G_{f^*}(u)\big)\big]+
\text{Var}\big[
\mathbb{E}_u\big( G_f(u) - G_{f^*}(u)\big)\big],
\end{align*}
where $\text{Var}_u\big( G_f(u) - G_{f^*}(u)\big)$ is the conditional variance of $G_f(u)-G_{f^*}(u)$ with fixed user $u$ and $\mathbb{E}_u\big( G_f(u) - G_{f^*}(u)\big)$ is the taken with respect to synthetic rankings conditional on user $u$.

By assuming items are uniformly generated with fixed size, $\text{Var}_u\big( G_f(u) - G_{f^*}(u)\big)$ can be upper bounded as
\begin{align*}
&\text{Var}_u\big( G_f(u) - G_{f^*}(u)\big) \\
 =& \frac{1}{m^2(m-1)^2}
\text{Var}\Big(
\sum_{ i\neq j}
I(\widetilde{\Phi}_u(\mathcal{I}_i)>\widetilde{\Phi}_u(\mathcal{I}_j)))\overline{\upsilon}\big( f(\bm{x}_u,\bm{y}_i)-f(f(\bm{x}_u,\bm{y}_j)\big)
\Big) \\
\leq &
\frac{1}{m(m-1)^2}
 \sum_{ i\neq j}
 \text{Var}\Big(
I(\widetilde{\Phi}_u(\mathcal{I}_i)>\widetilde{\Phi}_u(\mathcal{I}_j)))\overline{\upsilon}\big( f(\bm{x}_u,\bm{y}_i)-f(\bm{x}_i,\bm{y}_j)\big)
\Big) ,
\end{align*}
where $\overline{\upsilon}(f(\bm{x}_u,\bm{y}_i)-f(\bm{x}_u,\bm{y}_j)) = \upsilon(f(\bm{x}_u,\bm{y}_i)-f(\bm{x}_u,\bm{y}_j)) -\upsilon(f^*(\bm{x}_i,\bm{y}_i)-f^*(\bm{x}_i,\bm{y}_j))$ and the inequality follows from the fact that partial orders are partially correlated.

For any $i,j\in [m]$, we have
\begin{align*}
 &\text{Var}\Big(
I(\widetilde{\Phi}_u(\mathcal{I}_i)>\widetilde{\Phi}_u(\mathcal{I}_j))\overline{\upsilon}\big( f(\bm{x}_u,\bm{y}_i)-f(\bm{x}_u,\bm{y}_j)\big)
\Big) \\
\leq &
\mathbb{P}\big( \widetilde{\Phi}_u(\mathcal{I}_i)>\widetilde{\Phi}_u(\mathcal{I}_j)\big)
\overline{\upsilon}^2\big( f(\bm{x}_u,\bm{y}_i)-f(\bm{x}_u,\bm{y}_j)\big) \\
&+
\mathbb{P}\big( \widetilde{\Phi}_u(\mathcal{I}_j)>\widetilde{\Phi}_u(\mathcal{I}_i)\big)
\overline{\upsilon}^2\big( f(\bm{x}_u,\bm{y}_j)-f(\bm{x}_u,\bm{y}_i)\big) \\
=&
\widetilde{\eta}_{uij}\overline{\upsilon}^2\big( f(\bm{x}_u,\bm{y}_i)-f(\bm{x}_u,\bm{y}_j)\big) +
(1-\widetilde{\eta}_{uij})
\overline{\upsilon}^2\big( f(\bm{x}_u,\bm{y}_j)-f(\bm{x}_u,\bm{y}_i)\big) \\
\leq &
 K_{\upsilon}^2 \Big[  f(\bm{x}_u,\bm{y}_i)-f(\bm{x}_i,\bm{y}_j) -  f^*(\bm{x}_u,\bm{y}_i)+f^*(\bm{x}_u,\bm{y}_j)\Big]^2,
\end{align*}
where the last inequality follows from the Lipschitz continuity of $\upsilon$. For ease of notation, we let $g_{f}(u,i,j) =   f(\bm{x}_u,\bm{y}_i)-f(\bm{x}_u,\bm{y}_j)$. By the fact that $f^*$ is the optimal minimizer, then for each user $u$, $\mathbb{E}_u\big( G_f(u) - G_{f^*}(u)\big)$ can be lower-bounded as
\begin{align}
\label{Low-noise_inequ}
&\mathbb{E}_{u}\big( G_f(u) - G_{f^*}(u)\big)\\
= &
\frac{1}{m(m-1)}\sum_{i <j} 
\widetilde{\eta}_{uij} \overline{\phi}\big( f(\bm{x}_u,\bm{y}_i)-f(\bm{x}_u,\bm{y}_j)\big)+
(1-\widetilde{\eta}_{uij})\overline{\phi}\big( f(\bm{x}_u,\bm{y}_j)-f(\bm{x}_u,\bm{y}_i)\big)\notag \\
\geq &
\frac{T_1}{m(m-1)}
\sum_{i <j} \big(  f(\bm{x}_u,\bm{y}_i)-f(\bm{x}_u,\bm{y}_j) -  f^*(\bm{x}_u,\bm{y}_i)+f^*(\bm{x}_u,\bm{y}_j))^2|1-2\widetilde{\eta}_{uij}| \notag \\
\geq &
T_1 \mathbb{E}_u
 \big[(
g_f(u,i,j) - g_{f^*}(u,i,j))^2 \beta I(|1-2\widetilde{\eta}_{uij}| \geq \beta)\big],
\end{align}
for some positive constants $T_1>0$ and the first inequality follows from the fact that $f^*$ is the optimal minimizer and the inequality holds when $\widetilde{\eta}_{uij}=1/2$ almost surely or $f=f^*$.

Combining $(\ref{Low-noise_inequ})$ and Lemma \ref{lemma:low-noise}, it follows that 
\begin{align*}
\mathbb{E}_u\big( G_f(u) - G_{f^*}(u)\big) \geq&
T_1 \beta \mathbb{E}_u\big[(
g_f(u,i,j) - g_{f^*}(u,i,j))^2 \big] - 
C_2 \Psi_{u}^{\gamma} \beta^{\gamma+1} \\
= &
T_1 (\frac{T_2}{2C_2})^{\frac{\gamma+1}{\gamma}} 
 \mathbb{E}^{\frac{\gamma+1}{\gamma}}_u\big[(
g_f(u,i,j) - g_{f^*}(u,i,j))^2\big] \Psi_{u}^{-1},
\end{align*}
where the last equality follows by taking $\beta=(2^{-1}T_1C_2^{-1})^{\frac{1}{\gamma}}\Psi_u^{-1}  \mathbb{E}^{\frac{1}{\gamma}}\big[(
g_f(u,i,j) - g_{f^*}(u,i,j))^2\big]$. Therefore, it holds that for some positive constants $C_4$ depending on $\gamma$,
\begin{align*}
\text{Var}_u\big( G_f(u) - G_{f^*}(u)\big) \leq  
\frac{C_4}{|\Omega|-1}
\Psi_{u}^{\frac{\gamma}{\gamma+1}} 
\mathbb{E}_u^{\frac{\gamma}{\gamma+1}}\big( G_f(u) - G_{f^*}(u)\big).
\end{align*}
Taking the expectation of both sides yields that
\begin{align}
\label{Var_first}
\mathbb{E}\big[
\text{Var}_u\big( G_f(u) - G_{f^*}(u)\big) 
\big] \leq &C_4 \mathbb{E}\big[
(|\Omega|-1)^{-1}
\Psi_{u}^{\frac{\gamma}{\gamma+1}} 
\mathbb{E}_u^{\frac{\gamma}{\gamma+1}}\big( G_f(u) - G_{f^*}(u)\big)
\big] \notag \\
\leq &\frac{C_4}{|\Omega|-1}
\mathbb{E}^{\frac{1}{1+\gamma}}\big[
\Psi_{u}^{\gamma}\big]
\big(e_{\upsilon}(f,f^*)\big)^{\frac{\gamma}{\gamma+1}},
\end{align} 
where the last inequality follows from the H\"{o}lder's inequality.

Next, we proceed to establish the relation between $\text{Var}\big[
\mathbb{E}_u\big( G_f(u) - G_{f^*}(u)\big)\big]$ and $e_{\upsilon}(f,f^*)$. 
\begin{align}
\label{Var_second}
&\text{Var}\big[
\mathbb{E}_u\big( G_f(u) - G_{f^*}(u)\big)\big] = 
\mathbb{E}\Big(\mathbb{E}_u^2\big( G_f(u) - G_{f^*}(u)\big)\Big) - e_{\upsilon}^2(f,f^*)\notag \\
=&\mathbb{E}\Big(
\frac{1}{m(m-1)}\sum_{i<j} 
\widetilde{\eta}_{uij} \overline{\upsilon}\big( f(\bm{x}_u,\bm{y}_i)-f(\bm{x}_u,\bm{y}_j)\big)+
(1-\widetilde{\eta}_{uij})\overline{\upsilon}\big( f(\bm{x}_u,\bm{y}_j)-f(\bm{x}_u,\bm{y}_i)\big)\Big)^2 - e_{\upsilon}^2(f,f^*)\notag \\
\leq &
\mathbb{E}\Big(
\frac{1}{m(m-1)}\sum_{i<j} 
\widetilde{\eta}_{uij} \overline{\upsilon}^2\big( f(\bm{x}_u,\bm{y}_i)-f(\bm{x}_u,\bm{y}_j)\big)+
(1-\widetilde{\eta}_{uij})\overline{\upsilon}^2\big( f(\bm{x}_u,\bm{y}_j)-f(\bm{x}_u,\bm{y}_i)\big) \Big)-e_{\upsilon}^2(f,f^*) 
\notag \\
\leq & 
C_4
\mathbb{E}^{\frac{1}{1+\gamma}}\big[
\Psi_{u}^{\gamma}\big]
\big(e_{\upsilon}(f,f^*)\big)^{\gamma/(\gamma+1)},
\end{align}
where the last inequality by applying the same steps as above. Combining $(\ref{Var_first})$ and $(\ref{Var_second})$, we have
\begin{align*}
\text{Var}\big( G_f(u) - G_{f^*}(u)\big) \leq 2 C_4
\mathbb{E}^{\frac{1}{1+\gamma}}\big[
\Psi_{u}^{\gamma}\big]
\big(e_{\upsilon}(f,f^*)\big)^{\frac{\gamma}{\gamma+1}}.
\end{align*}
This completes the proof. \qed

\begin{lemma}
\label{Bound_Mean}
Provided that $\big(n^{-1}|\Theta| \log  \big(|\Theta|^{-1}n )\big)^{\frac{\gamma+1}{\gamma+2}} \mathbb{E}^{\frac{1}{\gamma+2}}\big(\Psi_u^{\gamma}\big)=O(\delta_n)$, it holds that for $0 \leq i \leq i_0 $ and $0\leq j \leq j_0$,
\begin{equation}
\label{Equ:Bound_Mean}
\mathbb{E}\Big(\sup_{f \in \mathcal{H}_{ij}} 
U_n(f)\Big) \leq 1/2 M(i,j),
\end{equation}
where $U_n(f) =n^{-1}\sum_{u=1}^n \big[DG_f(u) - \mathbb{E} \big(DG_f(u)\big) \big]$.
\end{lemma}

\noindent
\textbf{Proof of Lemma \ref{Bound_Mean}}:
The main idea to bound $\mathbb{E}(\sup_{f \in \mathcal{H}_{ij}} 
U_n(f))$ is based on Rademacher complexity. First, by a symmetrization argument, we have
\begin{align*}
&\mathbb{E}\Big(\sup_{f \in \mathcal{H}_{ij}} 
U_n(f)\Big) = \mathbb{E}\Big(\sup_{f \in \mathcal{H}_{ij}} 
\frac{1}{n}\sum_{u=1}^n \big(DG_f(u)\big) - \mathbb{E} \big(DG_f(u)\big) \Big) \\
=&
 \mathbb{E}_{\Omega}\Big[\sup_{f \in \mathcal{H}_{ij}} \mathbb{E}_{\Omega'}\Big(
\frac{1}{n}\sum_{u,u'} \big(DG_f(u)-  DG_f(u') \big)\Big| \Omega \Big) 
\Big] 
 \leq 
 \mathbb{E}_{\mathcal{U},\mathcal{U}'}\Big[\sup_{f \in \mathcal{H}_{ij}} 
\frac{1}{n}\sum_{u,u'} \big(DG_f(u)-  DG_f(u') \big) 
\Big]  \\
= &
 \mathbb{E}_{\mathcal{U},\mathcal{U}'}\Big[\sup_{f \in \mathcal{H}_{ij}} \mathbb{E}_{\bm{\upsilon}}\Big(
\frac{1}{n}\sum_{u,u'} \sigma_u  \big(DG_f(u) -  DG_f(u') \big) \Big)
\Big] \leq  2
\mathbb{E}_{\mathcal{U}} \Big[\frac{1}{n}\mathbb{E}_{\bm{\sigma}}\Big(\sup_{f \in \mathcal{H}_{ij}} 
\sum_{u=1}^n \sigma_u DG_f(u) \Big)
\Big]\\
 =& 2 \mathbb{E}_{\mathcal{U}}[\mathcal{R}_{n}(\mathcal{H}_{ij})],
\end{align*}
where the first inequality follows from Jensen's inequality, $(\sigma_u)_{u=1}^n$ are independent Rademacher random variables, $\mathcal{U}$ denote the set of users, $\mathcal{U}'$ is an independent copy of $\mathcal{U}$, and $ \mathbb{E}[\mathcal{R}_{n}(\mathcal{H}_{ij})]$ is the Rademacher complexity of $\mathcal{H}_{ij}$.

Next, we turn to bound $\mathbb{E}[\mathcal{R}_{n}(\mathcal{H}_{ij})]$. By Theorem 3.11 in \citet{koltchinskii2011oracle}, there exists some constants $A_1>0$ such that 
\begin{align}
\label{Entropy}
\mathbb{E}[\mathcal{R}_{n}(\mathcal{H}_{ij})] \leq 
A_1 n^{-1/2} \mathbb{E} \Big[
\int_{0}^{2\sigma_n(\mathcal{H}_{ij})} \sqrt{\log \mathcal{N}(\mathcal{H}_{ij},v,\Vert \cdot \Vert_{\mathcal{U}})}dv \Big],
\end{align}
where $\Vert f \Vert_{\mathcal{U}} =\sqrt{n^{-1}\sum_{u=1}^n DG_f^2(u)}$ and $\sigma_n^2(\mathcal{H}_{ij}) = \sup_{f \in \mathcal{H}_{ij}}n^{-1}\sum_{u=1}^n DG_f^2(u)$. Next, we derive an upper bound for $\log \mathcal{N}(\mathcal{H}_{ij},v,\Vert \cdot \Vert_{\mathcal{U}})$. For any $f_1,f_2 \in \mathcal{H}$, 
\begin{align*}
\Vert f_1 - f_2 \Vert_{\mathcal{U}}^2 = &
\frac{1}{n}\sum_{u=1}^n \Big(
DG_{f_1}(u) - DG_{f_2}(u) 
\Big)^2 = \frac{1}{n}\sum_{u=1}^n \Big(
G_{f_1}(u) - G_{f_2}(u) 
\Big)^2 \\
\leq & 
\frac{1}{n}\sum_{u=1}^n
\frac{1}{m(m-1)}\sum_{i \neq j} K_{\upsilon}^2\big(g_{f_1}(u,i,j) - g_{f_2}(u,i,j)\big)^2 \\
\leq & 
\frac{1}{n}\sum_{u=1}^n
\frac{1}{m(m-1)}\sum_{i \neq j} K_{\upsilon}^2\Big(2\big(f_1(\bm{x}_u,\bm{y}_i)-f_2(\bm{x}_u,\bm{y}_i)\big)^2+2
\big(f_1(\bm{x}_u,\bm{y}_j)-f_2(\bm{x}_u,\bm{y}_j)\big)^2 \Big) \\
\leq &
\frac{1}{n}\sum_{u=1}^n
\frac{1}{m(m-1)}\sum_{i \neq j} K_{\upsilon}^2\big(f_1(\bm{x}_u,\bm{y}_i)-f_2(\bm{x}_u,\bm{y}_i)\big)^2  \\
\leq&   
\frac{C_{1}}{n}\sum_{u=1}^n \Big(
\Vert \bm{x}_u \Vert_2^2+ \frac{1}{m} \sum_{i \in [m]} \Vert \bm{y}_i \Vert_2^2 \Big)
 \Vert 
 \Theta_{f_1} - \Theta_{f_2} \Vert_{\infty}^2,
\end{align*}
where the first inequality follows from the fact that $\phi(\cdot)$ is a $K_{\upsilon}$-Lipschitz function and the last inequality follows from Assumption 3. 

For ease of notation, we denote that $F^2(u) = \Vert \bm{x}_u \Vert_2^2+  \frac{1}{m} \sum_{i \in [m]}  \Vert \bm{y}_i \Vert_2^2$. Notice that $f \in \mathcal{H}_{ij}$ implies that $\Vert \Theta_f \Vert_{\infty} \leq 2^{j} J_0$. Let $B_{\Theta}(2^{j} J_0)$ denote the cube centered at the origin and of length $2^{j+1} J_0$ in $\mathbb{R}^{|\Theta|}$. It then can be verified that
\begin{align*}
\mathcal{N}(\mathcal{H}_{ij},v,\Vert \cdot \Vert_{\mathcal{U}}) 
\leq 
\mathcal{N}\Big(B_{\Theta}(2^{j} J_0),\frac{v}{\sqrt{C_1} \Vert F \Vert},\Vert \cdot \Vert\Big),
\end{align*}
where $\Vert F \Vert = \sqrt{n^{-1}\sum_{u=1}^n F^2(u)}$. It then follows that
\begin{align}
\label{Upper_Bound_Entropy}
\log \mathcal{N}(\mathcal{H}_{ij},v,\Vert \cdot \Vert_{\mathcal{U}}) \leq 
|\Theta| 
\log \Big( \max\big\{ \frac{\sqrt{C_1} 2^{j+1} J_0 \Vert F \Vert}{v}, 1\big\}\Big),
\end{align}
Next, combining $(\ref{Entropy})$ and $(\ref{Upper_Bound_Entropy})$ yields that
\begin{align*}
&\mathbb{E}[\mathcal{R}_{n}(\mathcal{H}_{ij})] \leq 
A_1 n^{-1/2} \mathbb{E} \Big[
\int_{0}^{2\sigma_n(\mathcal{H}_{ij})} \sqrt{|\Theta| 
\log \Big( \max\Big\{\frac{\sqrt{C_1} 2^{j+1} J_0\Vert F \Vert}{v},1\Big\}\Big)}dv \Big] \\
=& A_1 n^{-1/2} \mathbb{E} \Big[
\int_{0}^{2\sigma_n(\mathcal{H}_{ij})} \sqrt{|\Theta| 
\log \Big( \max\Big\{\frac{\sqrt{C_1} 2^{j+1} J_0\Vert F \Vert}{v},1\Big\}\Big)}dv I\big(\Vert F \Vert^2 > 2\mathbb{E}F^2(u)\big)\Big] \\
+&
A_1 n^{-1/2} \mathbb{E} \Big[
\int_{0}^{2\sigma_n(\mathcal{H}_{ij})} \sqrt{|\Theta| 
\log \Big( \max\Big\{\frac{\sqrt{C_1} 2^{j+1} J_0\Vert F \Vert}{v},1\Big\}\Big)}dv I\big(\Vert F \Vert^2 \leq 2\mathbb{E}F^2(u)\big)\Big] \\
=& V_1 + V_2.
\end{align*}
Next, it suffices to bound $V_1$ and $V_2$ separately. For $V_1$, we first note that
\begin{align*}
&\int_{0}^{2\sigma_n(\mathcal{H}_{ij})} \sqrt{|\Theta| 
\log \Big( \max\Big\{\frac{\sqrt{C_1} 2^{j+1} J_0\Vert F \Vert}{v},1\Big\}\Big)}dv \\
 =&\sqrt{C_1} 2^{j+1} J_0\Vert F \Vert
\int_{0}^{2\frac{\sigma_n(\mathcal{H}_{ij})}{\sqrt{C_1} 2^{j+1} J_0\Vert F \Vert}}
\sqrt{|\Theta| 
\log \Big( \max\Big\{\frac{1}{v},1\Big\}\Big)}d\epsilon  \\
\leq & 
\sqrt{C_1} 2^{j+1} J_0\Vert F \Vert
\int_{0}^{1}
\sqrt{|\Theta| 
\log \big( \epsilon^{-1}\big)}d\epsilon  \\
\leq& 
\sqrt{C_1} 2^{j+1} J_0\Vert F \Vert \sqrt{|\Theta|}
\int_{1}^{\infty} u^{-2} \sqrt{\log (u)}du \\
\leq &A_2 2^{j+1} J_0\Vert F \Vert \sqrt{|\Theta|},
\end{align*}
where $A_2 = \sqrt{C_1} \int_{0}^{\infty} u^{-3/2}du$ and the last inequality follows from the fact that $\sqrt{\log(u)} \leq \sqrt{u-1}$ for $u \geq 1$. With this, $V_1$ can be bounded as
\begin{align}
\label{Bound_V1}
V_1 \leq &A_1 n^{-1/2} \mathbb{E}\Big[
A_2 2^{j+1} J_0\Vert F \Vert\sqrt{|\Theta|} 
I\big(\Vert F \Vert^2 > 2\mathbb{E}F^2(u)\big)
\Big] \notag \\
\leq &
A_1A_2  2^{j+1} J_0 n^{-1/2} \mathbb{E}\Big[
\Vert F \Vert \sqrt{|\Theta|} 
I\big(\Vert F \Vert^2 > 2\mathbb{E}F^2(u)\big)
\Big]\notag \\
\leq &A_1A_2  2^{j+1} J_0 n^{-1/2}\sqrt{|\Theta| \mathbb{E}F^2(u)} 
\mathbb{P}\Big(
\Vert F \Vert^2 > 2\mathbb{E}F^2(u)
\Big)\notag \\
\leq & 
A_1A_2  2^{j_0+1} J_0 n^{-1/2}\sqrt{|\Theta| \mathbb{E}F^2(u)} 
\mathbb{P}\Big(
\Vert F \Vert^2 > 2\mathbb{E}F^2(u)
\Big) 
\end{align}
where the third inequality follows from H\"{o}lder's inequality. By Bernstein's inequality, the right-hand side of $(\ref{Bound_V1})$ can be bounded as
\begin{align}
\label{Bern}
\mathbb{P}\Big(
\Vert F \Vert^2 > 2\mathbb{E}F^2(u)
\Big)  \leq 
\exp\Big(-
\frac{1/2n^2(\mathbb{E}F^2(u))^2 }{nA_4+1/3A_3\mathbb{E}F^2(u)}
\Big),
\end{align}
where $A_3 \geq F(u)$ for any $u$ and $A_4 \geq \frac{1}{n}\sum_{u=1}^n\mathbb{E}[F^4(u)]$, which holds true by the fact that $\mathcal{X}$ and $\mathcal{Y}$ are both compact spaces. Combining $(\ref{Bound_V1})$ and $(\ref{Bern})$ shows that there exists some positive constants $A_5>0$ such that
\begin{align*}
V_1 \leq A_5 n^{-1}.
\end{align*}

To bound $V_2$, we first note that for $u \geq 1$
\begin{align}
\label{Equi}
\frac{d u^{-2} \sqrt{\log (\max\{u,1\})}}{du} = 
\frac{1}{u^3 \sqrt{\log(u)}} (-2\log(u)+1).
\end{align}
Setting $(\ref{Equi})$ to 0 yields that $u = \exp(1/2)$, hence $u^{-2} \log (\max\{u,1\})$ is increasing for $1 \leq u \leq \exp(1/2)$ and decreasing for $u >\exp(1/2)$. It follows that for any $A>0$,
\begin{align*}
&\int_{A}^{+\infty}  u^{-2} \sqrt{\log (\max\{u,1\})} du = 
\int_{A}^{\exp(1/2)}  u^{-2} \sqrt{\log (\max\{u,1\})} du+
\int_{\exp(1/2)}^{+\infty}  u^{-2} \sqrt{\log (u)} du \\
\leq &
2^{-1/2}A^{-1}-2^{-1/2}\exp(-1/2)+2^{-1/2}\exp(-1/2)+
\int_{\exp(1/2)}^{+\infty}  \frac{u^{-2}}{\sqrt{\log (u)}} du \\
\leq  &
2^{-1/2}A^{-1}+A^{-a} 
\int_{\exp(1/2)}^{+\infty}  \frac{u^{-2+a}}{\sqrt{\log (u)}} du,
\end{align*}
where $0 \leq a<1$. It can be verified that $\int_{\exp(1/2)}^{+\infty}  \frac{u^{-2+a}}{\sqrt{\log (u)}} du$ is finite when $0 \leq a<1$. Hence, there exists constant $A_6$ such that for any $A>0$
\begin{align}
\label{C_4}
\int_{A}^{+\infty}  u^{-2} \sqrt{\log (\max\{u,1\})} du \leq A_6 A^{-1}.
\end{align}
For $V_2$, by the concavity of $\sqrt{\log(x)}$, it holds that
\begin{align}
\label{V_22}
V_2 = &A_1 n^{-1/2} \mathbb{E} \Big[
\int_{0}^{2\sigma_n(\mathcal{H}_{ij})} \sqrt{|\Theta| 
\log \Big( \max\Big\{\frac{\sqrt{C_1} 2^{j+1} J_0\Vert F \Vert}{v},1\Big\}\Big)}dv I\big(\Vert F \Vert^2 \leq 2\mathbb{E}F^2(u)\big)\Big] \notag \\
\leq &
A_1 n^{-1/2} 
\int_{0}^{2\sqrt{\mathbb{E}\sigma_n^2(\mathcal{H}_{ij})}} \sqrt{|\Theta| 
\log \Big( \max\Big\{\frac{\sqrt{C_1} 2^{j+1} J_0\sqrt{2\mathbb{E}[F^2(u)]}}{v},1\Big\}\Big)}dv\notag \\
=&
A_1 \sqrt{C_1} n^{-1/2} 2^{j+1} J_0\sqrt{2|\Theta| \mathbb{E}[F^2(u)]}
\int_{0}^{\frac{\sqrt{\mathbb{E}\sigma_n^2(\mathcal{H}_{ij})}}{\sqrt{C_1} 2^{j} J_0\sqrt{2\mathbb{E}[F^2(u)]}}} \sqrt{
\log \Big( \max\Big\{\frac{1}{v},1\Big\}\Big)}dv \notag \\
\leq &
A_1 \sqrt{C_1} n^{-1/2} 2^{j+1} J_0\sqrt{2|\Theta| \mathbb{E}[F^2(u)]}
\int_{\frac{\sqrt{C_1} 2^{j+1} J_0\sqrt{\mathbb{E}[F^2(u)]}}{2\sqrt{\mathbb{E}\sigma_n^2(\mathcal{H}_{ij})}} }^{+\infty}u^{-2} \sqrt{
\log (\max\{u,1\})}du\notag \\
\leq &2
A_1 A_6\sqrt{C_1} n^{-1/2} \sqrt{|\Theta|\mathbb{E}\sigma_n^2(\mathcal{H}_{ij})},
\end{align}
where the last inequality follows from $(\ref{C_4})$

As proved in Theorem $2$, we have
\begin{align}
\label{Inequ:meanvari}
\mathbb{E}\sigma_n^2(\mathcal{H}_{ij})  \leq 8 C_{\upsilon} \mathbb{E}\big[ \mathcal{R}(\mathcal{H}_{ij})\big]+
C_5 
\mathbb{E}^{\frac{1}{1+\gamma}}\big[
\Psi_{u}^{\gamma}\big]M^{\frac{\gamma}{\gamma+1}}(i,j),
\end{align}
Plugging $(\ref{Inequ:meanvari})$ into $(\ref{V_22})$ yields that for some positive constants $A_7>0$
\begin{align*}
V_2 \leq 
A_7 \sqrt{n^{-1}|\Theta|} \Big(
 \sqrt{8 C_{\upsilon} \mathbb{E}\big[ \mathcal{R}(\mathcal{H}_{ij})\big]}+C_5
 \mathbb{E}^{\frac{1}{2(1+\gamma)}}\big[
\Psi_{u}^{\gamma}\big](2^i \delta_n)^{\frac{\gamma}{2(\gamma+1)}}
\Big).
\end{align*}
Consequently, we get
\begin{align}
\label{Equation}
\mathbb{E}[\mathcal{R}_{n}(\mathcal{H}_{ij})]  \leq  A_6 n^{-1}
+
A_7 \sqrt{n^{-1}|\Theta|} \Big(
 \sqrt{8 C_{\upsilon}\mathbb{E}\big[ \mathcal{R}(\mathcal{H}_{ij})\big]}+
C_5
 \mathbb{E}^{\frac{1}{2(1+\gamma)}}\big[
\Psi_{u}^{\gamma}\big](2^i \delta_n)^{\frac{\gamma}{2(\gamma+1)}} \Big).
\end{align}
Solving $(\ref{Equation})$ yields that $\mathbb{E}[\mathcal{R}_{n}(\mathcal{H}_{ij})] \leq A_8 n^{-1/2}|\Theta|^{-1/2}  (2^i \delta_{ij})^{\frac{\gamma}{2(\gamma+1)}} \mathbb{E}^{\frac{1}{2(1+\gamma)}}\big[
\Psi_{u}^{\gamma}\big]+A_9 n^{-1}|\Theta|$ for some constants $A_8,A_9>0$.

Provided that $\big(n^{-1}|\Theta| \log  \big(|\Theta|^{-1}n )\big)^{\frac{\gamma+1}{\gamma+2}} \mathbb{E}^{\frac{1}{\gamma+2}}\big(\Psi_u^{\gamma}\big)=O(\delta_n)$, it follows that
\begin{align*}
\frac{\mathbb{E}[\mathcal{R}_{n}(\mathcal{H}_{ij})] }{M(i,j)} \leq 
\frac{A_8 n^{-1/2} |\Theta|^{1/2} \mathbb{E}^{\frac{1}{2(1+\gamma)}}\big[
\Psi_{u}^{\gamma}\big]}{(2^{i}\delta_n)^{\frac{\gamma+2}{2(\gamma+1)}}} +
\frac{A_9n^{-1}|\Theta|}{2^{i}\delta_n}
 \leq 1/4,
\end{align*}
where the last inequality holds when $n$ goes to infinity, which then implies $\mathbb{E}\big(\sup_{f \in \mathcal{H}_{ij}} 
U_n(f)\big) \leq 1/2 M(i,j)$, and this completes the proof. \qed

\noindent
\textbf{Proof of Theorem \ref{Thm: Main}}: By Assumption 2, it holds that for  $\delta_n>0$,
\begin{align*}
\mathbb{P}\big( e(\widetilde{f} , f^*) \geq \delta_n^{\alpha} \big) 
\leq 
\mathbb{P}\big( e_{\upsilon}(\widetilde{f},f^*) \geq \delta_n \big).
\end{align*}
Next, we proceed to bound $\mathbb{P}\big( e_{\phi}(\widehat{f},f^*_{\phi}) \geq \delta_n \big)$. By the definition of $\widehat{f}$, we have
\begin{align}
\label{I_bound}
\mathbb{P}\Big(
e_{\upsilon}(\widetilde{f},f^*)\geq \delta_n
\Big) \leq 
\mathbb{P}\Big( \sup_{f \in  \mathcal{H}} 
\widetilde{R}_{n,\upsilon}(f^*_{\mathcal{F}})+\lambda_n J_0-
\widetilde{R}_{n,\upsilon}(f)-\lambda_n J(f) \geq 0
\Big) \equiv I,
\end{align}
where $J_0 = \max\{J(f^*_{\mathcal{F}}),1 \}$.  Let $i_0$ denote the integer such that $2^{i_0} \delta_n \geq \sup_{f}e_{\upsilon}(f,f^*)$ and $j_0$ denote the integer such that for any $f \in \mathcal{F}$ satisfies $J(f) \geq 2^{j_0} J_0$ must have $\Vert \Theta_f \Vert_{min} \geq C_{\mathcal{F}}$, where $\Vert \cdot \Vert_{min}$ denotes the minimum absolute value of $ \Theta_f$.

 For $0 \leq i \leq i_0$ and $0 \leq j \leq j_0$, we define
\begin{align*}
&\mathcal{H}_{ij} = \big\{ f \in \mathcal{F}:2^{i-1}\delta_n < e_{\upsilon}(f , f^*) \leq 2^i \delta_n, 2^{j-1} J_0 <J(f)\leq 2^j J_0  \big\}, 1 \leq i \leq i_0, 1 \leq j \leq j_0 \\
& \mathcal{H}_{i0}= \big\{ f\in \mathcal{F}: 2^{i-1}\delta_n < e_{\upsilon}(f , f^*)   \leq 2^i \delta_n,J(f) \leq J_0 \big\}, 1 \leq i \leq i_0.
\end{align*}
It is easy to verify that $\mathcal{H}$ can be represented as $\mathcal{H} =\cup_{i=1}^{i_0} \cup_{j=0}^{j_0} \mathcal{H}_{ij}$. With this, $(\ref{I_bound})$ can be upper bounded as
\begin{align*}
I  
=&
\mathbb{P}\Big( \sup_{f \in \cup_{i=1}^{i_0} \cup_{j=0}^{j_0} \mathcal{H}_{ij}} 
\widetilde{R}_{n,\upsilon}(f^*_{\mathcal{F}})+\lambda_n J_0-
\widetilde{R}_{n,\upsilon}(f)-\lambda_n J(f) \geq 0
\Big) \\
 \leq &
 \sum_{i=0}^{i_0} \sum_{j=0}^{j_0}
\mathbb{P}\Big( \sup_{f \in \mathcal{H}_{ij}} 
\widetilde{R}_{n,\upsilon}(f^*_{\mathcal{F}})+\lambda_n J_0-
\widetilde{R}_{n,\upsilon}(f)-\lambda_n J(f) \geq 0
\Big) \\
=&
 \sum_{i=1}^{i_0}\sum_{j=1}^{j_0} \mathbb{P}\Big( \sup_{f \in \mathcal{H}_{ij}} 
\widetilde{R}_{n,\upsilon}(f^*_{\mathcal{F}})+\lambda_n J_0-
\widetilde{R}_{n,\upsilon}(f)-\lambda_n J(f) \geq 0
\Big) \\
&+ \sum_{i=1}^{i_0} \mathbb{P}\Big( \sup_{f \in \mathcal{H}_{i0}} 
\widetilde{R}_{n,\upsilon}(f^*_{\mathcal{F}})+\lambda_n J_0-
\widetilde{R}_{n,\upsilon}(f)-\lambda_n J(f) \geq 0
\Big)   
\equiv  I_1 + I_2.
\end{align*}
Therefore, it suffices to bound $I_1$ and $I_2$ separately. By Assumption \ref{APPROX}, for any $1 \leq i \leq i_0$ and $j \geq 0$, we get
\begin{align}
\label{Sieve_I1I2}
&\mathbb{P}\Big( \sup_{f \in \mathcal{H}_{ij}} 
\widetilde{R}_{n,\upsilon}(f^*_{\mathcal{F}})+\lambda_n J_0-
\widetilde{R}_{n,\upsilon}(f)-\lambda_n J(f) \geq 0
\Big)  \notag \\
\leq &\mathbb{P}\Big(  \sup_{f \in \mathcal{H}_{ij}} \big(
\widetilde{R}_{n,\upsilon}(f^*_{\mathcal{F}})-
\widetilde{R}_{n,\upsilon}(f)  - \widetilde{R}_\upsilon(f^*_{\mathcal{F}})+ \widetilde{R}_\upsilon(f) \big) \notag \\
& \qquad \qquad \geq \lambda_n
\inf_{f \in \mathcal{H}_{ij}} (J(f) - J_0)
+\inf_{f \in \mathcal{H}_{ij}} \widetilde{R}_\phi(f) -\widetilde{R}_\upsilon(f^*_{\mathcal{F}})
\Big)  \notag \\
\leq & \mathbb{P}\Big(  \sup_{f \in \mathcal{H}_{ij}}\big(
\widetilde{R}_{n,\upsilon}(f^*_{\mathcal{F}})-
\widetilde{R}_{n,\upsilon}(f)  - \widetilde{R}_\upsilon(f^*_{\mathcal{F}})+ \widetilde{R}_\upsilon(f) \big)  \geq \lambda_n
(2^{j-1}-1) J_0
+2^{i-1}\delta_n 
\Big) \notag \\
\leq &\mathbb{P}\Big(  \sup_{f \in \mathcal{H}_{ij}} \big(
\widetilde{R}_{n,\upsilon}(f^*_{\mathcal{F}})-
\widetilde{R}_{n,\upsilon}(f) - \widetilde{R}_\upsilon(f^*_{\mathcal{F}})+ \widetilde{R}_\upsilon(f)  \big)  \geq \lambda_n
(2^{j-1}-1) J_0
+2^{i-1}\delta_n
\Big)\notag  \\
\leq & \mathbb{P}\Big(  \sup_{f \in \mathcal{H}_{ij}} \big(
\widetilde{R}_{n,\upsilon}(f^*_{\mathcal{F}})-
\widetilde{R}_{n,\upsilon}(f)  - \widetilde{R}_\upsilon(f^*_{\mathcal{F}})+ \widetilde{R}_\upsilon(f) \big)  \geq M(i,j)
\Big),
\end{align}
where $M(i,j) =\lambda_n(2^{j-1}-1) J_0+2^{i-1}\delta_n$ for $1 \leq i \leq i_0$ and $j \geq 1$ and $M(i,0) = 2^{i_0}\delta_n$ for $i \geq 1$. Further, we define 
$$G_f(u) =\frac{1}{m(m-1)}\sum\limits_{i \neq j}I\Big(\widetilde{\phi}_u(\mathcal{I}_i)>\widetilde{\phi}_u(\mathcal{I}_j)\Big) \upsilon(g_{f}(u,i,j)).$$ Then the right-hand side of $(\ref{Sieve_I1I2})$ can be written as
\begin{align}
\label{Step_2}
 &\mathbb{P}\Big(  \sup_{f \in \mathcal{H}_{ij}} \big(
\widetilde{R}_{n,\upsilon}(f^*_{\mathcal{F}})-
\widetilde{R}_{n,\upsilon}(f)  - \widetilde{R}_\upsilon(f^*_{\mathcal{F}})+ \widetilde{R}_\upsilon(f) \big)  \geq M(i,j)
\Big) \notag \\
=&
\mathbb{P}\Big(  \sup_{f \in \mathcal{H}_{ij}} 
\big[\frac{1}{n}\sum_{u=1}^n \big(G_{f^*_{\mathcal{F}}}(u) - G_f(u)\big) - 
\mathbb{E} \big(G_{f^*_{\mathcal{F}}}(u) - G_f(u)\big)\big] \geq M(i,j) \Big)\notag \\
=&
 \mathbb{P}\Big(  \sup_{f \in \mathcal{H}_{ij}} 
\big[\frac{1}{n}\sum_{u=1}^n \big(DG_f(u)\big) - 
\mathbb{E} \big(DG_f(u)\big)\big] \geq M(i,j) \Big),
\end{align}
where $DG_f(u) =G_{f^*_{\mathcal{F}}}(u) - G_f(u)$. Let $U_n(f) =n^{-1}\sum_{u=1}^n \big[DG_f(u) - \mathbb{E} \big(DG_f(u)\big) \big]$, then $(\ref{Step_2})$ can be re-written as 
\begin{align}
\label{Talagrand_pre}
 &\mathbb{P}\Big(  \sup_{f \in \mathcal{H}_{ij}} 
\big[\frac{1}{n}\sum_{u=1}^n \big(DG_f(u)\big) - 
\mathbb{E} \big(DG_f(u)\big)\big] \geq M(i,j) \Big) \notag \\
=& 
 \mathbb{P}\Big(  \sup_{f \in \mathcal{H}_{ij}} 
U_n(f) - \mathbb{E}\big(\sup_{f \in \mathcal{H}_{ij}} 
U_n(f)\big)\geq M(i,j)- \mathbb{E}\big(\sup_{f \in \mathcal{H}_{ij}} 
U_n(f)\big) \Big)\notag \\
\leq &
 \mathbb{P}\Big(  \sup_{f \in \mathcal{H}_{ij}} 
U_n(f) - \mathbb{E}\big(\sup_{f \in \mathcal{H}_{ij}} 
U_n(f)\big)\geq 1/2M(i,j) \Big),
\end{align}
where the last inequality follows from Lemma \ref{Bound_Mean}. With this, we have
\begin{align*}
I_1 &
\leq \sum_{i=1}^{i_0}\sum_{j=1}^{j_0}
 \mathbb{P}\Big(  \sup_{f \in \mathcal{H}_{ij}} 
U_n(f) - \mathbb{E}\big(\sup_{f \in \mathcal{H}_{ij}} 
U_n(f)\big)\geq 1/2M(i,j) \Big), \\
I_2&   
\leq \sum_{i=1}^{i_0}
 \mathbb{P}\Big(  \sup_{f \in \mathcal{H}_{i0}} 
U_n(f) - \mathbb{E}\big(\sup_{f \in \mathcal{H}_{i0}} 
U_n(f)\big)\geq 1/2M(i,j) \Big).
\end{align*}
Clearly, it suffices to bound $ \mathbb{P}\big(  \sup_{f \in \mathcal{H}_{ij}} 
U_n(f) - \mathbb{E}\big(\sup_{f \in \mathcal{H}_{ij}} 
U_n(f)\big)\geq 1/2M(i,j) \big)$ for any $1 \leq i\leq i_0 $ and $0\leq j \leq j_0$, to which we apply the Talagrand's inequality in the following step. 

Denote that $\mathbb{E}[\sigma_n^2(\mathcal{H}_{ij})]=\mathbb{E}[\sup_{f \in \mathcal{H}_{ij}}n^{-1}\sum_{i=1}^n DG_f^2(u)]$. Then, we establish the relation between $\mathbb{E}[\sigma_n^2(\mathcal{H}_{ij})]$ and $M(i,j)$. By a symmetrization argument, we have
\begin{align}
\label{Process:tala_vari}
\mathbb{E}[\sigma_n^2(\mathcal{H}_{ij})] 
\leq&
\mathbb{E}\Big[\sup_{f \in \mathcal{H}_{ij}}\frac{1}{n}\sum_{u=1}^n  DG_f^2(u)-\mathbb{E}\big(DG_f^2(u)\big) \Big]+
\sup_{f \in \mathcal{H}_{ij}}\frac{1}{n}\sum_{u=1}^n\mathbb{E}\big(DG_f^2(u)\big)\notag \\
\leq &8 C_{\upsilon} \mathbb{E}\big[ \mathcal{R}(\mathcal{H}_{ij})\big]+ \sup_{f\in \mathcal{H}_{ij}} 
\frac{1}{n}\sum_{u=1}^n \mathbb{E}\Big[
\big( G_{f^*_{\mathcal{F}}}(u) - G_f(u) \big)^2\Big].
\end{align}
The right-hand side of $(\ref{Process:tala_vari})$ can be bounded by
\begin{align*}
&\sup_{f\in \mathcal{H}_{ij}}\frac{1}{n}\sum_{u=1}^n \mathbb{E}\Big[
\big( G_{f^*_{\mathcal{F}}}(u) - G_f(u) \big)^2\Big] \\
\leq& \frac{2}{n}\sum_{u=1}^n \mathbb{E}\Big[
\big( G_{f^*_{\mathcal{F}}}(u) -G_{f^*}(u)\big)^2\Big]+  \sup_{f\in \mathcal{H}_{ij}} \frac{2}{n}\sum_{u=1}^n\mathbb{E}\Big[\big( G_{f^*}(u)-  G_{f}(u) \big)^2\Big] \\
\leq &
\frac{2}{n}\sum_{u=1}^n \text{Var}
\big( G_{f^*_{\mathcal{F}}}(u) - G_{f^*}(u)\big)+  \sup_{f\in \mathcal{H}_{ij}} \frac{2}{n}\sum_{u=1}^n\text{Var}\big(  G_{f^*}(u)-  G_u(f) \big)\notag \\
& + 2 e_{\upsilon}^2(f^*_{\mathcal{F}},f^*)+2 \sup_{f\in \mathcal{H}_{ij}}e_{\upsilon}^2(f,f^*_{\mathcal{F}}) \\
\leq &
4 C_4
\mathbb{E}^{\frac{1}{1+\gamma}}\big[
\Psi_{u}^{\gamma}\big]\sup_{f\in \mathcal{H}_{ij}}
\big(e_{\upsilon}(f,f^*)\big)^{\gamma/(\gamma+1)} + \sup_{f\in \mathcal{H}_{ij}}e_{\upsilon}^2(f,f^*_{\mathcal{F}}).
\end{align*}
Suppose that $\mathbb{E}\big[\Psi_{u}^{\gamma}\big] = O(\delta_n)$. Since $\gamma/(\gamma+1)<2$, there exists some constants $C_5$ such that
\begin{equation}
\label{Expect_second}
\sup_{f\in \mathcal{H}_{ij}}\frac{1}{n}\sum_{u=1}^n \mathbb{E}\Big[
\big( G_{f^*_{\mathcal{F}}}(u) - G_f(u) \big)^2\Big] \leq C_5 
\mathbb{E}^{\frac{1}{1+\gamma}}\big[
\Psi_{u}^{\gamma}\big]M^{\frac{\gamma}{\gamma+1}}(i,j).
\end{equation}
Combined with Lemma \ref{Bound_Mean}, it holds that
\begin{align*}
\mathbb{E}\big[\sigma_n^2(\mathcal{H}_{ij})\big]  \leq 2 C_{\upsilon} M(i,j)+
C_5 
\mathbb{E}^{\frac{1}{1+\gamma}}\big[
\Psi_{u}^{\gamma}\big]M^{\frac{\gamma}{\gamma+1}}(i,j),
\end{align*}
where $C_{\upsilon}$ such that $G_f(u) \leq C_{\upsilon}$ for any $f \in \mathcal{F}$. Since $M(i,j)$ is asymptotically smaller then 1, there exists some constants $T_4$ such that
\begin{align}
\label{Vari_Bound}
\mathbb{E}\sigma_n^2(\mathcal{H}_{ij})
\leq  T_4 \mathbb{E}^{\frac{1}{1+\gamma}}\big[
\Psi_{u}^{\gamma}\big] M^{\frac{\gamma}{\gamma+1}}(i,j),
\end{align}
for $1 \leq i \leq i_0$ and $0 \leq j \leq j_0$. Then, plugging $(\ref{Vari_Bound})$ into $(\ref{Talagrand_pre})$ yields that for some constants $T_5>0$
\begin{align*}
 &\mathbb{P}\Big(  \sup_{f \in \mathcal{H}_{ij}} 
U_n(f) - \mathbb{E}\big(\sup_{f \in \mathcal{H}_{ij}} 
U_n(f)\big)\geq 1/2M(i,j) \Big)\notag \\
\leq &
T_1 \exp \Big(
- \frac{nM(i,j)}{2BT_1} \log\big(1+\frac{B \big(M(i,j)\big)}{T_4 \mathbb{E}^{\frac{1}{1+\gamma}}(
\Psi_{u}^{\gamma}) M^{\gamma/(\gamma+1)}(i,j)} \big)
\Big) \\
\leq &
T_1 \exp \Big(
-\frac{T_5}{2T_1T_4}
n \mathbb{E}^{-\frac{1}{1+\gamma}}(
\Psi_{u}^{\gamma}) \big(M(i,j)\big)^{\frac{\gamma+2}{\gamma+1}}
\Big),
\end{align*}
where $T_5$ satisfies that $\log(1+x) \geq T_5 x$ for $x \in [0,(T_4 \mathbb{E}^{\frac{1}{1+\gamma}}(
\Psi_{u}^{\gamma}) )^{-1}B \big(M(i,j)\big)^{1/(1+\gamma)}]$, which holds true by the boundedness of $M(i,j)$. 

We let $T_6 = \frac{T_5}{2T_1T_4}$, then $I_3$ can be bounded as
\begin{align*}
I_1 
= &
\sum_{i=1}^{i_0}\sum_{j=1}^{j_0} \mathbb{P}\Big(  \sup_{f \in \mathcal{H}_{ij}} 
U_n(f) - \mathbb{E}\big(\sup_{f \in \mathcal{H}_{ij}} 
U_n(f)\big)\geq 1/2M(i,j) \Big)
 \\
\leq & \sum_{i=1}^{i_0}\sum_{j=1}^{j_0}
T_1 \exp \Big\{
-T_6 n\mathbb{E}^{-\frac{1}{1+\gamma}}(
\Psi_{u}^{\gamma}) \Big( \delta_{n}^{\frac{\gamma+2}{\gamma+1}}  (2^{i-1})^{\frac{\gamma+2}{\gamma+1}}
-(\lambda_nJ_0)^{\frac{\gamma+2}{\gamma+1}}(2^{j-1}-1)^{\frac{\gamma+2}{\gamma+1}}\Big)
\Big\}\\
\leq & \sum_{i=2}^\infty\sum_{j=1}^{\infty}
T_1 \exp\Big\{
-T_6 n\mathbb{E}^{-\frac{1}{1+\gamma}}(
\Psi_{u}^{\gamma}) \Big(\delta_{n}^{\frac{\gamma+2}{\gamma+1}} i
-(\lambda_nJ_0)^{\frac{\gamma+2}{\gamma+1}}(j-1)\big)
\Big) \Big\}\\
&+
\sum_{j=1}^{\infty}
T_1 \exp\Big\{ \Big(
-T_6 n\mathbb{E}^{-\frac{1}{1+\gamma}}(
\Psi_{u}^{\gamma})
\Big(\delta_{n}^{\frac{\gamma+2}{\gamma+1}} 
-(\lambda_nJ_0)^{\frac{\gamma+2}{\gamma+1}}(j-1)\Big)
\Big\}  \\
\leq &
T_1 \frac{\exp(-2T_6 n
\mathbb{E}^{-\frac{1}{1+\gamma}}(\Psi_{u}^{\gamma})
\delta_{n}^{\frac{\gamma+2}{\gamma+1}}  )}{1-\exp(-T_6 n
\mathbb{E}^{-\frac{1}{1+\gamma}}(\Psi_{u}^{\gamma})
\delta_{n}^{\frac{\gamma+2}{\gamma+1}}  ))}
\frac{1}{1-\exp(-T_6n
\mathbb{E}^{-\frac{1}{1+\gamma}}(\Psi_{u}^{\gamma})
 )(\lambda_nJ_0)^{\frac{\gamma+2}{\gamma+1}})} \\
&+
T_1\exp(-T_6 n
\mathbb{E}^{-\frac{1}{1+\gamma}}(\Psi_{u}^{\gamma})
\delta_{n}^{\frac{\gamma+2}{\gamma+1}} )
\frac{1}{1-\exp(-T_6n
\mathbb{E}^{-\frac{1}{1+\gamma}}(\Psi_{u}^{\gamma})
 )(\lambda_nJ_0)^{\frac{\gamma+2}{\gamma+1}})}  \\
\leq &
6 T_1\exp\Big(-T_6 n
\mathbb{E}^{-\frac{1}{1+\gamma}}(\Psi_{u}^{\gamma})
\delta_{n}^{\frac{\gamma+2}{\gamma+1}} \Big),
\end{align*}
where the second last inequality follows from Cauchy-Schwartz inequality and the last inequality holds true when $n$ goes to infinity.

Similarly, for $I_2$, we have
\begin{align*}
I_4 = &
\sum_{i=1}^{i_0}\mathbb{P}\Big(  \sup_{f \in \mathcal{H}_{i0}} 
U_n(f) - \mathbb{E}\big(\sup_{f \in \mathcal{H}_{i0}} 
U_n(f)\big)\geq 1/2M(i,j) \Big)  \\
\leq &
 \sum_{i=1}^{i_0}
T_1 \exp \Big(
-T_6 n\mathbb{E}^{-\frac{1}{1+\gamma}}(\Psi_{u}^{\gamma}) \delta_{n}^{\frac{\gamma+2}{\gamma+1}}  (2^{i-1})^{\frac{\gamma+2}{\gamma+1}}
+T_6 n\mathbb{E}^{-\frac{1}{1+\gamma}}(\Psi_{u}^{\gamma}) (\lambda_nJ_0)^{\frac{\gamma+2}{\gamma+1}}
\Big) \\
\leq &
 \sum_{i=1}^{i_0}
T_1 \exp \Big(
-T_6 n \mathbb{E}^{-\frac{1}{1+\gamma}}(\Psi_{u}^{\gamma}) \delta_{n}^{\frac{\gamma+2}{\gamma+1}} (2^{i-1}-1/2)^{\frac{\gamma+2}{\gamma+1}}
\Big) \\
\leq &
T_1\frac{ \exp\big(
-2^{-1}T_6n\mathbb{E}^{-\frac{1}{1+\gamma}}(\Psi_{u}^{\gamma}) \delta_{n}^{\frac{\gamma+2}{\gamma+1}}
\big)}{1-\exp\big(-T_6n\mathbb{E}^{-\frac{1}{1+\gamma}}(\Psi_{u}^{\gamma}) \delta_{n}^{\frac{\gamma+2}{\gamma+1}}\big)}
\leq 
2 T_1 \exp\big(
-T_6n\mathbb{E}^{-\frac{1}{1+\gamma}}(\Psi_{u}^{\gamma}) \delta_{n}^{\frac{\gamma+2}{\gamma+1}}
\big).
\end{align*}
Finally, by setting $C_2 = T_1$ and $C_3 = T_6$ we get
\begin{align*}
I_1 + I_2 \leq 
8C_2  \exp\big(
-C_3n\mathbb{E}^{-\frac{1}{1+\gamma}}(\Psi_{u}^{\gamma}) \delta_{n}^{\frac{\gamma+2}{\gamma+1}}
\big).
\end{align*}
This completes the proof. \qed

\noindent
\textbf{Proof of Corollary \ref{Coro_1}}: First, by the property (2) in Theorem \ref{Thm: Main}, we have
\begin{align*}
\frac{|2\eta_{uij}-1|}{|2\widetilde{\eta}_{uij}-1|} \leq 
\frac{\exp\big((|\Omega|-1)^{-1}\epsilon_u\big)+1}{\exp\big((|\Omega|-1)^{-1}\epsilon_u\big)-1}=\Psi_u.
\end{align*}
By the definition of excess risk, we have
\begin{align*}
&\widetilde{R}(f) - \widetilde{R}(f^*) =\frac{1}{m(m-1)} \mathbb{E}_{u}\Big[\sum_{i<j}|2\widetilde{\eta}_{uij}-1|I\Big(\big(f(\bm{x}_u,\bm{y}_i)-f(\bm{x}_u,\bm{y}_j)\big)\big(f^*(\bm{x}_u,\bm{y}_i)-f^*(\bm{x}_u,\bm{y}_j)\big)<0\Big)\Big].
\end{align*}
Therefore, we have
\begin{align*}
&R(f) - R(f^*) \\
=&\frac{1}{m(m-1)} \mathbb{E}_{u}\Big[\sum_{i<j}|2\eta_{uij}-1|I\Big(\big(f(\bm{x}_u,\bm{y}_i)-f(\bm{x}_u,\bm{y}_j)\big)\big(f^*(\bm{x}_u,\bm{y}_i)-f^*(\bm{x}_u,\bm{y}_j)\big)<0\Big)\Big]\\
\leq &
\frac{1}{m(m-1)} \mathbb{E}_{u}\Big[\Psi_u\sum_{i<j}|2\widetilde{\eta}_{uij}-1|I\Big(\big(f(\bm{x}_u,\bm{y}_i)-f(\bm{x}_u,\bm{y}_j)\big)\big(f^*(\bm{x}_u,\bm{y}_i)-f^*(\bm{x}_u,\bm{y}_j)\big)<0\Big)\Big] \\
\leq &
\mathbb{E}_{u}(\Psi_u)
\frac{1}{m(m-1)} \mathbb{E}_{u}\Big[\sum_{i<j}|2\widetilde{\eta}_{uij}-1|I\Big(\big(f(\bm{x}_u,\bm{y}_i)-f(\bm{x}_u,\bm{y}_j)\big)\big(f^*(\bm{x}_u,\bm{y}_i)-f^*(\bm{x}_u,\bm{y}_j)\big)<0\Big)\Big] \\
=&
\mathbb{E}_{u}(\Psi_u)\Big(\widetilde{R}(f) - \widetilde{R}(f^*)\Big).
\end{align*}
It then follows from Theorem \ref{Thm: Main} that
\begin{align*}
R(\widetilde{f}) - R(f^*)= O_p\Big(
\mathbb{E}(\Psi_{u}) 
\mathbb{E}^{\frac{\alpha}{\gamma+2}}(\Psi_{u}^{\gamma})  \big(|\Theta| n^{-1}\log(n/|\Theta|)\big)^{\frac{\alpha(\gamma+1)}{\gamma+2}}\Big).
\end{align*}
If $\epsilon_u=\epsilon$ for any user $u$, $\alpha=1$, and $\epsilon=o(1)$, we have
\begin{align*}
\Psi_u = \frac{\exp\big((|\Omega|-1)^{-1}\epsilon_u\big)+1}{\exp\big((|\Omega|-1)^{-1}\epsilon_u\big)-1} \asymp \frac{1}{\epsilon}.
\end{align*}
Therefore, 
\begin{align*}
\mathbb{E}(\Psi_{u}) 
\mathbb{E}^{\frac{\alpha}{\gamma+2}}(\Psi_{u}^{\gamma})  \big(|\Theta| n^{-1}\log(n/|\Theta|)\big)^{\frac{\alpha(\gamma+1)}{\gamma+2}} 
\asymp
(\Psi_{u}^{\gamma})  \big(|\Theta| (\epsilon^2 n)^{-1}\log(n/|\Theta|)\big)^{\frac{\alpha(\gamma+1)}{\gamma+2}} .
\end{align*}
This completes the proof. \qed

\end{document}